\documentclass[a4paper,11pt,fleqn]{article}

\usepackage{amsmath}
\usepackage{amsthm} 
\usepackage{amssymb}  
\usepackage{mathrsfs}
\usepackage{graphicx}
\usepackage{array}
\usepackage{xspace}		
\usepackage[english,ruled,vlined]{algorithm2e}	
\usepackage{color}

\usepackage{scalefnt}
\usepackage{todonotes}
\usepackage{subcaption}
\usepackage{geometry}
\usepackage{enumitem}

\usepackage{cite}
\usepackage[normalem]{ulem}
\usepackage{amsmath,amssymb,amsfonts}
\usepackage{algorithmic}
\usepackage{graphicx}
\usepackage{grffile}
\usepackage{epstopdf}
\usepackage{float}
\usepackage{amsthm}
\usepackage{textcomp}
\usepackage{soul}
\usepackage{xcolor,cancel}
\usepackage{rotating}
\usepackage{multirow}
\usepackage{xcolor}
\usepackage{multirow}

\usepackage{xfakebold}

\renewcommand{\baselinestretch}{1.1}
\newcommand{\fbseries}{\unskip\setBold\aftergroup\unsetBold\aftergroup\ignorespaces}
\makeatletter
\newcommand{\setBoldness}[1]{\def\fake@bold{#1}}
\makeatother

\def\BibTeX{{\rm B\kern-.05em{\sc i\kern-.025em b}\kern-.08em
    T\kern-.1667em\lower.7ex\hbox{E}\kern-.125emX}}


\newtheorem{proposition}{Proposition}[section]
\newtheorem{theorem}{Theorem}[section]

\newtheorem{corollary}{Corollary}

\newcommand{\datad}{\mathrm{z}}
\newcommand{\estmsd}{\mathrm{x}}
\newcommand{\dualvar}{\mathrm{u}}
\newcommand{\datar}{\mathrm{y}}

\newcommand{\Id}{\ensuremath{\operatorname{Id}}}
\newcommand{\UNN}{PNN}

\newcommand{\prox}{\ensuremath{\operatorname{prox}}}
\newcommand{\proj}{\ensuremath{\operatorname{P}}}
\newcommand{\argmin}[1]{\ensuremath{\rm \underset{#1}{argmin} \;}}
\newcommand{\Argmin}[1]{\ensuremath{\rm \underset{#1}{Argmin} \;}}

\newcommand{\RR}{\ensuremath{\mathbb{R}}}

\newcommand{\D}{\mathrm{D}}
\newcommand{\A}{\mathrm{A}}
\newcommand{\Wrm}{\mathrm{W}}
\newcommand{\Vrm}{\mathrm{V}}
\newcommand{\Lrm}{\mathrm{L}}
\newcommand{\brm}{\mathrm{b}}
\newcommand{\NN}{\mathbb{N}}

\renewcommand\footnotemark{}

\definecolor{color_ista}{rgb}{1, 0., 0.}             \definecolor{color_fista}{rgb}{1, 0.59, 0.6}       
\definecolor{color_cp}{rgb}{0., 0.4 0.8}             \definecolor{color_sccp}{rgb}{0. , 0.6, 0.3}                   
\definecolor{DDFB_LNO}{rgb}{0.682, 0.110, 0.118}
\definecolor{DDFB_LFO}{rgb}{0.682, 0.451, 0.451}
\definecolor{DDIFB_LNO}{rgb}{0.996, 0.682, 0}
\definecolor{DDIFB_LFO}{rgb}{1, 1, 0.}
\definecolor{DCP_LNO}{rgb}{0.180, 0.325, 0.682}
\definecolor{DCP_LFO}{rgb}{0.478, 0.545, 0.682}
\definecolor{DScCP_LNO}{rgb}{0.133, 0.682, 0.514}
\definecolor{DScCP_LFO}{rgb}{0.475, 0.682, 0.616}
\definecolor{airforceblue}{rgb}{0.36, 0.54, 0.66}                    
  
\graphicspath{
	{figures/},
	{figures/u-v/},
}
\makeatletter
	\def\input@path{%
		{figures/},%
	}
\makeatother

\begin{document}

\title{Unfolded proximal neural networks for robust image Gaussian denoising
\thanks{This work was partly supported by the ANR (Agence Nationale de la Recherche) from France ANR-19-CE48-0009 Multisc'In,  Fondation Simone et Cino Del Duca (Institut de France), the CBP (Centre Blaise Pascal-ENS lyon), the Royal Society of Edinburgh, and the EPSRC grant EP/X028860.}}
\author{Hoang Trieu Vy Le$^\star$, Audrey Repetti$^\dagger$, and Nelly Pustelnik$^\star$ \\[0.3cm] 
\small{$^\star$ Laboratoire de Physique, ENSL, CNRS UMR 5672, F-69342, Lyon, France} \\
\small{(e-mail: hoang.le@ens-lyon.fr, nelly.pustelnik@ens-lyon.fr)} \\[0.2cm]
\small{$^\dagger$ School of Engineering and Physical Sciences and School of Mathematical and Computer Sciences} \\ 
\small{Heriot-Watt University, Edinburgh, EH14 4AS, UK (e-mail: a.repetti@hw.ac.uk)}
}
\date{}
\maketitle

\begin{abstract}
A common approach to solve inverse imaging problems relies on finding a maximum \textit{a posteriori} (MAP) estimate of the original unknown image, by solving a minimization problem. 
In this context, iterative proximal algorithms are widely used, enabling to handle non-smooth functions and linear operators. 
Recently, these algorithms have been paired with deep learning strategies, to further improve the estimate quality.
In particular, \textit{proximal neural networks} ({\UNN}s) have been introduced, obtained by unrolling a proximal algorithm as for finding a MAP estimate, but over a fixed number of iterations, with learned linear operators and parameters. 
As {\UNN}s are based on optimization theory, they are very flexible, and can be adapted to any image restoration task, as soon as a proximal algorithm can solve it. They further have much lighter architectures than traditional networks.
In this article we propose a unified framework to build {\UNN}s for the Gaussian denoising task, based on both the dual-FB and the primal-dual Chambolle-Pock algorithms. 
We further show that accelerated inertial versions of these algorithms enable skip connections in the associated NN layers. 
We propose different learning strategies for our {\UNN} framework, and investigate their robustness (Lipschitz property) and denoising efficiency. 
Finally, we assess the robustness of our {\UNN}s when plugged in a forward-backward algorithm for an image deblurring problem.

\end{abstract}


\section{Introduction}

Image denoising aims to find an estimate of an unknown image $\overline{\estmsd} \in \mathbb{R}^N$, from noisy measurements $\datad \in \mathbb{R}^N$. The present contribution focuses on the Gaussian denoising problem
\begin{equation}\label{denoising-model}
   \datad=\overline{\estmsd}+ \mathrm{b},
\end{equation}
where $\mathrm{b} \in \mathbb{R}^M$ models an additive white Gaussian noise with standard deviation $\delta>0$. 
A common method to denoise $\datad$ is to rely on a maximum \textit{a posteriori} (MAP) approach, and to define the estimate ${\widehat{\estmsd}_{\textrm{MAP}}} \in \mathbb{R}^N$ as a minimizer of a penalized least-squares objective function. A general formulation of this problem is to find
\begin{align}\label{prob:2-terms-minimization}
	{\widehat{\estmsd}_{\text{MAP}}} 
    = \argmin{\estmsd\in \RR^N} 
 F(\estmsd), 
\end{align}
where 
\begin{equation}    \label{eq:prox-function}
    F(\estmsd): =\frac{1}{2}\|\estmsd-\datad\|^2_2+ \nu g(\D\estmsd)+\iota_{C}(\estmsd),
\end{equation}
$C \subset \RR^N$ is a closed, convex, non-empty constraint set, $\nu>0$ is a regularization parameter proportional to $\delta^2$, $\D\colon \RR^N \to \RR^{\mid\mathbb{F}\mid }$ is a linear operator mapping an image from $\RR^N$ to a feature space $\RR^{\mid\mathbb{F}\mid}$, and $g \colon \RR^{\mid\mathbb{F}\mid}\to (-\infty,+\infty]$ denotes a proper, lower-semicontinous, convex function. 
 The function $g$ and the operator $\D$ are chosen according to  the type of images of interest. For instance, functions of choice for piece-wise constant images are those in the family of total variation (TV) regularizations \cite{rudin1992nonlinear}, which can be expressed as an $\ell_1$ (or an $\ell_{1,2}$) norm composed with a linear operator performing horizontal and vertical finite differences of the image. More generally, $\D$ can be chosen as a sparsifying operator (e.g., wavelet transform \cite{Mallat_S_1997_book_wav_tsp, Jacques_L_2011_j-sp_pan_mgr, Pustelnik_N_20016_j-w-enc-eee_wav_bid}), and $g$ as a function promoting sparsity (e.g., $\ell_1$).
For any choice of $g$ and $\D$, the parameter $\nu>0$ is used to balance the penalization term (i.e., function $g\circ \D$) with the data-fidelity term (i.e., least-squares function).

When $C=\RR^N$, and for simple choices of $g\circ \D$, \eqref{prob:2-terms-minimization} may have a closed form solution (see, e.g., \cite{chierchia2020proximity}, and references therein).
Otherwise, \eqref{eq:prox-function} can be minimized efficiently using proximal splitting methods \cite{Bauschke_H_2011_book_con_amo, Combettes2011, Chambolle_A_2016_an}. 
The most appropriate algorithm will be chosen depending on the properties of $g$, $\D$ and $C$. 
For simple sets $C$ and when $g \circ \D$ is proximable, Douglas-Rachford (DR) scheme \cite{Combettes_PL_2007_istsp_Douglas_rsatncvsr} can be considered, alternating, at each iteration, between a proximity step on the sum of the least-square term  and the indicator function, and a proximity step on $g \circ \D$. However, when $g \circ \D$ is not proximable nor differentiable (e.g., TV penalization, or when $\D$ is a redundant wavelet transform), more advanced algorithms must be used, splitting all the terms in~\eqref{eq:prox-function} to handle them separately. 
Such methods usually rely on the Fenchel-Rockafellar duality~\cite{Komodakis_N_2015_j-ieee-spm_pla_dor, Bauschke_H_2011_book_con_amo}.  On the one hand, some algorithms can evolve fully in the dual space, such as, e.g., ADMM \cite{Gabay1976, Fortin1983, Boyd2011} or the dual-Forward-Backward (FB) \cite{Combettes_P_2009_j-svva_dualization_srp, Combettes_Vu_2010}. Note however that ADMM requires the inversion of $\D^\top \D$. On the other hand, other algorithms can alternate between the primal and the dual spaces, namely primal-dual algorithms \cite{chambolle2011first, condat_primal-dual_2013, vu_splitting_2013, combettes2012primal}.

During the last decade, the performances of proximal algorithms have been pushed to the next level by mixing them with deep learning approaches \cite{venkatakrishnan2013plug, lecun2015deep,ongie2020deep} 
leading to {\UNN}s that consists in unrolling optimisation algorithms over a fixed number of iterations \cite{adler2018learned, Jiu_M_2021_jstsp_deep_pdp}. In the litterature, two classes of {\UNN}s can be encountered: \UNN-LO (i.e. {\UNN}  with learned Linear Operators) where the involved linear operators  are learned  or the \UNN-PO (i.e. {\UNN} with learned Proximal Operators)  where  the proximity operators are replaced by small neural networks, typically a Unet \cite{aggarwal2018modl}. 
{\UNN}s  have shown to be very efficient for denoising task \cite{Le_HTV_2022_p-eusipco-fas-pab} and further good performances for image restoration problems including deconvolution \cite{Jiu_M_2021_jstsp_deep_pdp,jiu2022alternative}, magnetic resonance imaging \cite{adler2018learned}, or computed tomography \cite{savanier2023deep}. 
Note that in the context of \UNN-PO, if only the denoiser is learned, and the algorithm is unrolled until convergence, then the resulting approach boils down to plug-and-play (PnP) or deep equilibrium \cite{bai2019deep, gilton2021deep, zou2023deep}.

\smallskip

%
%
%

\noindent \textbf{Contributions --}  We propose a unified framework to build unfolded {\UNN}-LO Gaussian denoisers. These are designed by unrolling proximal algorithms to obtain a MAP estimate of a Gaussian denoising problem of the form of~\eqref{prob:2-terms-minimization}-\eqref{eq:prox-function}, that is equivalent to computing a proximity operator. More specifically,
\begin{itemize}
\item  We propose a unified framework to build {\UNN} Gaussian denoisers. 
In this context, we design a primal-dual Arrow-Hurwicz building block for designing NN layers (allowing for skip connections), embedding in particular two proximal algorithms adapted to solve~\eqref{prob:2-terms-minimization}-\eqref{eq:prox-function}: the dual-FB (DFB) iterations and the primal-dual Chambolle-Pock (CP) iterations. We also show that accelerated versions of DFB and CP fit within the proposed primal-dual Arrow-Hurwicz building block.


\item Within the unified framework, we investigate two learning strategies for the learnable parameters of the proposed {\UNN}s, namely Learned Normalized Operators (LNO) strategy satisfying theoretical conditions dictated by optimization theory; and Learned Flexible Operator (LFO) strategy enabling more flexible learning strategies. 
 In particular,
for all the considered {\UNN}s, we show that the LNO strategy leads to more robust {\UNN}s with better denoising performances than the LFO strategy. This means that imposing some normalizing constraints on the network linearities helps to design a better architecture, hence showing that optimization theory can be useful in practice to design powerful (denoising) networks.


\item 
We show numerically that the proposed denoising {\UNN}s are highly competitive with state-of-the-art denoising networks. In particular, we show that we rich comparable performances as DRUnet, with potentially higher robustness, for a much lower computational cost (with  $10^3$ times less learnable parameters). 
Robustness properties of our {\UNN}s are investigated by computing their Lipchitz constants, and by studying their performances when applied in a different denoising setting than the Gaussian denoising task for which they have been trained (i.e., for non-Gaussian noise, and when plugged within a plug-and-play restoration algorithm). 
\end{itemize}

\smallskip

\noindent \textbf{Outline --} {The remainder of this paper is organized as follows. Section~\ref{s:denoising}  focuses on MAP denoising estimates, with a recall of the considered iterative schemes. 
Section~\ref{Sec:proposed-unfolded} is dedicated to the design of the proposed unified {\UNN} architectures, relying on the algorithmic schemes presented in Section~\ref{s:denoising}. 
We evaluate the performances of the proposed {\UNN}s in Section~\ref{s:exp}. In particular, we compare the denoising performances of our {\UNN}s to those of an established denoising network in terms of both reconstruction quality and robustness. We further present results when the proposed {\UNN}s are used in a PnP framework for an image deblurring task.}
Finally, our conclusions are given in Section~\ref{s:conclusion}.

\medskip

\noindent \textbf{Notation -- } In the remainder of this paper we will use the following notations. 
An element of $\RR^N$ is denoted by $\mathrm{x}$. For every $n \in \{1, \ldots, N\}$, the $n$-th coefficient of $\mathrm{x}$ is denoted by $\mathrm{x}^{(n)}$. {The spectral norm is denoted $\Vert \cdot \Vert_S$.} 
Let $C \subset \RR^N$ be a closed, non-empty, convex set. The indicator function of $C$ is denoted by $\iota_C$, and is equal to $0$ if its argument belongs to $C$, and $+\infty$ otherwise. Let $\mathrm{x} \in \RR^N$. The Euclidean projection of $\mathrm{x}$ onto $C$ is denoted by $\proj_C(\mathrm{x}) = \argmin{\mathrm{v} \in C} \| \mathrm{v}-\mathrm{x} \|^2$.
Let $\psi \colon \RR^N \to (-\infty, +\infty]$ be a convex, lower semicontinuous, proper function. The proximity operator of $\psi$ at $\mathrm{x}$ is given by $\prox_\psi(\mathrm{x}) = \argmin{\mathrm{v} \in \RR^N} \psi(\mathrm{v}) + \frac12 \| \mathrm{v} - \mathrm{x} \|^2 $. 
The Fenchel-Legendre conjugate function of $\psi$ is given by $\psi^*(\mathrm{x}) = \sup_{\mathrm{v} \in \RR^N} \mathrm{v}^\top \mathrm{x} - \psi(\mathrm{v})$. 
When ${\psi = \lambda \Vert\cdot \Vert_1}$ is the $\ell_1$ norm, then $\psi^* = \iota_{\mathcal{B}_{\infty}(0, \lambda )}$ corresponds to the indicator function of the $\ell_\infty$-ball centred in $0$ with radius $\lambda>0 $, i.e., $\mathcal{B}_{\infty}(0, \lambda ) = \{ \mathrm{x} \in \RR^N \mid (\forall n \in \{1, \ldots, N\}) \; -\lambda  \le \mathrm{x}^{(n)} \le \lambda \}$.

Let $f_\Theta^K$ be a feedforward NN, with $K$ layers and learnable parameters $\Theta$. It can be written as a composition of operators (i.e., layers)
$f_\Theta = \mathrm{L}_{\Theta_K}\circ\dots\circ \mathrm{L}_{\Theta_1}$, where, for every $k\in \{1, \ldots, K\}$, $\Theta_k$ are the learnable parameters of the $k$-th layer $\mathrm{L}_{\Theta_k}$. The $k$-th layer is defined as  
\begin{equation}    \label{def:layer-ff}
    \mathrm{L}_{\Theta_k}\colon \mathrm{u} \in \RR^{N_k} \mapsto \eta_k(\mathrm{W}_k \mathrm{u}+ \mathrm{b}_k),
\end{equation}
where $\eta_k$ is an (non-linear) activation function, $\mathrm{W}_k $ is a linear operator, and $\mathrm{b}_k$ is a bias. 
In \cite{1903.01014}, authors showed that most of usual activation function involved in NNs correspond to proximal operators.

\section{Denoising (accelerated) proximal schemes}
\label{s:denoising}

In this section, we introduce the two algorithmic schemes we will focus in this work, i.e., FB and CP. 
Although problem~\eqref{prob:2-terms-minimization}-\eqref{eq:prox-function} does not have a closed form solution in general, iterative methods can be used to approximate it. 
Multiple proximal algorithms can be used to solve~\eqref{prob:2-terms-minimization} as detailed in the introduction section. 
In this section, we describe two schemes enabling minimizing~\eqref{eq:prox-function}: the FB algorithm applied to the dual problem of~\eqref{prob:2-terms-minimization}, and the primal-dual CP algorithm directly applied to~\eqref{prob:2-terms-minimization}. In addition, for both schemes we also investigate their accelerated versions, namely DiFB (i.e., Dual inertial Forward Backward, also known as FISTA \cite{beck2009fast, chambolle2015convergence}), and ScCP (i.e., CP for strongly convex functions \cite{chambolle2011first}). For sake of simplicity, in the remainder of the paper we refer to these schemes as FB, CP, DiFB, and ScCP, respectively.

\smallskip

\noindent \textbf{D(i)FB} -- 
A first strategy to solve \eqref{prob:2-terms-minimization}-\eqref{eq:prox-function} consists in applying (i)FB to the dual formulation of problem~\eqref{prob:2-terms-minimization} (see \cite[Problem~3.2]{Combettes_Vu_2010} for details), whose associated iterations read:
\begin{equation}
\label{eq:fista}
\begin{array}{l}
    \text{for } k = 0, 1, \ldots \\
    \left\lfloor 
    \begin{array}{l}  
    	\dualvar_{k+1}= \prox_{\tau_k (\nu g)^*}\Big(\mathrm{v}_{k}+\tau_k \D \proj_{C}(\datad -\D^\top\mathrm{v}_{k})\Big),\\
    	\mathrm{v}_{k+1}= (1+\rho_k) \dualvar_{k+1}-\rho_k\dualvar_{k} ,
	\end{array}
    \right. 
\end{array}
\end{equation}
where {$(\dualvar_0, \mathrm{v}_0)\in \mathbb{R}^{\mid\mathbb{F}\mid} \times \mathbb{R}^{\mid\mathbb{F}\mid}$} and, for every $k \in \NN$, $\tau_k>0$ and $\rho_k\geq 0$ are step-size parameters. Note that when, for every $k\in \NN$, $\rho_k = 0$, then algorithm~\eqref{eq:fista} reduces to DFB and when $\rho_k > 0$ it leads to DiFB formalism.

The following convergence result applies.
\begin{theorem}[\hspace{-0.25pt}\cite{Combettes_Vu_2010, chambolle2015convergence}] \label{thm:cvg-fista}
Let $(\dualvar_k, \mathrm{v}_k)_{k\in \NN}$ be generated by \eqref{eq:fista}.
Assume that one of the following conditions is satisfied.
\begin{enumerate}
    \item 
    For every $k\in \NN$, $\tau_k \in (0, 2/ \| \D \|_S^2)$, and $\rho_k = 0$.
    \item 
    For every $k\in \NN$, $\tau_k \in (0, 1/ \| \D \|_S^2)$, and $\rho_k=\frac{t_k-1}{t_{k+1}}$ with $t_k=\frac{k+a-1}{a}$ and $a>2$.
\end{enumerate}
Then we have
{\begin{equation}
    \widehat{\estmsd}_{\mathrm{MAP}} = \lim_{k \to \infty} \proj_C ( \datad -\D^\top\dualvar_{k} ),
\end{equation}}
where $\widehat{\estmsd}_{\mathrm{MAP}}$ is defined in {\eqref{prob:2-terms-minimization}-\eqref{eq:prox-function}}.
\end{theorem}

\smallskip

\noindent \textbf{(Sc)CP} -- 
A second strategy to solve \eqref{prob:2-terms-minimization}-\eqref{eq:prox-function} consists in applying the (Sc)CP algorithm to problem \eqref{prob:2-terms-minimization}. 
The data-term being $\zeta$-strongly convex with parameter $\zeta=1$, the accelerated CP \cite{chambolle2011first}, dubbed ScCP, can be employed. As described in Appendix~\ref{s:annex}, the associated iterations can be reformulated as:
\begin{align}
\label{eq:sccp}
\begin{array}{l}
    \text{for } k = 0, 1, \ldots \\
    \left\lfloor 
    \begin{array}{l}  
    \estmsd_{k+1} 
    =   \proj_C \left(\frac{\mu_k}{1+\mu_k} ( \datad - \D^\top\dualvar_k ) + \frac{1}{1+\mu_k} \estmsd_k \right), \\
    \dualvar_{k+1} =\prox_{\tau_k (\nu g)^*} \left( \dualvar_k+\tau_k \D \Big( (1+\alpha_k ) \estmsd_{k+1}-\alpha_{k}\estmsd_{k} \Big) \right), 
	\end{array}
    \right. 
\end{array}
\end{align}
where $\estmsd_0 \in \RR^N$ and $\dualvar_0 \in \mathbb{R}^{\mid\mathbb{F}\mid} $. 
Note that when, for every $k \in \NN$, $\alpha_k = 1$, then algorithm~\eqref{eq:sccp} reduces to standard iterations of  the primal-dual CP algorithm \cite{chambolle2011first}. When $\alpha_k=(1+2\zeta\mu_k)^{-1/2}$ it leads to ScCP and when $\alpha_k = 0$, it leads to the classical Arrow-Hurwicz algorithm \cite{Arrow_K_1958_book_s_lnl}.

The following convergence result applies.
\begin{theorem}[\hspace{-0.25pt}\cite{chambolle2011first}] \label{thm:cvg-sccp}
Let $(\dualvar_k, \mathrm{x}_k)_{k\in \NN}$ be generated by \eqref{eq:sccp}.
Assume that $(\tau_k)_{k\in \NN}$ and $(\mu_k)_{k\in \NN}$ are positive sequences, and that one of the following conditions is satisfied.
\begin{enumerate}
    \item 
    For every $k\in \NN$, $\tau_k \mu_k \| \D \|_S^2 < 1$, and $\alpha_k = 1$.
    \item 
    For every $k\in \NN$, {$\alpha_k=(1+2\mu_k)^{-1/2}$}, $	\mu_{k+1}=\alpha_{k}\mu_k$, and $\tau_{k+1}= \tau_k \alpha_{k}^{-1}$ with $\mu_0 \tau_0\Vert \D \Vert_S^2\leq 1$.
\end{enumerate}
Then we have
\begin{equation}
    \widehat{\estmsd}_{\mathrm{MAP}} = \lim_{k \to \infty} \mathrm{x}_{k},
\end{equation}
where $\widehat{\estmsd}_{\mathrm{MAP}}$ is defined in {\eqref{prob:2-terms-minimization}-\eqref{eq:prox-function}}.
\end{theorem}


\section{Proposed unfolded denoising NNs}
\label{Sec:proposed-unfolded}

In this section we aim to design {\UNN}s $f_\Theta$ such that 
\begin{equation}    \label{eq:NN-estim}
    f_\Theta(\mathrm{z}) \approx \widehat{\estmsd},
\end{equation}
where $\widehat{\estmsd} \in \RR^N$ is an estimate of $\overline{\estmsd}$. 
As discussed in the introduction, such an estimate can correspond to the penalized least-squares MAP estimate of $\overline{\estmsd}$, defined as in \eqref{prob:2-terms-minimization}-\eqref{eq:prox-function}.

\subsection{Primal-dual building block iteration}

The iterations described previously in \eqref{eq:fista} and \eqref{eq:sccp} share a similar framework which yields:
\begin{align}
\label{eq:sccpfista}
\begin{array}{l}
    \text{for } k = 0, 1, \ldots \\
    \left\lfloor 
    \begin{array}{l}  
         \dualvar_{k+1} =\prox_{\tau_k (\nu g)^*} \Big(\dualvar_k+\tau_k \D \estmsd_k\Big) \\
    \estmsd_{k+1} =   \proj_C \left(\frac{\mu_k}{1+\mu_k} (\datad -\D^\top\dualvar_{k+1}) +\frac{1}{1+\mu_k}\estmsd_{k}\right).
	\end{array}
    \right. 
\end{array}
\end{align}
On the one hand, this scheme is a reformulation of the Arrow-Hurwicz (AH) iterations, i.e., algorithm~\eqref{eq:sccp} with $\alpha_k \equiv 0$. On the other hand, for the limit case when $\mu_k \to +\infty$, the DFB~\eqref{eq:fista} iterations are recovered. 
Further, the inertia step is activated either on the dual variable for DiFB 
\begin{equation}    \label{eq:fista-inertia}
\dualvar_{k+1} \leftarrow (1+\rho_{k}) \dualvar_{k+1}-\rho_k\dualvar_{k}
\end{equation}
or on the primal variable for {ScCP} 
\begin{equation}    \label{eq:sccp-inertia}
\estmsd_{k+1} \leftarrow(1+\alpha_{k}) \estmsd_{k+1}-\alpha_k\estmsd_{k}.
\end{equation}

Based on these observations we propose a strategy to build denoising {\UNN}s satisfying~\eqref{eq:NN-estim}, reminiscent to our previous works \cite{Le_HTV_2022_p-eusipco-fas-pab, Repetti_A_2022_eusipco_dual_fbu}, to unroll D(i)FB and (Sc)CP algorithms using a primal-dual perspective.

The result provided below aims to emphasize that each iteration of the primal-building block~\eqref{eq:sccpfista} can be viewed as the composition of two layers of feedforward networks, acting either on the image domain (i.e., primal domain $\RR^N$) or in the features domain (i.e., dual domain $\RR^{| \mathbb{F} |}$). {DFB}, DiFB, CP, and ScCP, hold the same structure, with extra steps that can be assimilated to skip connections {in the specific case of DiFB and ScCP}, enabling to keep track of previous layer's outputs (see \eqref{eq:fista-inertia} and \eqref{eq:sccp-inertia}).

\begin{proposition}\label{prop:unfolded_a}
Let $\datad \in \RR^N$, $\estmsd \in\mathbb{R}^{N}$, $\dualvar \in  \mathbb{R}^{| \mathbb{F} |}$, and $k\in \NN$.

\noindent
Let $\mathrm{L}_{ \nu, \Theta_{k, \mathcal{D}}, \mathcal{D}}\colon \mathbb{R}^{N} \!\! \times \! \mathbb{R}^{| \mathbb{F} |} \to \mathbb{R}^{| \mathbb{F} |}$, defined as
\begin{equation}    \label{prop:unfold:sublayD}
\mathrm{L}_{ \nu, \Theta_{k, \mathcal{D}}, \mathcal{D}} (\estmsd,\dualvar) = \eta_{ \nu,k,\mathcal{D}}\left(\Wrm_{k,\mathcal{D}} \estmsd + \Vrm_{k,\mathcal{D}}  \dualvar+\brm_{k,\mathcal{D}}\right),
\end{equation}
be a sub-layer acting on both the primal variable $\estmsd$ and the dual variable $\dualvar$ and returning a dual ($\mathcal{D}$) variable. In~\eqref{prop:unfold:sublayD} $\eta_{ \nu,k,\mathcal{D}} \colon \mathbb{R}^{| \mathbb{F} |} \to \mathbb{R}^{| \mathbb{F} |}$ is a fixed activation function with parameter $\nu>0$, and $\Theta_{k, \mathcal{D}}$ is a linear parametrization of the learnable parameters including $\Wrm_{k,\mathcal{D}} \colon \RR^N \to \mathbb{R}^{| \mathbb{F} |}$, $\Vrm_{k,\mathcal{D}} \colon \RR^{| \mathbb{F} |} \to \mathbb{R}^{| \mathbb{F} |}$ and $\brm_{k,\mathcal{D}} \in \mathbb{R}^{| \mathbb{F} |}$.

\noindent
Let $\mathrm{L}_{\datad, \Theta_{k, \mathcal{P}},\mathcal{P}} \colon \mathbb{R}^{N} \times \mathbb{R}^{| \mathbb{F} |}  \to  \mathbb{R}^{N}$, defined as
\begin{equation}    \label{prop:unfold:sublayP}
\mathrm{L}_{\datad, \Theta_{k, \mathcal{P}},\mathcal{P}} (\estmsd,\dualvar)  =  \eta_{k,\mathcal{P}}\left(\Wrm_{k,\mathcal{P}} \estmsd + \Vrm_{k,\mathcal{P}} \dualvar +\brm_{k,\mathcal{P}}\right),
\end{equation}
be a sub-layer acting on both the primal variable $\estmsd$ and the dual variable $\dualvar$ and returning a primal ($\mathcal{P}$) variable. In~\eqref{prop:unfold:sublayP} $\eta_{k,\mathcal{P}} \colon \RR^N \to \RR^N$ is a fixed activation functions, 
and $\Theta_{k, \mathcal{P}}$ is a linear parametrization of  the learnable parameters including $\Wrm_{k,\mathcal{P}} \colon \mathbb{R}^{N} \to \RR^N$, $\Vrm_{k,\mathcal{P}} \colon \mathbb{R}^{| \mathbb{F} |} \to \RR^N$ and $\brm_{ k,\mathcal{P}} \in \RR^N$. 

Then, the $k$-th iteration of the joint formulation \eqref{eq:sccpfista} can be written as a composition of two layers of the form of~\eqref{def:layer-ff}:
\begin{equation}    \label{prop:unfold:lay-ISTA}
{\begin{array}{lccl}
    \mathrm{L}_{\datad, \Theta_k}\colon 
    & \mathbb{R}^{N} \times \mathbb{R}^{| \mathbb{F} |} & \to & \mathbb{R}^{N} \\
    & (\estmsd_k ,\dualvar_k )   & \mapsto & \mathrm{L}_{\datad,\Theta_{k,\mathcal{P}},\mathcal{P}}( \estmsd,\mathrm{L}_{ \Theta_{k,\mathcal{D}},\mathcal{D}} ( \estmsd_k,\dualvar_k)),
\end{array}}
\end{equation}
where $\Theta_k$ is the combination of $\Theta_{k,\mathcal{P}}$ and $\Theta_{k,\mathcal{D}}$, i.e., the linear parametrization of all learnable parameters for layer $k$.

\end{proposition}

\begin{proof}
This result is obtained by noticing that, for every $k\in \NN$, the $k$-th iteration in algorithm~\eqref{eq:sccpfista} can be rewritten as \eqref{prop:unfold:lay-ISTA}, where the primal ($\mathcal{P}$) operators are given by 
$\Wrm_{k,\mathcal{P}} = \frac{1}{1+\mu_k}$,
$\Vrm_{k,\mathcal{P}} = -\frac{\mu_k}{1+\mu_k}\D^\top$,
$\brm_{ k,\mathcal{P}} = \frac{\mu_k}{1+\mu_k}\datad$, 
and $\eta_{k,\mathcal{P}} = \proj_C$, 
and the dual ($\mathcal{D}$) operators are given by
$\Wrm_{k,\mathcal{D}} = \tau_k\D$,
$\Vrm_{k,\mathcal{D}} = \Id$,  
$\brm_{k,\mathcal{D}} = 0$,
$\eta_{ \nu,k,\mathcal{D}} = \prox_{\tau_k (\nu g)^*}$.
\end{proof}

\begin{figure}[!]
\centering
\includegraphics[width=0.6\columnwidth]{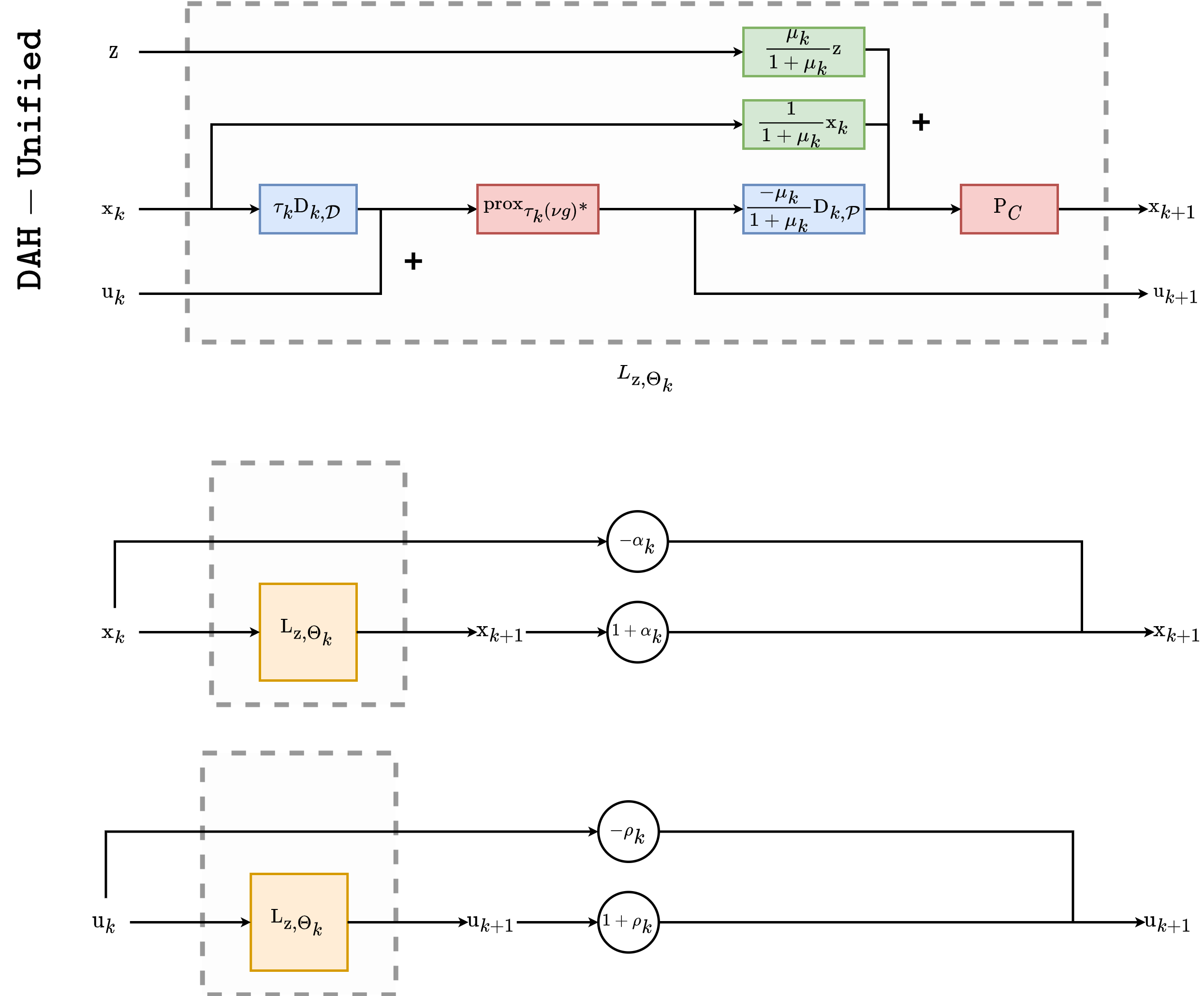}
\caption{Top: Architecture of the proposed {\tt DAH-Unified} block for the $k$-th layer. Linearities, biases, and activation functions are shown in blue, green and red, respectively.  
Bottom: {\tt Inertial step} for ScCP (top) and DiFB (bottom), for the $k$-th layer.}
\label{fig:unfoldv2}
\end{figure}

\subsection{Arrow-Hurwicz unfolded building block}
\label{ss:ahbb}
Our unrolled architectures rely on layer structures introduced in Proposition~\ref{prop:unfolded_a} where we allow the linear operator $\D$ to be different for each layer. For more flexibility, we also introduce, for every $k\in \{1, \ldots, K\}$, operators $\D_{k, \mathcal{D}} \colon \RR^N \to \mathbb{R}^{| \mathbb{F} |}$ and $\D_{k, \mathcal{P}} \colon \mathbb{R}^{| \mathbb{F} |} \to \RR^N $, to replace operators $ \D$ and $ \D^\top$, respectively, to allow a possible mismatch between operator $\D$ and its adjoint $\D^\top$. 
The resulting unfolding Deep Arrow-Hurwicz (DAH) building block is then given below:

    \begin{equation}    \label{nn:ah}
    {f}_{\datad, \Theta}^{K, \mathrm{DAH}}(\estmsd_0,\dualvar_0) = \Lrm^{\mathrm{DAH}}_{\datad, \nu, \Theta_K} \circ \cdots \circ \Lrm^{\mathrm{DAH}}_{\datad, \nu, \Theta_1} (\estmsd_0,\dualvar_0),
    \end{equation}
    where, for every $k\in\{1,\ldots,K\}$,
    \begin{align*}  
    &\dualvar_k = \mathrm{L}_{ \nu, \Theta_{k,\mathcal{D}},\mathcal{D}} ( \estmsd_{k-1},\dualvar_{k-1})   \\
    &\estmsd_k = \mathrm{L}_{\datad,\Theta_{k,\mathcal{P}},\mathcal{P}} ( \estmsd_{k-1} , \dualvar_k ) \\
    &\mathrm{L}^{\mathrm{DAH}}_{\datad, \nu, \Theta_k} (\estmsd_{k-1} ,\dualvar_{k-1} ) = (\estmsd_k, \dualvar_k).
    \end{align*}
    with
    \begin{equation*}
    \begin{cases}
    \Wrm_{k,\mathcal{D}} = \tau_k\D_{k, \mathcal{D}},\\
    \Vrm_{k,\mathcal{D}} = \Id,  \\
    \brm_{k,\mathcal{D}} = 0,\\
    \eta_{ \nu,k,\mathcal{D}} = \prox_{\tau_k (\nu g)^*},
    \end{cases}
    \text{ and }
    \begin{cases}
    \Wrm_{k,\mathcal{P}} = \frac{1}{1+\mu_k},\\
    \Vrm_{k,\mathcal{P}} = -\frac{\mu_k}{1+\mu_k}\D_{k, \mathcal{P}},\\
    \brm_{ k,\mathcal{P}} = \frac{\mu_k}{1+\mu_k}\datad, \\
    \eta_{k,\mathcal{P}} = \proj_C.    
    \end{cases}
    \end{equation*}

\subsection{Proposed unfolded strategies}
\label{Ssec:unfolded-str}

In this section we describe four unfolded strategies for building denoising NNs as defined in~\eqref{eq:NN-estim}. All strategies rely on Arrow-Hurwicz building block presented in Section~\ref{ss:ahbb}. 
\begin{itemize}
    \item 
 \textbf{DDFB} stands for Deep Dual Forward-Backward and it fits DAH when $\mu_k\to +\infty$. 
    \begin{equation}    \label{nn:DDFB}
    {f}_{\datad, \nu, \Theta}^{K, \mathrm{DDFB}}(\estmsd_0,\dualvar_0) = \Lrm^{\mathrm{DDFB}}_{\datad, \nu, \Theta_K} \circ \cdots \circ \Lrm^{\mathrm{DDFB}}_{\datad, \nu, \Theta_1} (\estmsd_0,\dualvar_0),
    \end{equation}
    where, for every $k\in\{1,\ldots,K\}$,
    \begin{align*}    
    &\dualvar_k = \mathrm{L}_{ \nu, \Theta_{k,\mathcal{D}},\mathcal{D}} ( \estmsd_{k-1},\dualvar_{k-1})   \\
    &\estmsd_k = \mathrm{L}_{\datad,\Theta_{k,\mathcal{P}},\mathcal{P}} ( \estmsd_{k-1} , \dualvar_k ) \\
    &(\estmsd_k, \dualvar_k) = \mathrm{L}^{\mathrm{DDFB}}_{\datad, \nu, \Theta_k} (\estmsd_{k-1} ,\dualvar_{k-1} ) .
    \end{align*}
    with
    \begin{equation*}
    \begin{cases}
    \Wrm_{k,\mathcal{D}} = \tau_k\D_{k, \mathcal{D}},\\
    \Vrm_{k,\mathcal{D}} = \Id,  \\
    \brm_{ k,\mathcal{D}} = 0,\\
    \eta_{ \nu,k,\mathcal{D}} = \prox_{\tau_k (\nu g)^*}  , 
    \end{cases}
    \text{ and }
    \begin{cases}
    \Wrm_{k,\mathcal{P}} = 0,\\
    \Vrm_{k,\mathcal{P}} = -\D_{k, \mathcal{P}},\\
    \brm_{\datad, k,\mathcal{P}} = \datad, \\
    \eta_{k,\mathcal{P}} = \proj_C.    
    \end{cases}
    \end{equation*}

    \item 
    \textbf{DDiFB:}  stands for Deep Dual inertial Forward-Backward interpreted as a DDFB with skip connections and defined as: 
    \begin{equation}    \label{nn:DDiFB}
   {f}_{\datad, \nu, \Theta}^{K, \mathrm{DDiFB}}(\estmsd_0,\dualvar_0)  = (\estmsd_K, \dualvar_K)
    \end{equation}
    where, for every $k\in\{1,\ldots,K\}$,
    \begin{align*}    
    (\estmsd_k ,\widetilde{\dualvar}_k ) &= 
    \mathrm{L}^{\mathrm{DDFB}}_{\datad, \nu, \Theta_k}(\estmsd_{k-1} ,\dualvar_{k-1} )\\
     \dualvar_k &= (1+\rho_k) \widetilde{\dualvar}_k-\rho_k \dualvar_{k-1}.
    \end{align*}


     \item 
    \textbf{DCP} stands for Deep Chambolle-Pock relying on DAH with a special update of the primal variable leading to: 
    \begin{equation}    \label{nn:dcp}
   {f}_{\datad, \nu, \Theta}^{K, \mathrm{DCP}}(\estmsd_0,\dualvar_0)  = (\estmsd_K, {\dualvar}_K)
    \end{equation}
    where, for every $k\in\{1,\ldots,K\}$,
    \begin{align*}    
    (\widetilde{\estmsd}_k ,\dualvar_k ) 
    &= 
    \mathrm{L}^{\mathrm{DAH}}_{\datad, \nu, \Theta_k}(\estmsd_{k-1} ,\dualvar_{k-1} )\\
     {\estmsd}_k 
     &= 2\widetilde{\estmsd}_k-\estmsd_{k-1}.
    \end{align*}

    \item 
    \textbf{DScCP}  stands for Deep Strongly convex Chambolle-Pock interpreted as a DAH with skip connections on the primal variable: 
    \begin{equation}    \label{nn:dsccp}
   {f}_{\datad, \nu, \Theta}^{K, \mathrm{DScCP}}(\estmsd_0,\dualvar_0)  = (\estmsd_K, {\dualvar}_K)
    \end{equation}
    where, for every $k\in\{1,\ldots,K\}$,
    \begin{align*}    
    (\widetilde{\estmsd}_k ,\dualvar_k ) 
    &= \mathrm{L}^{\mathrm{DAH}}_{\datad, \nu, \Theta_k}(\estmsd_{k-1} ,\dualvar_{k-1} )\\
     {\estmsd}_k 
     &= (1+\alpha_{k}) \widetilde{\estmsd}_k-\alpha_k\estmsd_{k-1}.
    \end{align*}
\end{itemize}
Illustration of a single layer of the resulting DD(i)FB and D(Sc)CP architectures are provided in Figure~\ref{fig:unfoldv2}.

Since these architectures are reminiscent of D(i)FB and (Sc)CP, given in {Section~\ref{s:denoising}}, we can deduce limit cases for the proposed unfolded strategies, when $K\to +\infty$ and linear operators are fixed over the layers. 

\begin{corollary}[Limit case for deep unfolded NNs] \label{cor:limit-NN}
We consider the unfolded NNs DD(i)FB and D(Sc)CP defined in Section~\ref{Ssec:unfolded-str}.
Assume that, for every $k\in \{1, \ldots, K\}$, $\D_{k, \mathcal{D}} = \D$ and $\D_{k, \mathcal{P}} = \D^\top$, for $\D \colon \RR^N \to \RR^{| \mathbb{F} |} $. 
In addition, for each architecture, we further assume that, for every $k\in \{1, \ldots, K\}$, 
\begin{itemize}
    \item \emph{{DDFB:}} $\tau_k \in (0, 2/ \|\D\|_S^2)$.
    \item \emph{{DDiFB:}} $\tau_k \in (0, 1/ \|\D\|_S^2)$ and $\rho_k = \frac{t_k-1}{t_{k+1}}$ with $t_k = \frac{k+a-1}{a}$ and $a>2$.
    \item \emph{{DCP:}} $(\tau_k, \mu_k)\in (0, +\infty)^2$ such that $\tau_k \mu_k  \|\D\|_S^2 < 1$.
    \item \emph{{DScCP:}} {$\alpha_k = (1 + 2 \mu_k)^{-1/2}$}, $\mu_{k+1} = \alpha_k \mu_k$,  and  $\tau_{k+1} = \tau_k\alpha_k^{-1}$ with $\tau_0 \mu_0  \|\D\|_S^2 \le 1$.
\end{itemize}
Then, we have $\estmsd_K \to \widehat{\estmsd}$ when $K\to +\infty$, where $\estmsd_K$ is the output of either of the unfolded NNs DD(i)FB or D(Sc)CP, and $\widehat{\estmsd}$ is a solution to \eqref{prob:2-terms-minimization}
\end{corollary}

\subsection{Proposed learning strategies}

The proposed DD(i)FB and D(Sc)CP allow for flexibility in the learned parameters. In this work, we propose two learning strategies, either satisfying conditions described in Corollary~\ref{cor:limit-NN} (LNO), or giving flexibility to the parameters (LFO). These two strategies are described below. 
\begin{enumerate}
    \item
    \textbf{Learned Normalized Operators (LNO):} 
    This strategy uses theoretical conditions ensuring convergence of D(i)FB and (Sc)CP, i.e., choosing the stepsizes appearing in deep-D(i)FB and deep-(Sc)CP according to the conditions given in Corollary~\ref{cor:limit-NN}. 
    In this context, for every $k\in \{1, \ldots,K\}$, we choose $\D_{k, \mathcal{P}} $ to be equal to the adjoint $\D_{k, \mathcal{D}}^\top$ of $\D_{k, \mathcal{D}}$. However, unlike in Corollary~\ref{cor:limit-NN}, we allow $\D_{k, \mathcal{D}}$ to vary for the different layers $k \in \{1, \ldots, K\}$.
    \item 
    \textbf{Learned Flexible Operator (LFO):} 
    This strategy, introduced in \cite{Le_HTV_2022_p-eusipco-fas-pab}, consists in learning the stepsizes appearing in deep-D(i)FB and deep-(Sc)CP without constraints, as well as allowing a mismatch in learning the adjoint operator $\D^\top$ of $\D$, i.e., learning $\D_{k, \mathcal{D}}$ and $\D_{k, \mathcal{P}}$ independently.
\end{enumerate}
The learnable parameters are summarized in Table~\ref{tab:learnable-param}.

\begin{table}[t!]
    \centering
    \caption{Learnable paramaters of each {\UNN} scheme}
    \label{tab:learnable-param}
    
    \footnotesize
    \begin{tabular}{@{}l|l|l@{}}
        &  $\Theta_{k}$         & Comments\\
        \hline\hline
     DDFB-LFO  &  $\D_{k, \mathcal{P}}$, $\D_{k, \mathcal{D}}$           
        & absorb $\tau_{k}$ in $\D_{k, \mathcal{D}}$\\[0.2cm]
     DDiFB-LFO &  $\D_{k, \mathcal{P}}$, $\D_{k, \mathcal{D}}$, $\alpha_k$  
        & fix $\alpha_k$, and absorb $\tau_{k}$ in $\D_{k, \mathcal{D}}$\\[0.2cm]
     DDFB-LNO   &  $\D_{k, \mathcal{P}}= \D_{k, \mathcal{D}}^\top$           
        & define $\tau_k = 1.99\|\D_k\|^{-2}$\\[0.2cm]
     DDiFB-LNO  & $\D_{k, \mathcal{P}}= \D_{k, \mathcal{D}}^\top$  
        & fix $\alpha_k=\frac{t_k-1}{t_{k+1}}$, $t_{k+1}=\frac{k+a-1}{a}$, $a>2$, and $\tau_k = 0.99\|\D_k\|^{-2}$\\
     \hline\hline
      DCP-LFO    &  $\D_{k, \mathcal{P}}$, $\D_{k, \mathcal{D}}$, $\mu $
        & learn $\mu=\mu_0=\dots=\mu_K$, and absorb $\tau_{k}$ in $\D_{k, \mathcal{D}}$\\[0.2cm]
     DScCP-LFO  &  $\D_{k, \mathcal{P}}$, $\D_{k, \mathcal{D}}$, $\mu_0 $
        &  learn $\mu_0$, absorb $\tau_{k}$ in $\D_{k, \mathcal{D}}$, and fix $\alpha_k=(1+2\mu_k)^{-1/2}$, and $\mu_{k+1}=\alpha_{k}\mu_k$\\[0.2cm]
     DCP-LNO     &  $\D_{k, \mathcal{P}}= \D_{k, \mathcal{D}}^\top$, $\mu$
        & learn $\mu=\mu_0=\dots=\mu_K$, and fix $\tau_k = 0.99\mu^{-1} \|\D_k\|^{-2}$\\[0.2cm]
     DScCP-LNO   &  $\D_{k, \mathcal{P}}= \D_{k, \mathcal{D}}^\top$, $ \mu_k$
        &learn $\mu_k$, and fix $\alpha_k=(1+2\mu_k)^{-1/2}$, and $\tau_k = 0.99 \mu_k^{-1} \|\D_k\|^{-2}$\\
     \hline
\end{tabular}
\end{table}

\section{Experiments}
\label{s:exp}

In this section we present simulation results to illustrate the performance of the proposed unfolded network architectures. 
In Section~\ref{Ssec:exp:training} we describe the training setting adopted for our networks. In Section~\ref{Ssec:exp:architecture} we present an architecture comparison in terms of computational complexity between the proposed unfolded network, a state-of-the-art denoising network, namely DRUnet, and a handcrafted denoiser, namely BM3D. 
We then present in Section~\ref{Ssec:exp:perf-denoise} the denoising performances of the different methods. 
Finally, Sections~\ref{Ssec:exp:robust-lip} to~\ref{s:pnp} are dedicated
to a study of the robustness of the networks, looking at three metrics: Lipschitz constants (Section~\ref{Ssec:exp:robust-lip}),
application of the network for noise distributions different from the training setting (Section~\ref{Ssec:exp:perf-denoise-vs-robustness}), and
plugging the denoising networks into a plug-and-play algorithm (Section~\ref{s:pnp}).

\subsection{Denoising training setting}
\label{Ssec:exp:training}

\noindent \textbf{Training dataset} -- 
We consider two sets of images: the \emph{training set} $(\overline{\mathrm{x}}_s,\datad_s)_{s\in \mathbb{I}}$ of size $\vert \mathbb{I}\vert $ and the \emph{test set} $(\overline{\mathrm{x}}_s,\datad_s)_{s\in \mathbb{J}}$ of size $\vert \mathbb{J} \vert $. For both sets, each couple $(\overline{\mathrm{x}}_s, \datad_s)$ consists of a clean multichannel image $\overline{\mathrm{x}}_s$ of size $N_s = C \times \widetilde{N}_s$ (where $C$ denotes the number of channels, and $\widetilde{N}_s$ the number of pixels in each channel), and a noisy version of this image given by $ \datad_s = \overline{\mathrm{x}}_s + \varepsilon_s$ with $\varepsilon_s\sim \mathcal{N}(0,\delta^2 \mathrm{Id})$ for $\delta>0$.  

\smallskip

\noindent \textbf{Training strategy for unfolded denoising networks} -- 
The network parameters are optimized by minimizing the $\ell^2$ empirical loss between noisy and ground-truth images:
\begin{equation} \label{eq:newlossk_bis}
		\widehat{\Theta} \in \underset{\Theta}{\textrm{Argmin}}\;\frac{1}{ \vert \mathbb{I}\vert}\sum_{s\in \mathbb{I}} \mathcal{L}(\overline{\mathrm{x}}_s, \datad_s ; \Theta)
\end{equation}
where
\begin{equation*}
\mathcal{L}(\overline{\mathrm{x}}_s,\datad_s;\Theta):=  \frac12 \Vert \overline{\mathrm{x}}_s - f^K_{\mathrm{z}_s, \delta^2, \Theta} \left(\mathrm{z}_s,\D_{1,\mathcal{D}}(\mathrm{z}_s)\right) \Vert^2 ,
\end{equation*}
and $f^K_{\mathrm{z}_s, \delta^2, \Theta}$ is either of the unfolded networks described in Section~\ref{Ssec:unfolded-str}.
The loss~\eqref{eq:newlossk_bis} will be optimized in Pytorch with Adam algorithm \cite{kingma2014adam}. 
For the sake of simplicity, in the following we drop the indices in the notation of the network  $f^K_{\mathrm{z}_s, \delta^2, \Theta}$ and use the notation $f_{\Theta}$.

\smallskip

\noindent \textbf{Architectures} --  
We will compare the different architectures introduced in Section~\ref{Ssec:unfolded-str}, namely {DDFB, DDiFB, DDCP, and DScCP,} considering both LNO and LFO learning strategies (see Table~\ref{tab:learnable-param} for details). 
For each architecture and every layer $k\in \{1, \ldots, K\}$, the weight operator $\D_{k, \mathcal{D}}$ consists of $J$ convolution filters (features), mapping an image in $\RR^N$ to features in $\RR^{\vert \mathbb{F}\vert}$, with $\vert \mathbb{F}\vert = J \widetilde{N} $. For LFO strategies, weight operator $\D_{k, \mathcal{P}}$ consists of $J$ convolution filters mapping from $\RR^{\vert \mathbb{F}\vert}$ to $\RR^N$. All convolution filters considered in our work have the same  kernel size of $3\times3$. 

We evaluate the performance of the proposed models, varying the numbers of layer $K$ and the number of convolution filters $J$. 
In our experiments we consider $g  =\Vert \cdot \Vert_1$ leading to HardTanh activation function as in \cite{Le_HTV_2022_p-eusipco-fas-pab} and recalled below. 
\begin{proposition}
    The proximity operator of the conjugate of the $\ell_1$-norm scaled by parameter $ \nu>0$ is equivalent to the HardTanh activation function, i.e., for every $\mathrm{x} = (\mathrm{x}^{(n)})_{1\leq n \leq N}$:
    \begin{equation*}
        (\mathrm{p}^{(n)})_{1\leq n \leq N}
        = \prox_{(\nu\| \cdot \|_1)^*}(\mathrm{x}) = \mathrm{P}_{\Vert \cdot \Vert_\infty \leq \nu}(\mathrm{x}) 
        = \mathrm{HardTanh}_\nu(\mathrm{x})
    \end{equation*}
where\vspace{-0.3cm}
\begin{equation*}
\mathrm{p}^{(n)} = \begin{cases}
- \nu & \mbox{if} \quad \mathrm{x}^{(n)}<- \nu,\\
 \nu & \mbox{if} \quad \mathrm{x}^{(n)}> \nu,\\
\mathrm{x}^{(n)} & \mbox{otherwise}.
\end{cases}
\end{equation*}
\end{proposition}

\smallskip

\noindent \textbf{Experimental settings} -- To evaluate and compare the proposed unfolded architectures, we consider 2 training settings. In both cases, we consider RGB images (i.e., $C=3$).

\begin{itemize}
\item \noindent \textbf{Training Setting 1 -- Fixed noise level}: 
The {\UNN}s are trained with $\vert \mathbb{I} \vert = 200$ images extracted from BSDS500 dataset \cite{amfm_pami2011}, with a fixed noise level $\delta=0.08$. The learning rate for ADAM is set to $8\times10^{-5}$ (all other parameters set as default), we use batches of size of $10$ and patches of size $50\times50$ randomly selected. 

We train the proposed unfolded NNs considering multiple sizes of convolution filters $J\in \{8,16,32,64\}$ and for multiple numbers of layers $K\in \{5,10,15,20,25\}$.

\item \noindent \textbf{Training Setting 2 -- Variable noise level}: 
The NNs are trained using the test dataset of ImageNet \cite{ILSVRC15} ($\vert \mathbb{I} \vert = 5\times 10^4$), with learning rate for ADAM set to $1\times10^{-3}$ (all other parameters set as default), batch size of $200$, and patches of size $50\times50$ randomly selected. Further, we consider a variable noise level, i.e., images are degraded by a Gaussian noise with standard deviation $\delta_i$, for $i\in \mathbb{I}$, selected randomly with uniform distribution in~$[0,0.1]$.

All {\UNN}s are trained with $(J,K)= (64,20)$. 
\end{itemize}

In our experiments, we aim to compare the proposed {\UNN}s for three metrics: (i)~architecture complexity, (ii) robustness, and (iii) denoising performance. 
These metrics will also be provided for a state-of-the-art denoising network, namely DRUnet \cite{zhang2021plug}. For the sake of completeness, we will also compare the Gaussian denoising power of our {\UNN}s to the hand-crafted denoiser BM3D \cite{Dabov_2007_j-tip_image_den_bm3d}.

For both Settings 1 and 2, our test set $\mathbb{J}$ corresponds to randomly selected subsets of images from the BSDS500 validation set. The size of the test set $| \mathbb{J}|$ will vary depending on the experiments, hence will be specified for each case.

\subsection{Architecture comparison}
\label{Ssec:exp:architecture}

We first compare the proposed unfolded NNs (for both LNO and LFO learning strategies) in terms of runtime, FLOPs and number of learnable parameters (i.e., $|\Theta|$). 
These values, for $(K,J)= (20, 64)$ are summarized in Table~\ref{tab:runtime+flop}, also including the metrics for DRUnet. 
The experiments are conducted in PyTorch, using an Nvidia Tesla V100 PCIe 16GB.
From the table, it is obvious that the unfolded NNs have much lighter architectures than DRUnet. 
For comparison, we also provide the runtime of BM3D on CPU.

\begin{table}[h!]
    \centering 
    \caption{
    \textbf{Architecture comparison.} 
    Runtime (in sec.), number of parameters $| \Theta |$ and FLOPs (in G) of the denoisers when used on $100$ images of size $3\!\times\! 481 \!\times\! 321$. 
    Values for the {\UNN}s are given for fixed $(K,J)= (20, 64)$.}
    \label{tab:runtime+flop}
    \vspace*{-0.2cm}
    
    \footnotesize
    \setlength\tabcolsep{0.1cm}
   \begin{tabular}{@{}ll|c|c|c@{}}
        & & Time (msec) & $| \Theta |$ & FLOPs ($\times 10^3$ G) \\
        \hline\hline
        \multicolumn{2}{l|}{BM3D}  &~$13 \!\! \times \!\! 10^3 \pm 317$ & -- & -- \\
        \hline
        \multicolumn{2}{l|}{DRUnet} & $96\pm 21$ & $32, 640, 960$ & $137.24$ \\
        \hline
        \multirow{4}{*}{\begin{turn}{90}LNO\end{turn}} & DDFB & $3\pm 1.5$ & $34, 560$ & \multirow{4}{*}{$2.26$}  \\
        & DDiFB &$3\pm 0.5$& $34,560$& \\
        & DCP &$6\pm 1$& $34,561$&  \\
        & DScCP & $7\pm 1$  & $34, 580$ &   \\
        \hline
        \multirow{4}{*}{\begin{turn}{90}LFO\end{turn}} & DDFB &$4\pm 17$ & $69,120$ & \multirow{4}{*}{$2.26$}  \\
        & DDiFB & $5\pm 15$ & $69,121$ & \\
        & DCP & $7\pm 14$ &$69,121$ &  \\
        & DScCP & $9\pm 15$& $69,160$ &   \\
        \hline
    \end{tabular}
\end{table}

\begin{figure}[!t]
	\scriptsize
    \centering
	\hspace*{-0.2cm}\begin{tabular}{@{}c@{}c@{}c@{}c@{}}
        DDFB-LNO& DDiFB-LNO &DCP-LNO &DScCP-LNO\\[-0.05cm]
        \includegraphics[width=0.26\columnwidth]{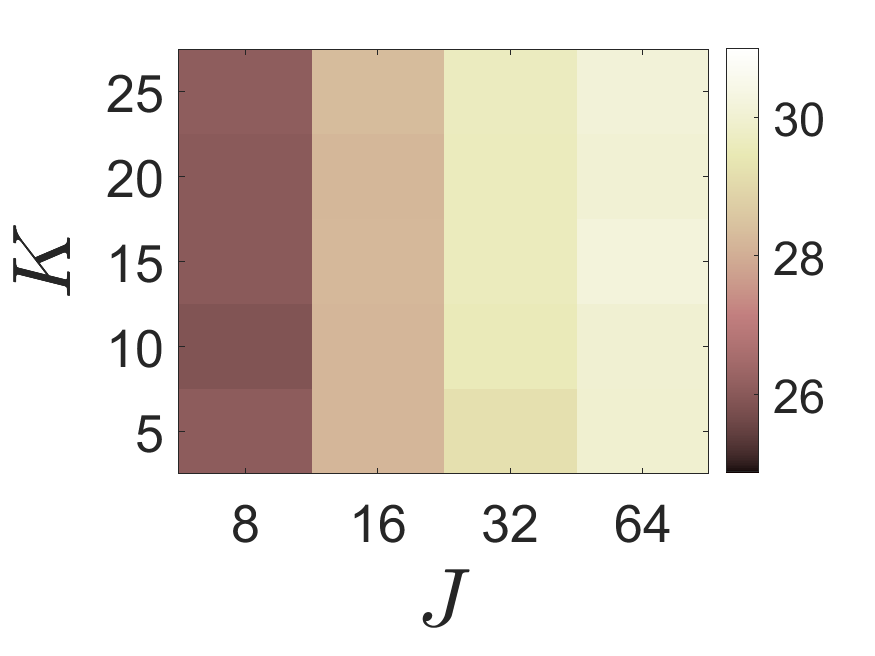}&
        \includegraphics[width=0.26\columnwidth]{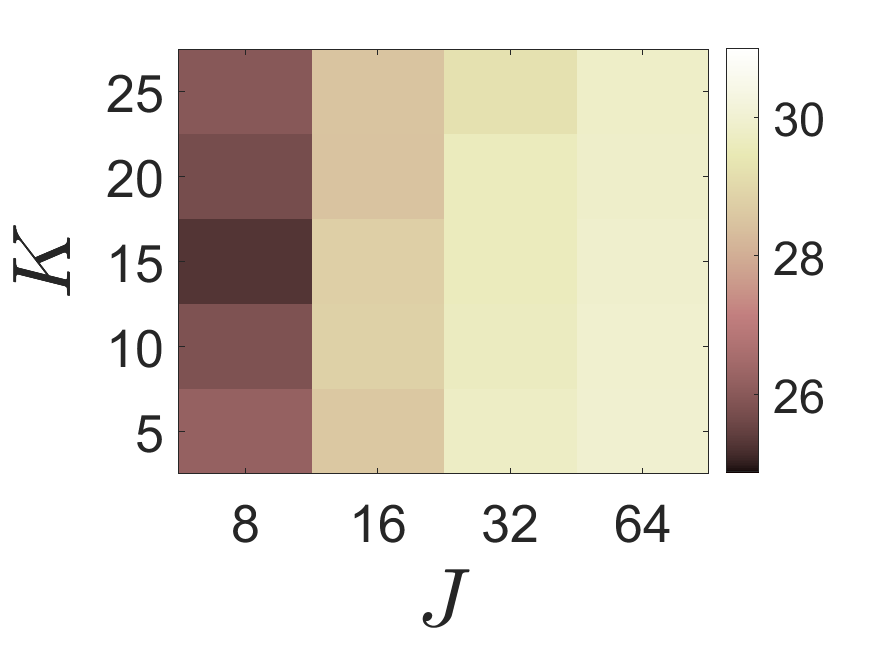}&
        \includegraphics[width=0.26\columnwidth]{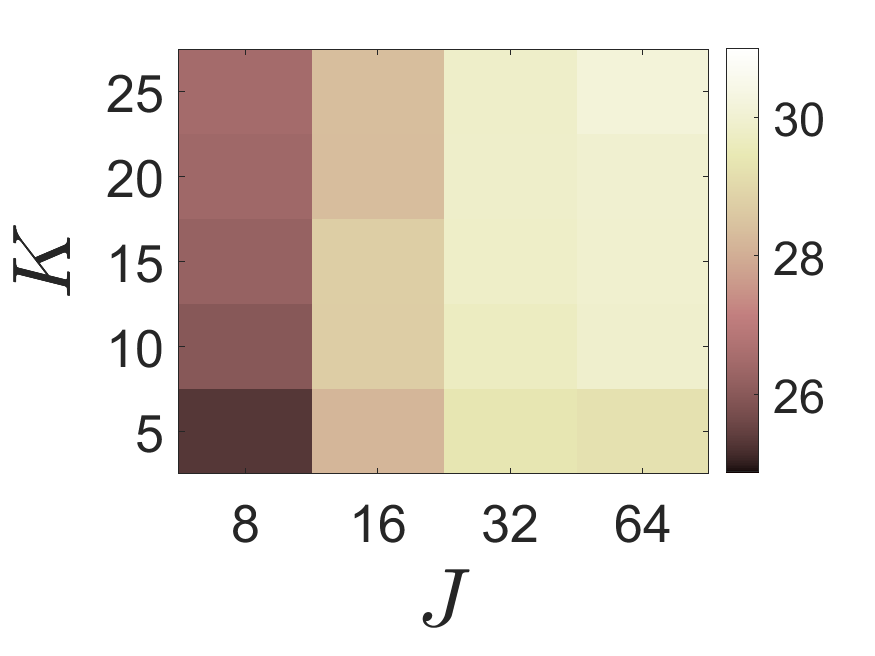}&
        \includegraphics[width=0.26\columnwidth]{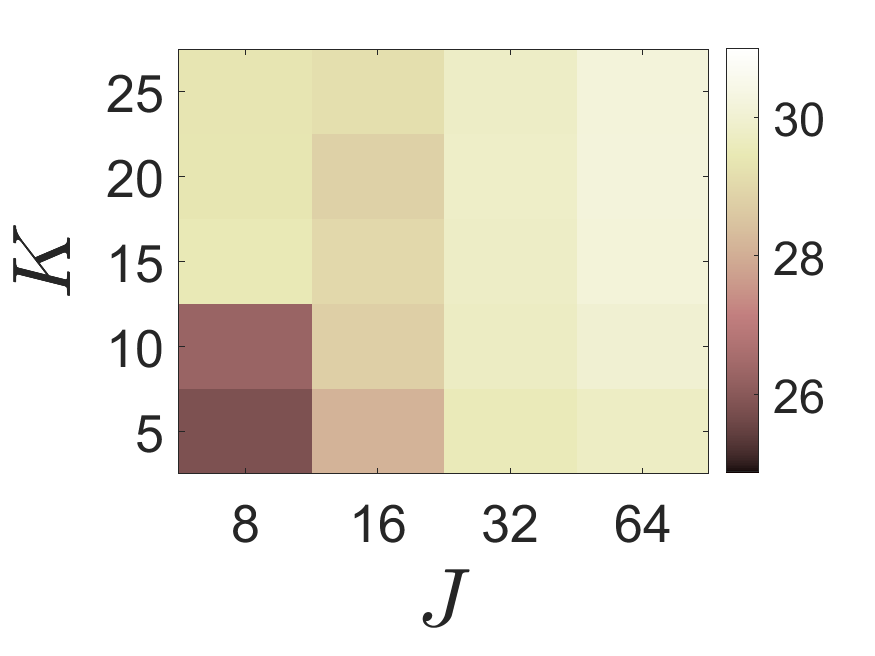}\\
        DDFB-LFO& DDiFB-LFO &DCP-LFO &DScCP-LFO\\[-0.05cm]
        \includegraphics[width=0.26\columnwidth]{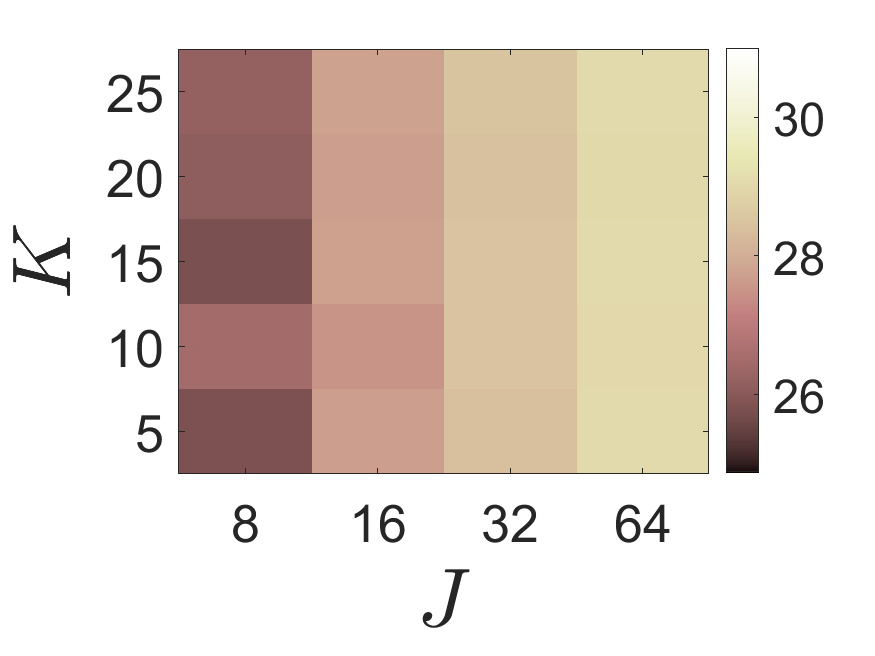}&
        \includegraphics[width=0.26\columnwidth]{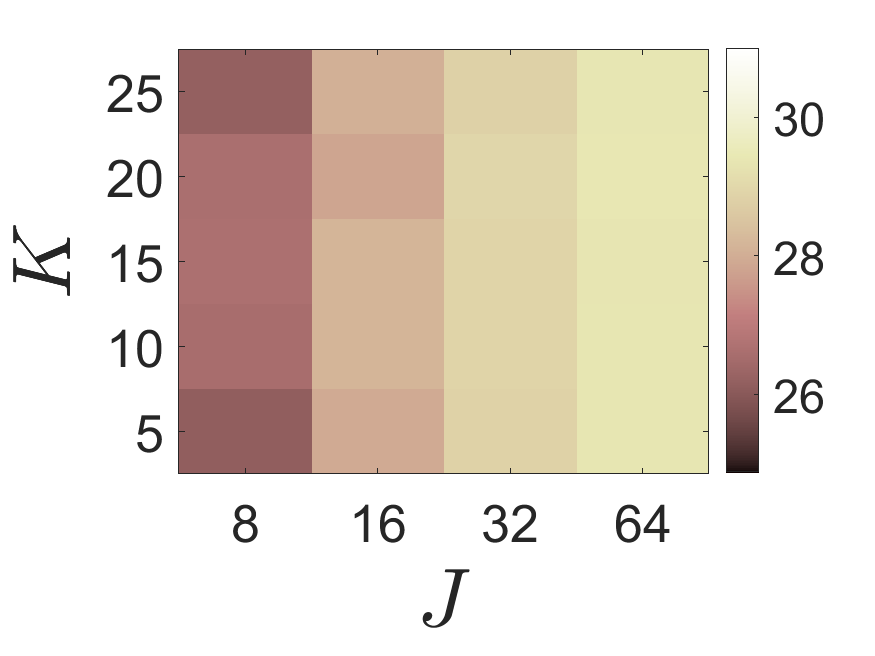}&
        \includegraphics[width=0.26\columnwidth]{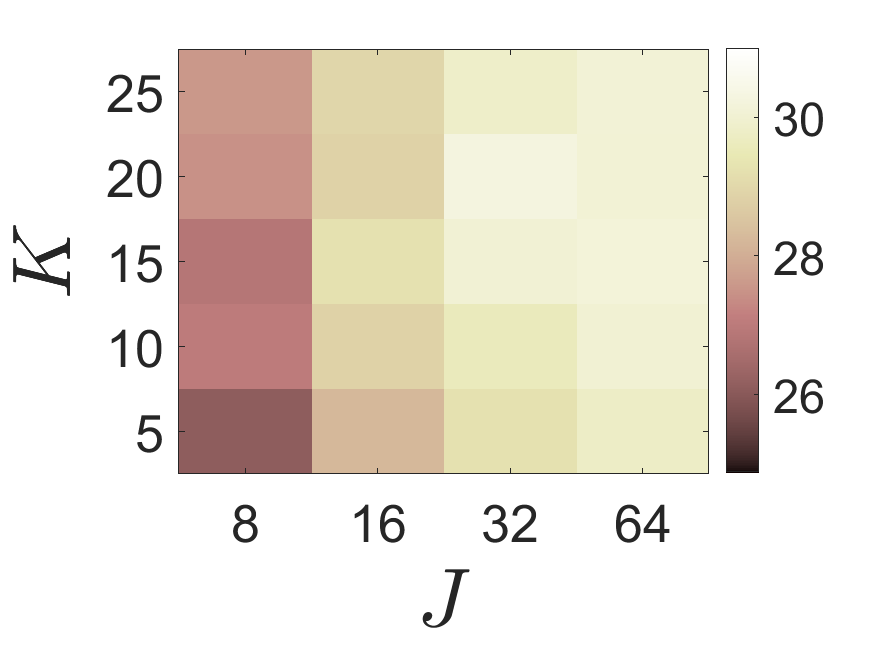}&
        \includegraphics[width=0.26\columnwidth]{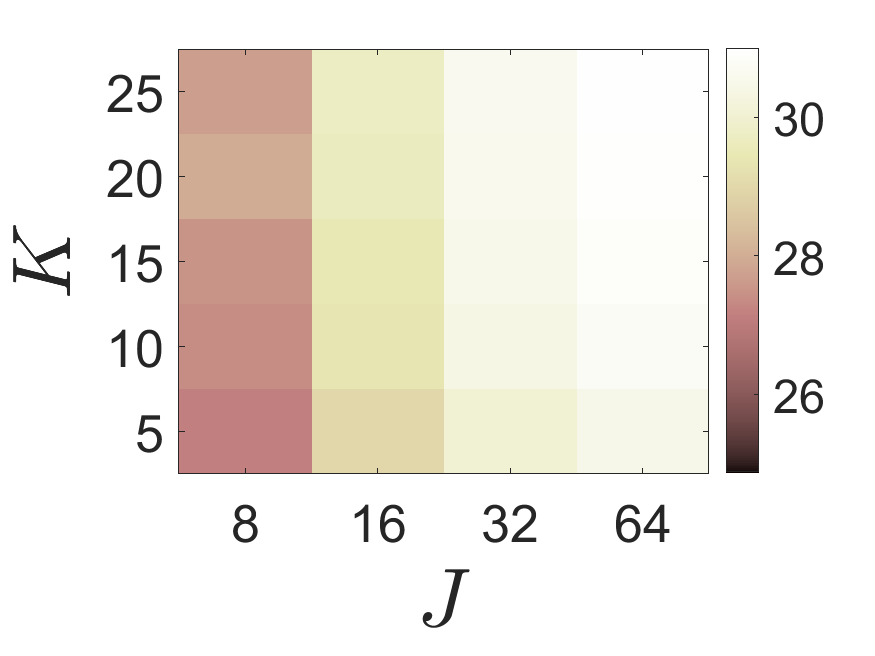}
	\end{tabular}

    \vspace*{-0.2cm}
 
	\caption{\textbf{{\UNN} denoising performance on Gaussian noise (Training Setting 1).} Average PSNR obtained with the proposed {\UNN}s, on $100$ images from BSD500 degraded with noise level $\delta=0.08$. Results are shown for $J\in \{8,16,32,64\}$ and $K\in \{5,10,15,20,25\}$. \textbf{Top row}: LNO settings. \textbf{Bottom row}: LFO settings.}
    \label{fig:psnr4methods}
    \vspace*{-0.6cm}
\end{figure}

\begin{figure}[!t]
    \centering
    \includegraphics[trim={4.8cm 2.5cm 1cm 0cm}, clip,width=9.5cm]{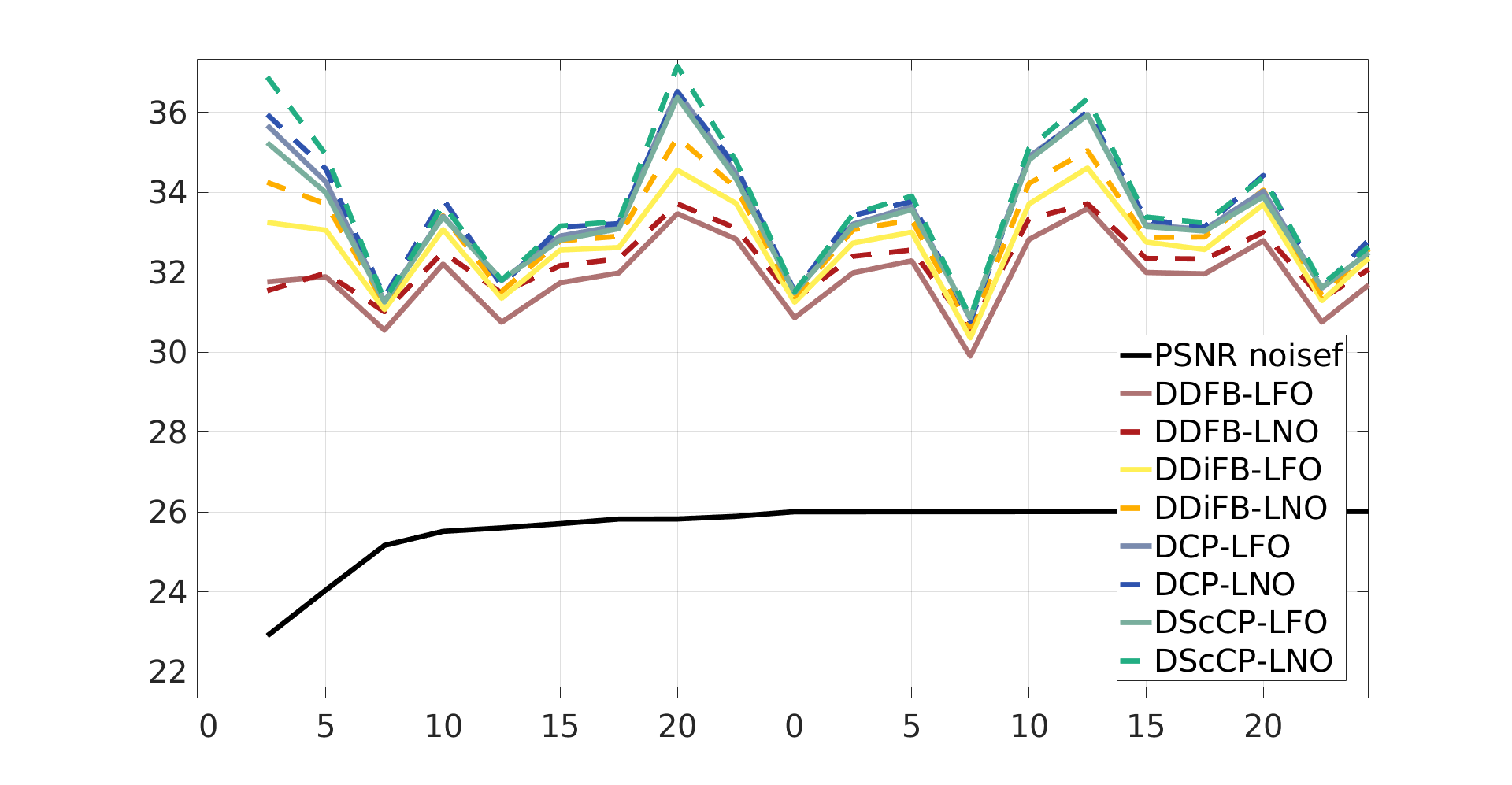}

    \vspace*{-0.2cm}
    
    \caption{\textbf{{\UNN} denoising performance on Gaussian noise (Training Setting 2).} PSNR values obtained with the proposed {\UNN}s (with $(K,J) = (20, 64)$), for 20 images of BSDS500 validation set, degraded with noise level $\delta=0.05$.   
    }
    \label{fig:setting1-denoise-compare}
\end{figure}

\subsection{{\UNN} denoising performance comparison}   
\label{Ssec:exp:perf-denoise}

\noindent \textbf{Training Setting 1 -- Fixed noise level}: 
We evaluate the denoising performance of the four proposed architectures considering either the LNO or LFO learning strategy, varying $K$ and $J$, on $|\mathbb{J}| = 100$ noisy images obtained from the test set, with noise standard deviation $\delta=0.08$. 
The average PSNR value for these test noisy images is $21.88$~dB. 

In Figure~\ref{fig:psnr4methods}, we show the averaged PSNR values obtained with the proposed unfolded networks. We observe a strong improvement of the denoising performances of all the NNs when increasing the size $J$ of convolution filters, as well as a moderate improvement when increasing the depth $K$ of the NNs.
All methods have very similar performances, with the exception of ScCP (both LNO and LFO) which has much higher denoising power.

\begin{figure*}[h]
    \centering\footnotesize
    \setlength\tabcolsep{0.1cm}
    \begin{tabular}{ccc}
     \includegraphics[width=3cm]{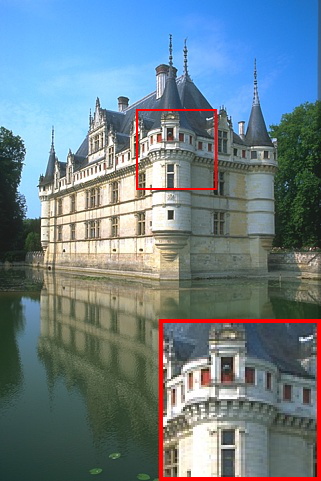}&
     \includegraphics[width=3cm]{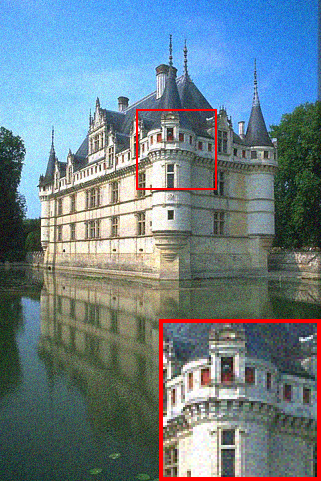}&
     \includegraphics[width=3cm]{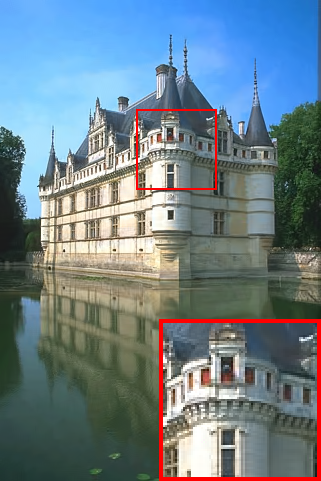} \\   
     &Noisy -- $26.03$dB &{BM3D -- $35.11$dB}  \\
     \includegraphics[width=3cm]{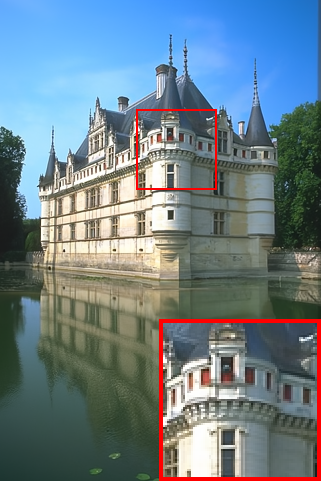}&
     \includegraphics[width=3cm]{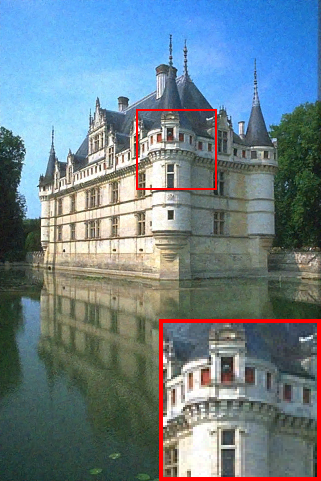}&
     \includegraphics[width=3cm]{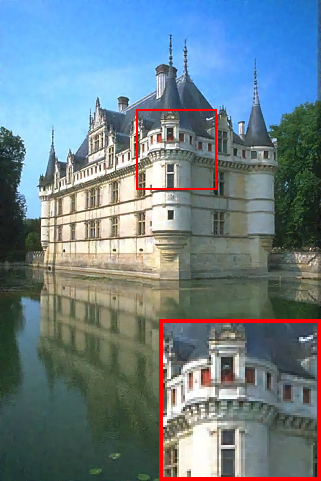}\\
     {DRUnet -- ${35.81}$dB} & DDFB-LNO -- $32.81$dB & DScCP-LNO -- $34.74$dB
    \end{tabular}
    
    \vspace*{-0.1cm}
    
    \caption{\textbf{Denoising performance on Gaussian noise (Training Setting 2).} Example of denoised images (and PSNR values) for Gaussian noise $\delta=0.05$ obtained with BM3D, DRUnet, and the proposed DDFB-LNO and DScCP-LNO, with $(K,J)=(20,64)$.}
    \label{fig:visual-denoise}

    \vspace*{-0.2cm}
\end{figure*}

\noindent \textbf{Training Setting 2 -- Variable noise level}: 
We evaluate the denoising performances of the proposed unfolded {\UNN}s on images degraded by a Gaussian noise with standard deviation $\delta=0.05$.
Figure~\ref{fig:setting1-denoise-compare} gives the PSNR values obtained from the proposed unfolded {\UNN}s, when applied to a random subset of $|\mathbb{J}| = 20$ images of BSDS500 validation set. Furthermore, Table~\ref{tab:compdenoised} presents the average PSNR values computed for a separate subset of $|\mathbb{J}| = 100$ images from the BDSD500 validation set. In the table, we further give the average PSNR value for DRUnet and BM3D. These results show that for all unfolding strategy, the LNO learning strategy improves the denoising performance over LFO. We further observe that D(Sc)CP outperforms DD(i)FB. DRUnet and BM3D however outperform the unfolded {\UNN}s on this experiments for PSNR values. For completeness, we give examples of a denoised image in Figure~\ref{fig:visual-denoise} obtained with DRUnet, BM3D, DDFB and DScCP. On visual inspection, results from DDFB may appear still slightly noisy compared with DRUnet, BM3D and DScCP. However results from DRUnet, BM3D and DScCP are very comparable, and DScCP might reproduce slightly better some textures (e.g., the water around the castle is slightly over-smoothed with DRUnet). In terms of trade-off performance versus computational time, DScCP offers the best compromise among all the methods.

\begin{table}[h!]
    \centering 
    \caption{
    \textbf{Denoising performance on Gaussian noise (Training Setting 2).} Average PSNR (and standard deviation) values (in dB) obtained with the proposed {\UNN}s for $(K,J)=(20,64)$, with DRUnet and BM3D, for $100$ noisy images of BSDS500 validation set ($\delta=0.05$, input PSNR$=25.94$dB). }
    \label{tab:compdenoised}

    \vspace*{-0.2cm}

    \footnotesize
    \setlength\tabcolsep{0.1cm}
   \begin{tabular}{@{}ll|c}
        & & PNSR  \\
        \hline\hline
        \multicolumn{2}{l|}{BM3D}  &$\mathbf{34.0\pm\! 1.78\!}$\\
        \hline
        \multicolumn{2}{l|}{DRUnet} & $\mathbf{34.7 \pm \! 1.89 \!}$ \\
        \hline
        \multirow{4}{*}{\begin{turn}{90}LNO\end{turn}} & DDFB & $32.4\pm \! 0.86 \!$\\
        & DDiFB &$33.1\pm \! 1.27 \!$\\
        & DCP & $33.5\pm \! 1.46 \!$\\
        & DScCP & $\mathbf{33.6\pm \! 1.53 }$  \\
        \hline
        \multirow{4}{*}{\begin{turn}{90}LFO\end{turn}} & DDFB &$32.1\pm \! 0.91 \!$ \\
        & DDiFB & $32.8\pm \! 1.12 \!$\\
        & DCP & $33.3\pm \! 1.35 \!$\\
        & DScCP & $33.3\pm \! 1.32 $ \\
        \hline
    \end{tabular}
\end{table}

\subsection{{\UNN} robustness comparison} 
\label{Ssec:exp:robust-lip}

Multiple works in the literature investigated NN robustness against perturbations \cite{jakubovitz2018improving, hoffman2019robust, pesquet2021learning}. 
Formally, given an input $\datad$ and a perturbation $\epsilon$, the error on the output can be upper bounded via the inequality
\begin{equation}    \label{eq:lipschitz}
    \Vert f_\Theta (\datad + \epsilon) - f_\Theta(\datad)\Vert \leq  \chi\Vert \epsilon\Vert.
\end{equation} 
The parameter $\chi>0$ can then be used as a certificate of the robustness for the network.
This analysis is important and complementary to quality recovery performance to choose a reliable model. 
According to \cite{Combettes_P_2020_siam-jmds_lip_cln}, in the context of feedforward NNs as those proposed in this work and described in Section~\ref{Sec:proposed-unfolded}, $\chi$ can be upper bounded by looking at the norms of each linear operator, i.e.,
\begin{equation}
    \chi \le \prod_{k=1}^K \Big(\|W_{k, \mathcal{P}} \|_S \times \|W_{k, \mathcal{D}} \|_S \Big).
\end{equation}
Unfortunately, as shown in \cite{Le_HTV_2022_p-eusipco-fas-pab}, this bound can be very pessimistic and not ideal to conclude on the robustness of the network. 
Instead, a tighter bound can be computed using the definition of Lipschitz continuity. Indeed, by definition the parameter $\chi$ in~\eqref{eq:lipschitz} is a Lipschitz constant of $f_\Theta$. This Lipschitz constant can be computed by noticing that it corresponds to the maximum of $\| \operatorname{J} f_\Theta(\datad) \|_S$ over all possible images $\datad \in \RR^N$, where $\operatorname{J}$ denotes the Jacobian operator. 
Since such a value is impossible to compute for all images of $\RR^N$, we can restrict our study on images similar to those used for training the network, i.e.,
\begin{equation}\label{eq:robustness}
    \chi 
    \approx \max_{ (\datad_s)_{s\in \mathbb{I}}} \| \operatorname{J} f_\Theta(\datad_s) \|_S.
\end{equation}
Such an approach has been proposed in~\cite{pesquet2021learning} for constraining the value of $\chi$ during the training process. In practice, the norm is computed using power iterations coupled with automatic differentiation in Pytorch. 

\smallskip

Motivated by these facts, we evaluate the robustness of our models by computing an approximation of $\chi$, as described in~\eqref{eq:robustness}, considering images in the test set $\mathbb{J}$ instead of the training set $\mathbb{I}$.
Here, $\mathbb{J}$ corresponds to 100 images randomly selected from BSD500 validation set, and for every $s \in \mathbb{J}$, $\datad_s = \overline{\estmsd}_s + \mathrm{w}_s$, where $\mathrm{w}_s \sim \mathcal{N}(0, \delta_s \mathrm{Id})$.

\smallskip

\begin{figure}[t!]
	\setlength{\tabcolsep}{0pt}
	\scriptsize
	\hspace*{-0.2cm}\begin{tabular}{@{}c@{}c@{}c@{}c@{}}
   DDFB-LNO& DDiFB-LNO &DCP-LNO &DScCP-LNO\\
		\includegraphics[width=0.26\columnwidth]{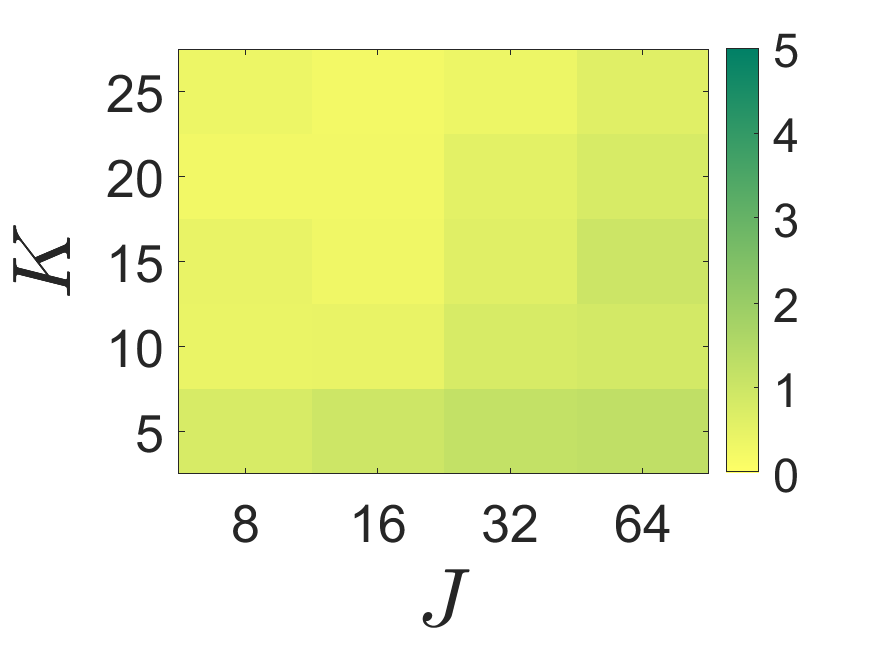}&
  \includegraphics[width=0.26\columnwidth]{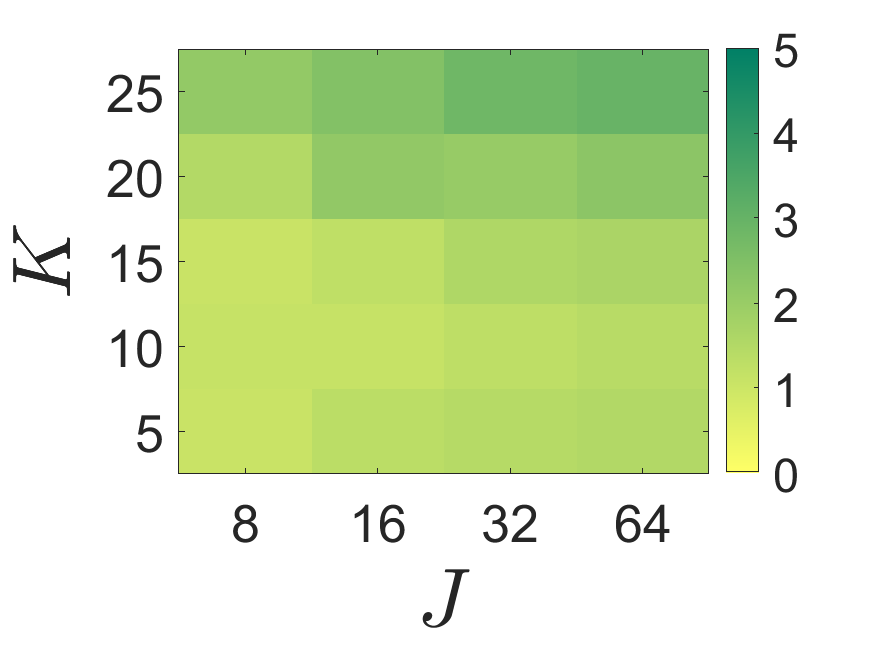}&
  \includegraphics[width=0.26\columnwidth]{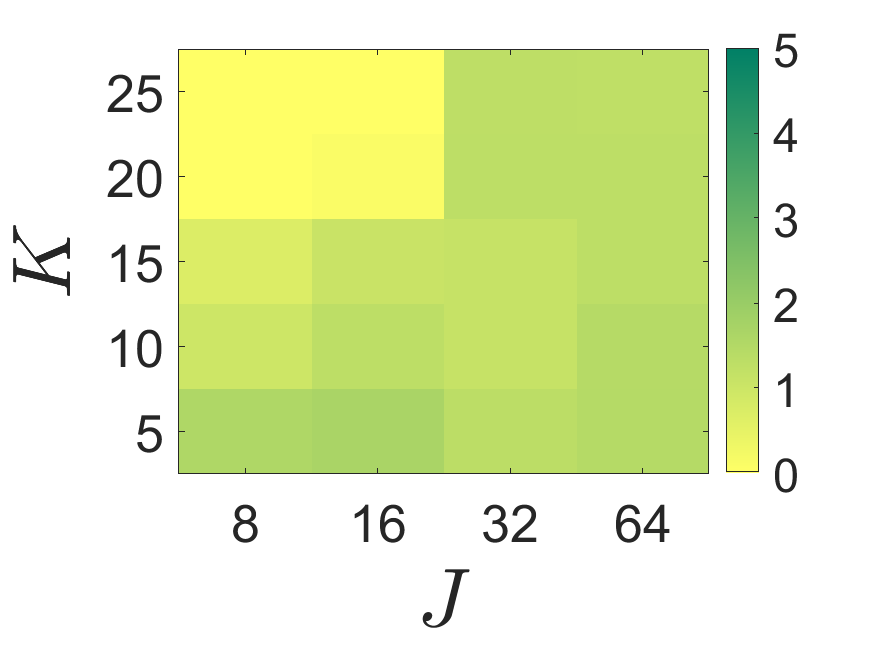}&
  \includegraphics[width=0.26\columnwidth]{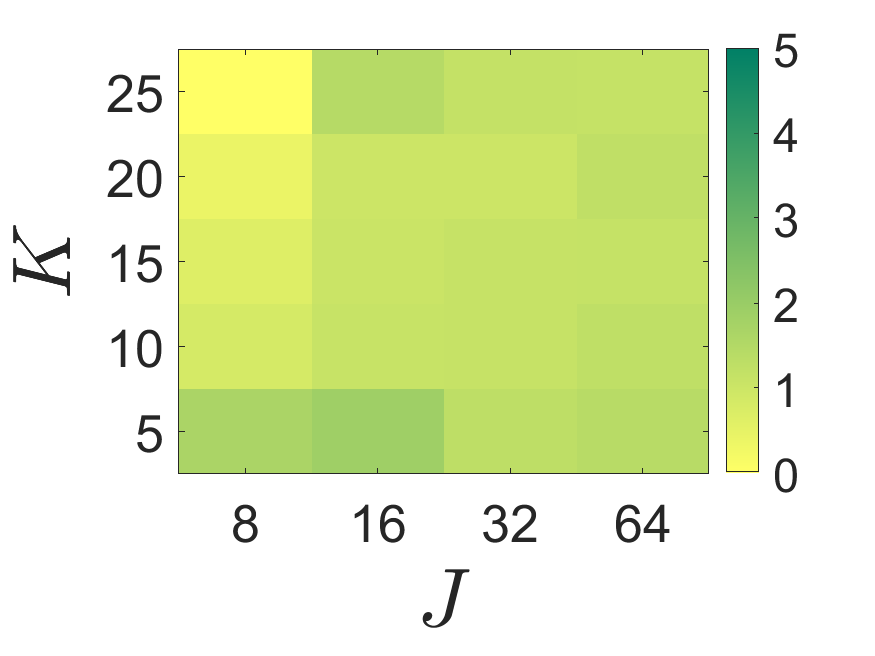}\\
		DDFB-LFO& DDiFB-LFO &DCP-LFO &DScCP-LFO\\
		\includegraphics[width=0.26\columnwidth]{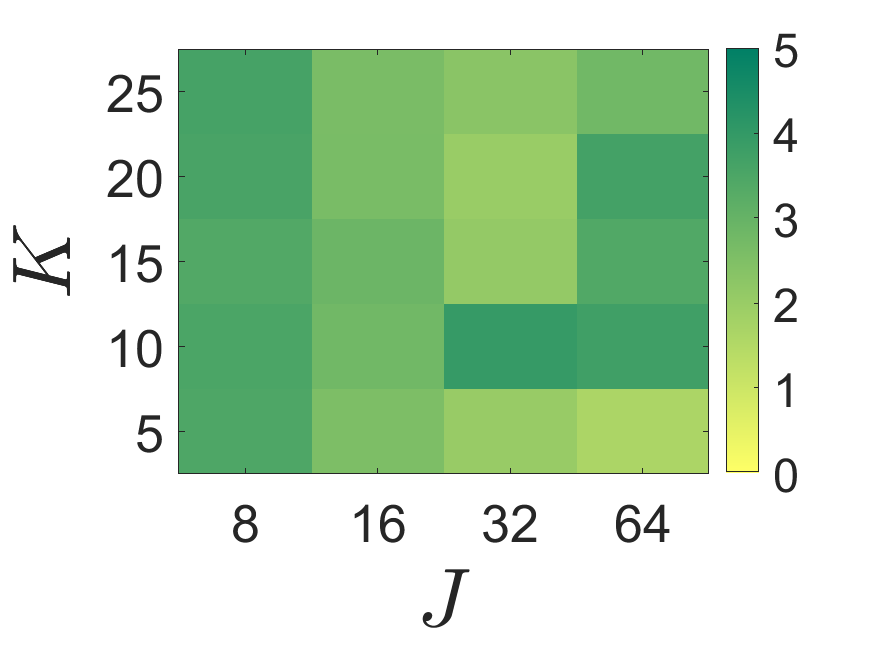}&
		\includegraphics[width=0.26\columnwidth]{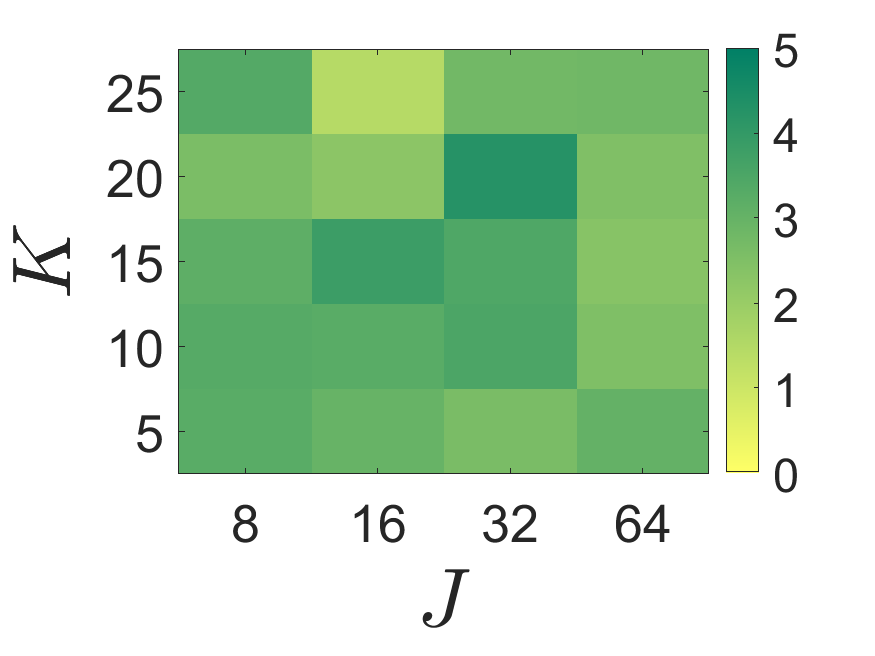}&
        \includegraphics[width=0.26\columnwidth]{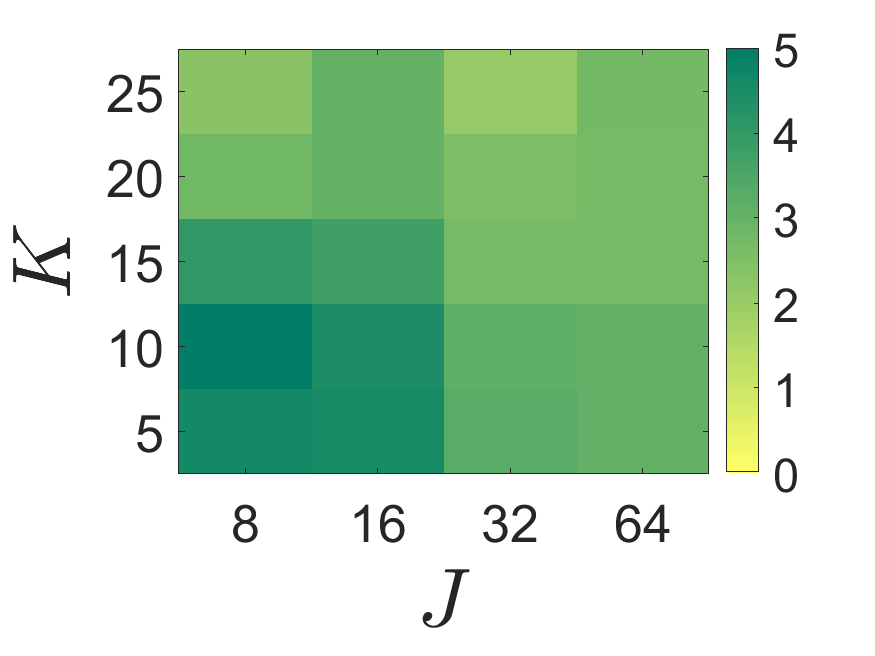}&
		\includegraphics[width=0.26\columnwidth]{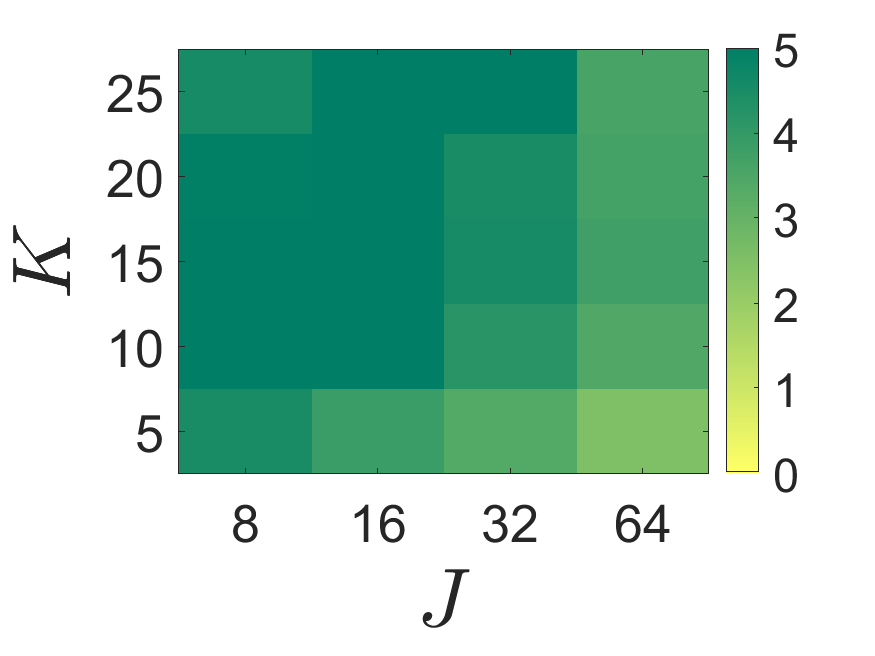}\\
	\end{tabular}

    \vspace{-0.2cm}
    
    \caption{\textbf{{\UNN} robustness comparison (Training Setting 1).} Values $\chi = \max_{s \in \mathbb{J}} \|\operatorname{J} f_\Theta(\datad_s) \|_S$ ($\log_2$ scale) for the proposed {\UNN}s, with $J\in \{8,16,32,64\}$ and  $K\in \{5,10,15,20,25\}$. \textbf{Top row}: LNO settings. \textbf{Bottom row}: LFO settings.
    }
    \label{fig:norm}
    
    \vspace{-0.3cm}
\end{figure}

\noindent \textbf{Training Setting 1 -- Fixed noise level}: 
For this setting we fix $\delta_s \equiv 0.08$. Corresponding values of $\chi$ for DD(i)FB and D(Sc)CP, with both LNO and LFO learning strategies, are reported in Figure~\ref{fig:norm}. 
For this setting, we show the evolution of the value of $\chi$ for $K\in \{5, 10, 15, 20, 25\}$ and $J \in \{8, 16, 32, 64\}$. We observe that the value of $\chi$ for LFO schemes are higher than their LNO counterparts, i.e., LFO schemes are less robust than LNO according to~\eqref{eq:lipschitz}. In addition, DDFB-LNO and D(Sc)CP-LNO seems to be the most robust schemes. Their $\chi$ value decreases slightly when $K$ increases, and increases slighly when $J$ increases.

\begin{figure}[t!]
    \centering
    \begin{center}
        \includegraphics[trim={4.8cm 0.cm 4.8cm 0cm},clip,scale=0.3]{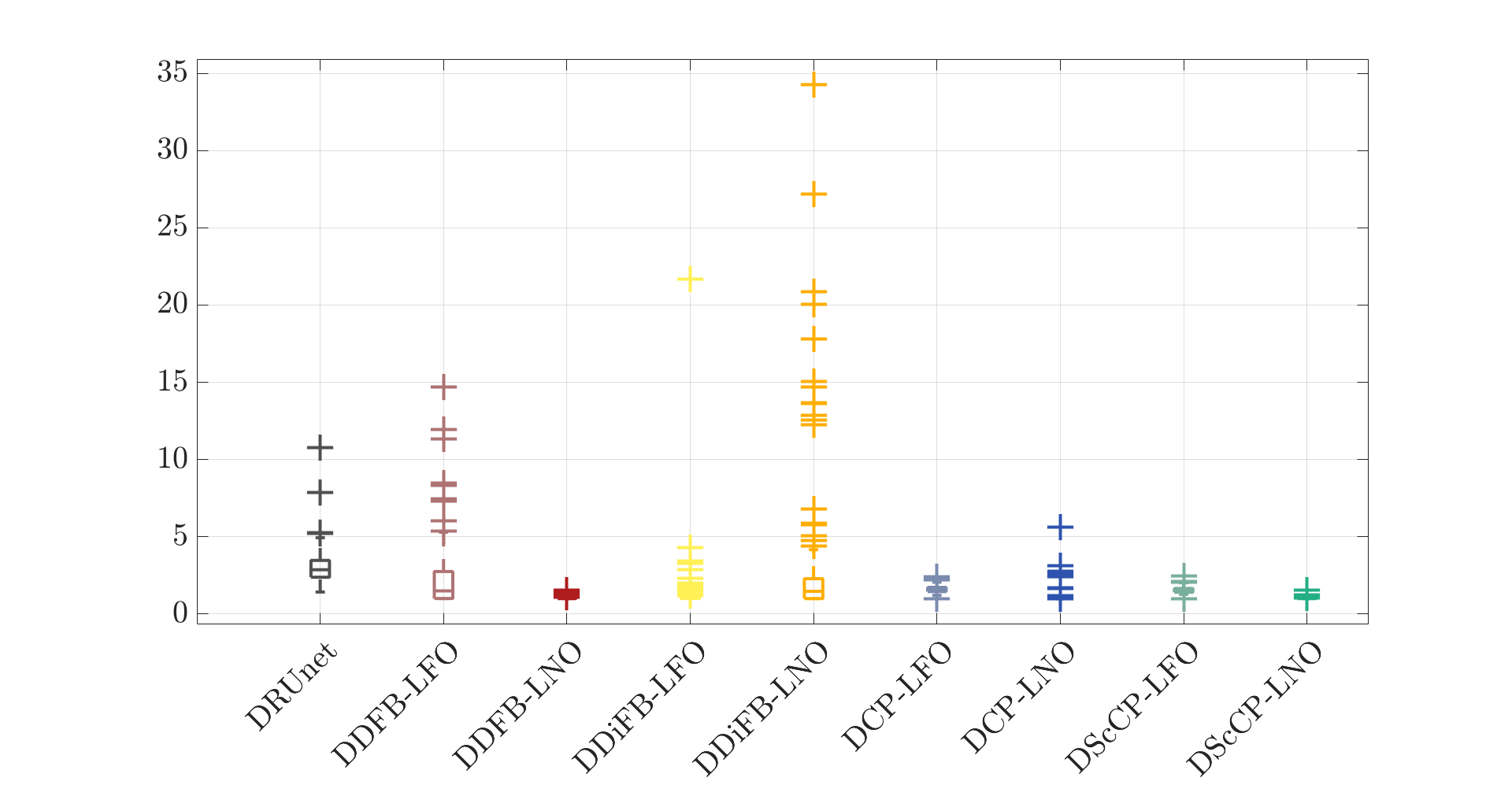}

        \vspace{0.1cm}
        {\footnotesize
        \setlength{\tabcolsep}{0.12cm}
        \begin{tabular}{@{}l||c|cc|cc|cc|cc@{}}
            &\multirow{2}{*}{DRUnet} & \multicolumn{2}{c|}{DDFB} & \multicolumn{2}{c|}{DDiFB} & \multicolumn{2}{c|}{DCP} & \multicolumn{2}{c}{DScCP} \\
            \cline{3-10}
            && LFO & LNO & LFO & LNO & LFO & LNO & LFO & LNO \\
            \hline\hline
            Mean & $1.92$ & $2.47$ & $1.04$ & $1.48$ & $3.80$ & $1.59$ & $1.15$ & $1.56$ & $1.02$ \\
            Median & $1.81$ & $1.49$ & $1.00$ & $1.13$ & $1.47$ & $1.60$ & $0.99$ & $1.52$ & $1.01$
        \end{tabular}}
    \end{center}

    \vspace{-0.2cm}
    
    \caption{\textbf{{\UNN} robustness comparison (Training Setting 2).} Distribution of $(\|\operatorname{J} f_{\Theta}(\datad_s)\|_S)_{s \in \mathbb J}$ for $100$ images extracted from BSDS500 validation dataset $\mathbb J$, for the proposed {\UNN}s and DRUnet.}
    \label{fig:fne}
\end{figure}

\noindent\textbf{Training Setting 2 -- Variable noise level}: 
For this setting, Figure~\ref{fig:fne} gives the box plots showing the distribution of the $100$ values of $(\|\operatorname{J} f_\Theta(\datad_s) \|_S)_{s \in \mathbb{J}}$, with $\delta_s \sim \mathcal{U}([0, 0.01])$. 
For the sake of completeness, the norms are also computed for DRUnet\cite{zhang2021plug}. 
Results are similar to the ones obtained with \textbf{Training Setting 1}.
Starting from the more robust schemes and moving to the less robust ones, we observe that DDFB-LNO and DScCP-LNO have the smallest values of $(\|\operatorname{J} f_\Theta(\datad_s) \|_S)_{s \in \mathbb J}$. D(Sc)CP-LFO have similar values to DDFB-LNO and DScCP-LNO, slightly larger and more spread out. Then DCP-LNO and DDiFB-LFO  are comparable, with larger values than the previously mentioned schemes, with a few outliers. Finally DDFB-LFO and DRUnet are comparable, with more outliers and higher median, Q1 and Q3 values, first and third quartiles respectively, followed by DDiFB-LNO that may have very high norm values, depending on the image (although Q3 is smaller than the Q1 value of DRUnet). Note that, overall DRUnet has the worst median, Q1 and Q3 values.

\subsection{{\UNN} robustness comparison on non-Gaussian noise}   
\label{Ssec:exp:perf-denoise-vs-robustness}

According to previous observations on our experiments, in the remainder we will focus on DDFB-LNO and DScCP-LNO, both appearing to have the best compromise in terms of denoising performance and robustness. 

\begin{figure*}[h]
 \setlength{\tabcolsep}{0.1pt}
    \centering\small
    \begin{tabular}{cccc} 
         \includegraphics[width=4.0cm]{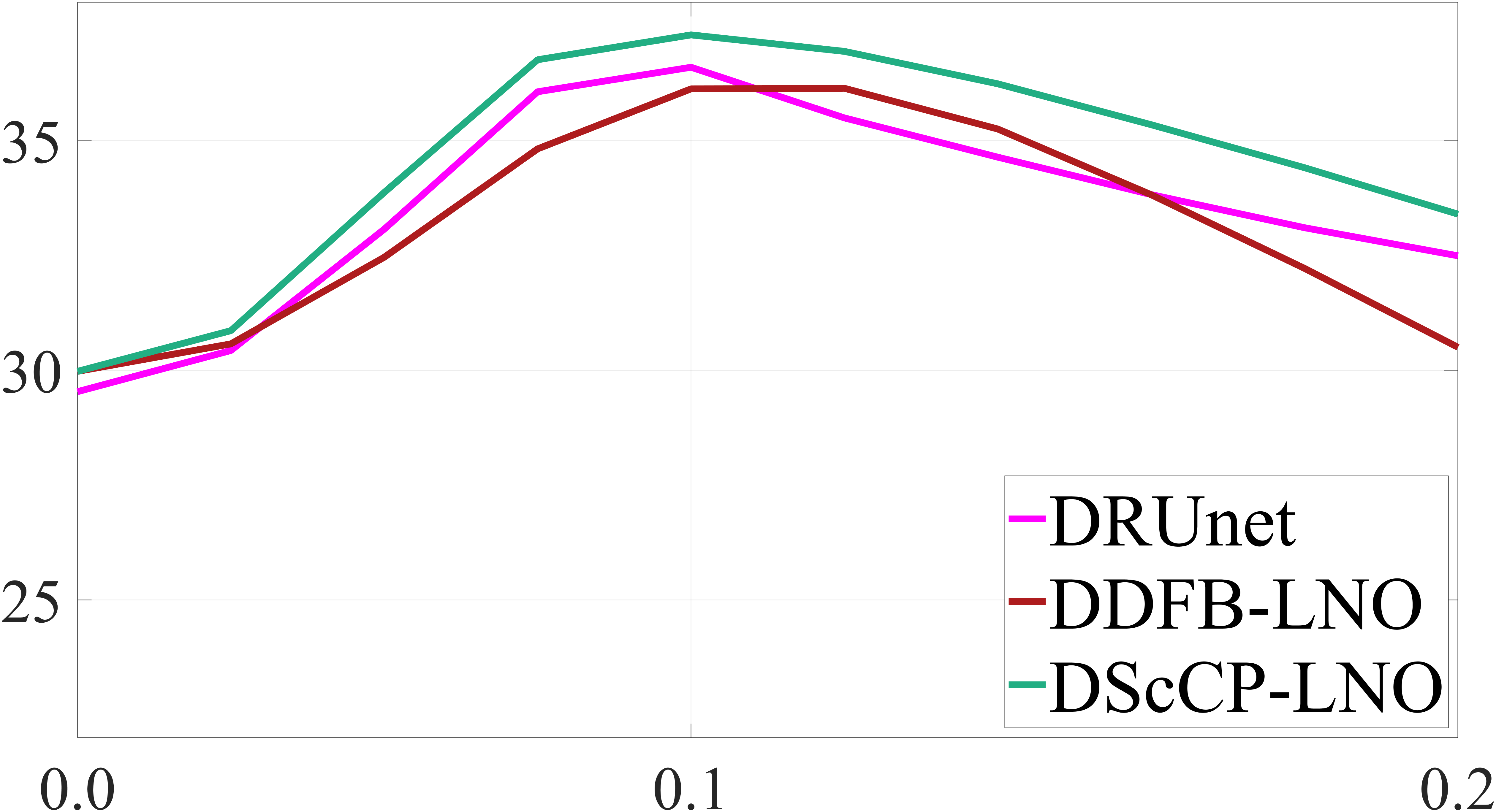}
         &\includegraphics[width=4.0cm]{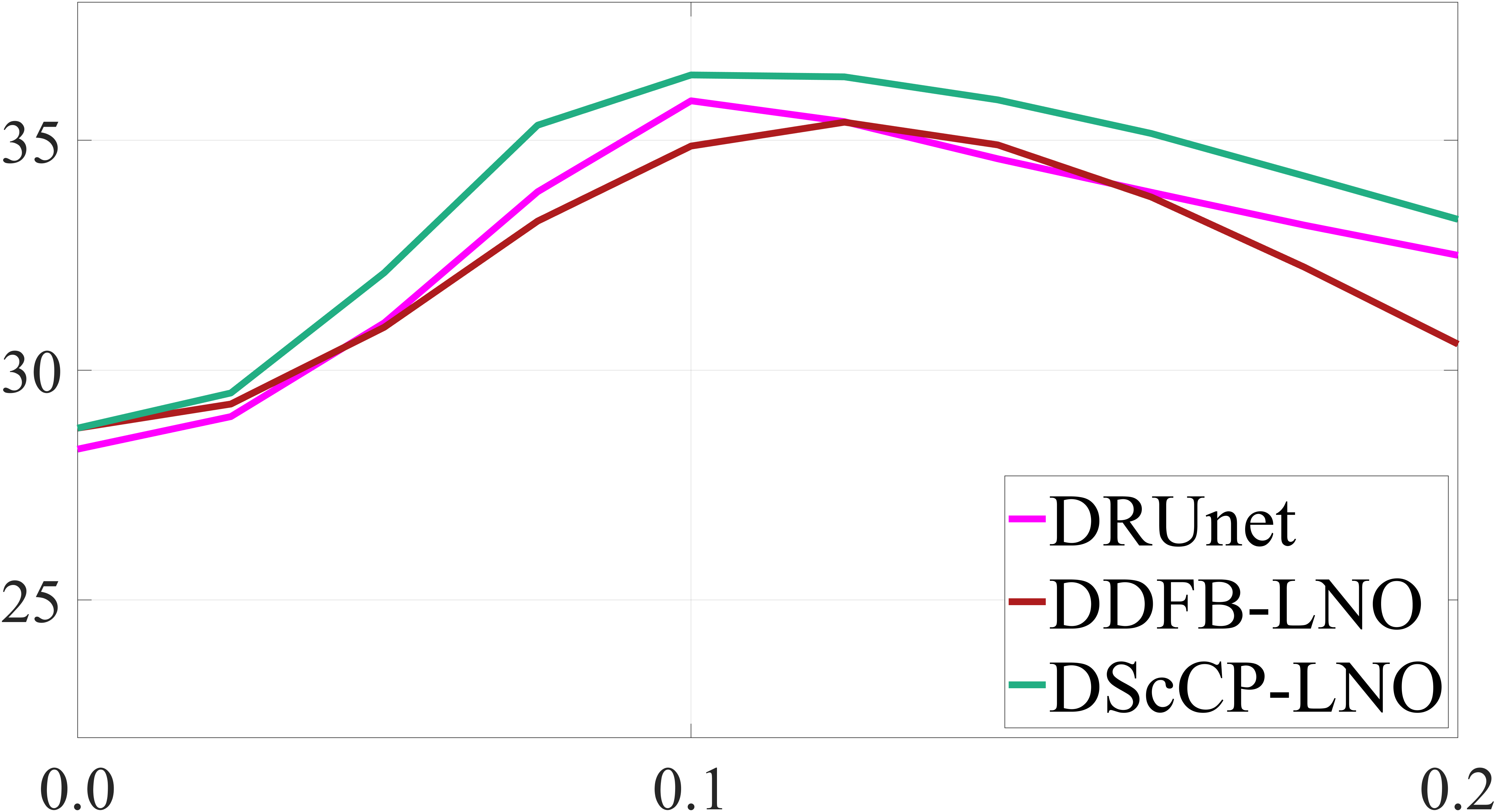}
         &\includegraphics[width=4.0cm]{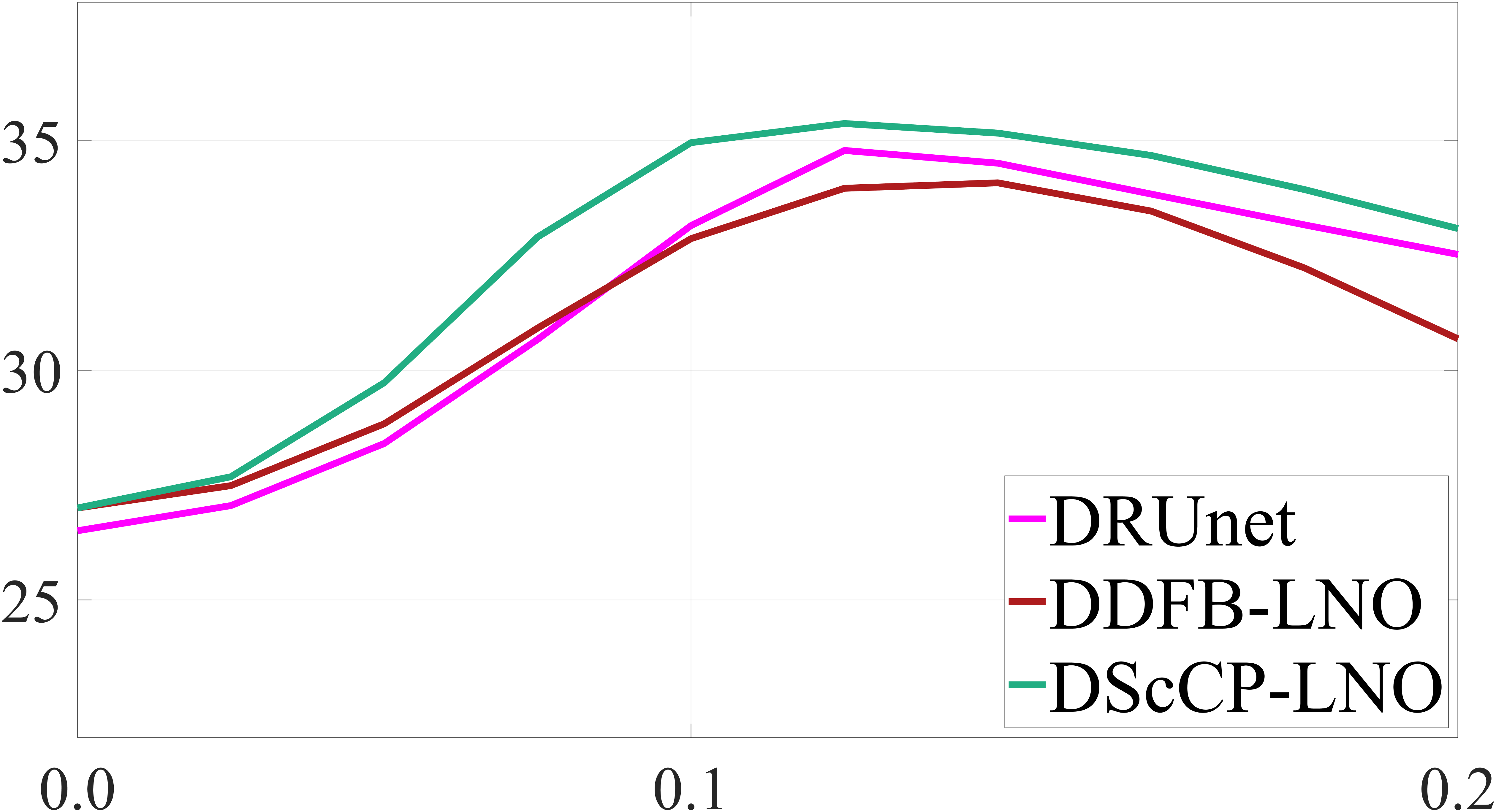}
         &\includegraphics[width=4.0cm]{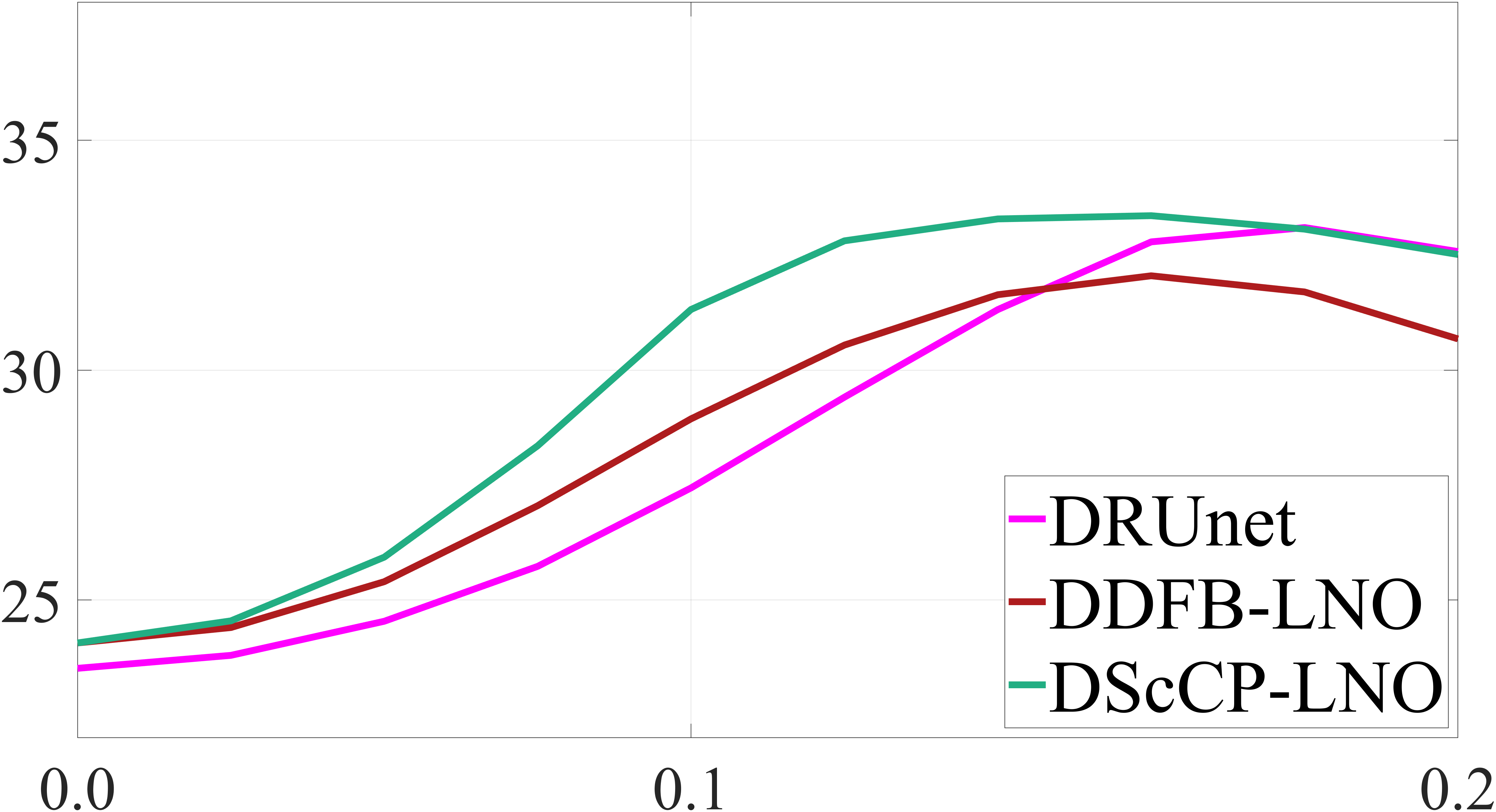}  \\
         \includegraphics[width=4.0cm]{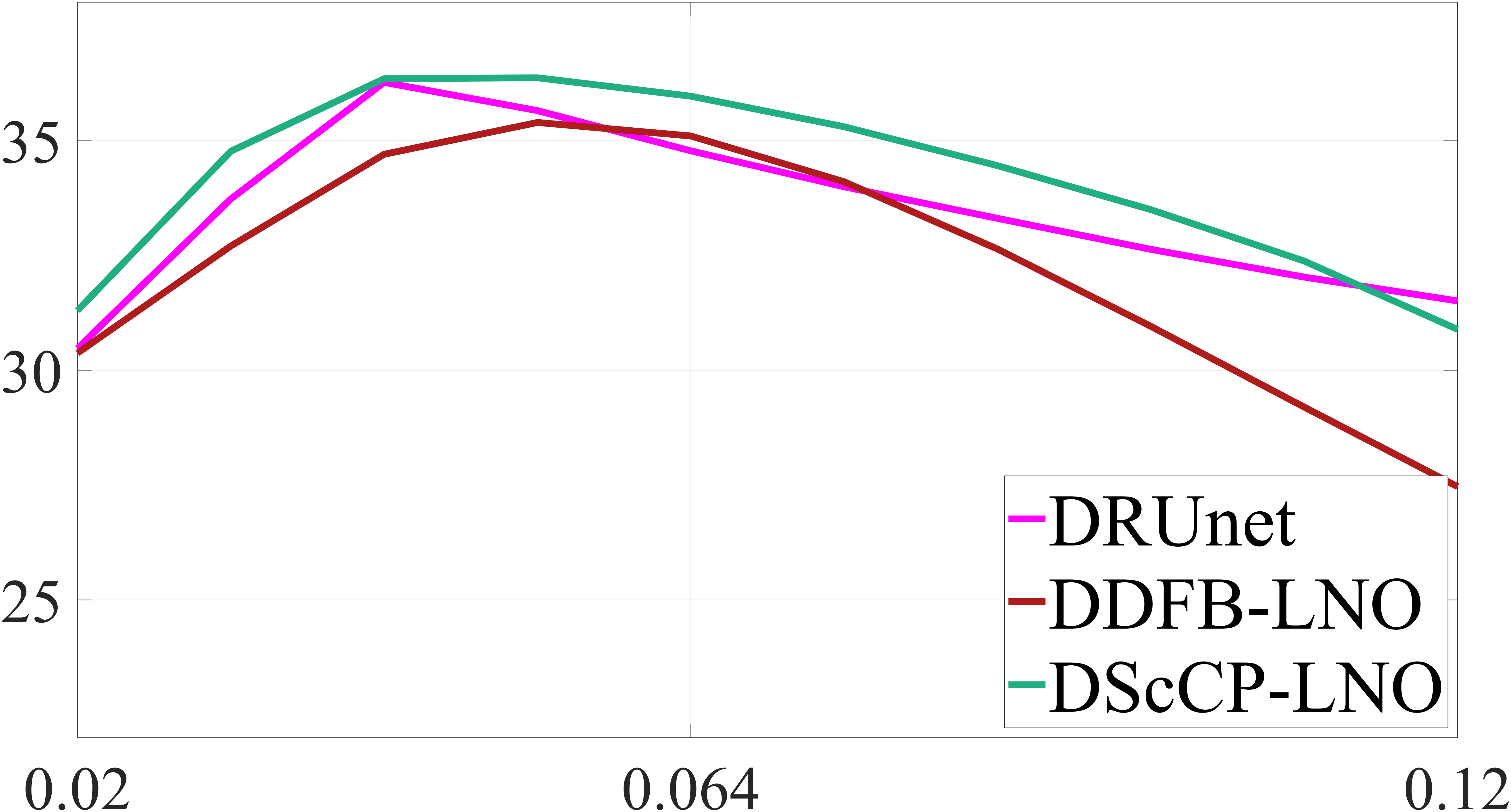}
         &\includegraphics[width=4.0cm]{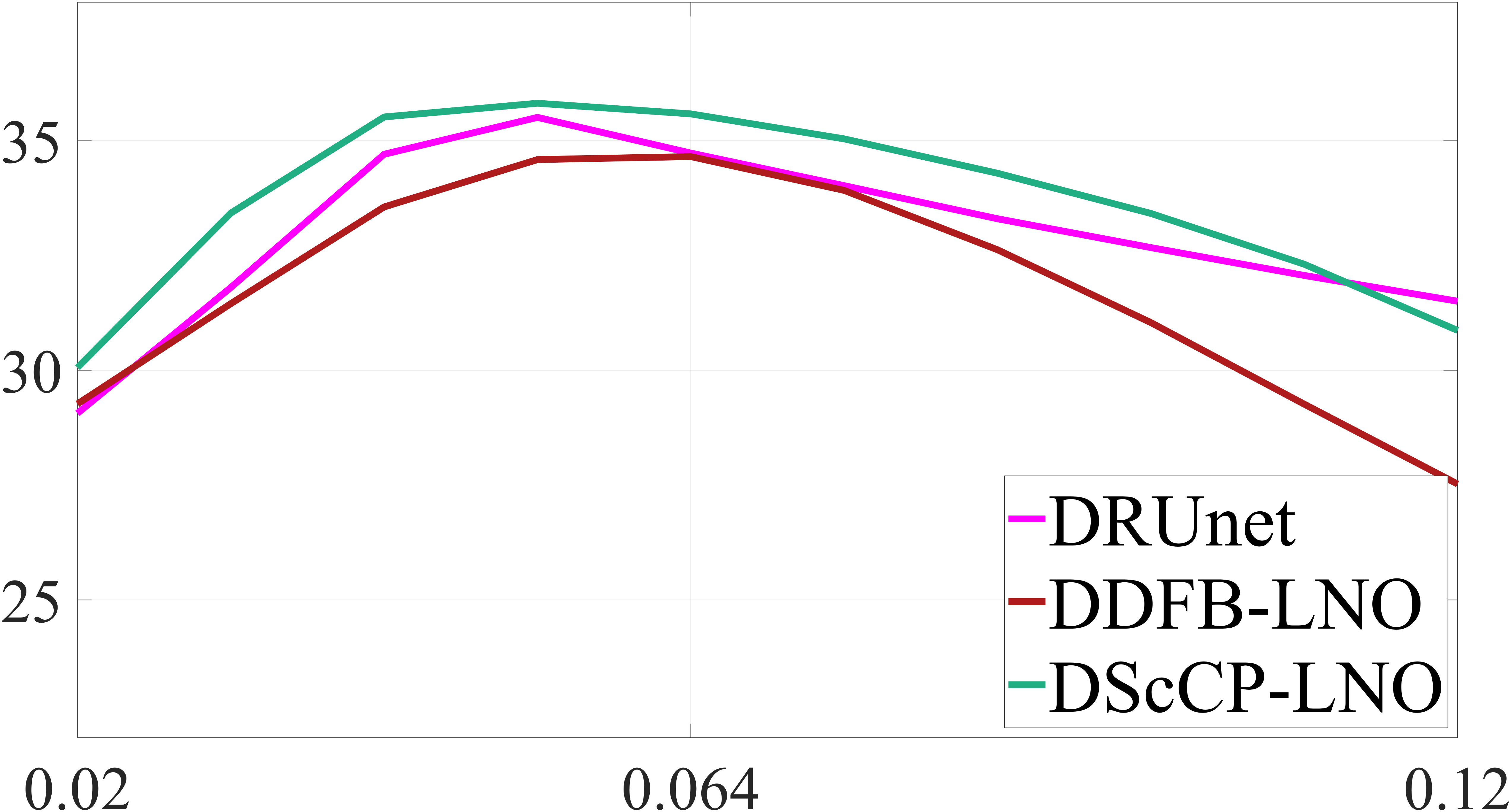}
         &\includegraphics[width=4.0cm]{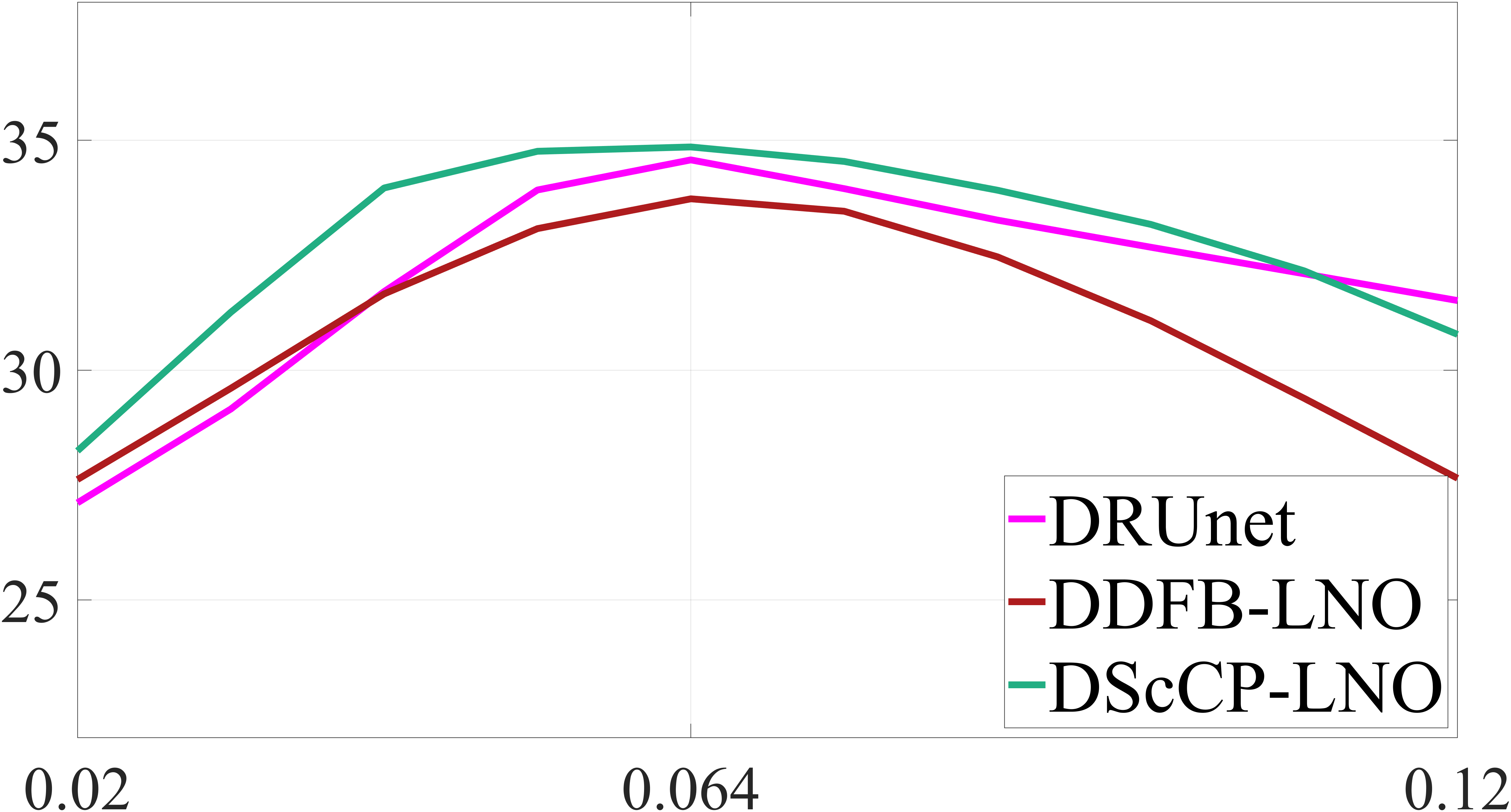}
         &\includegraphics[width=4.0cm]{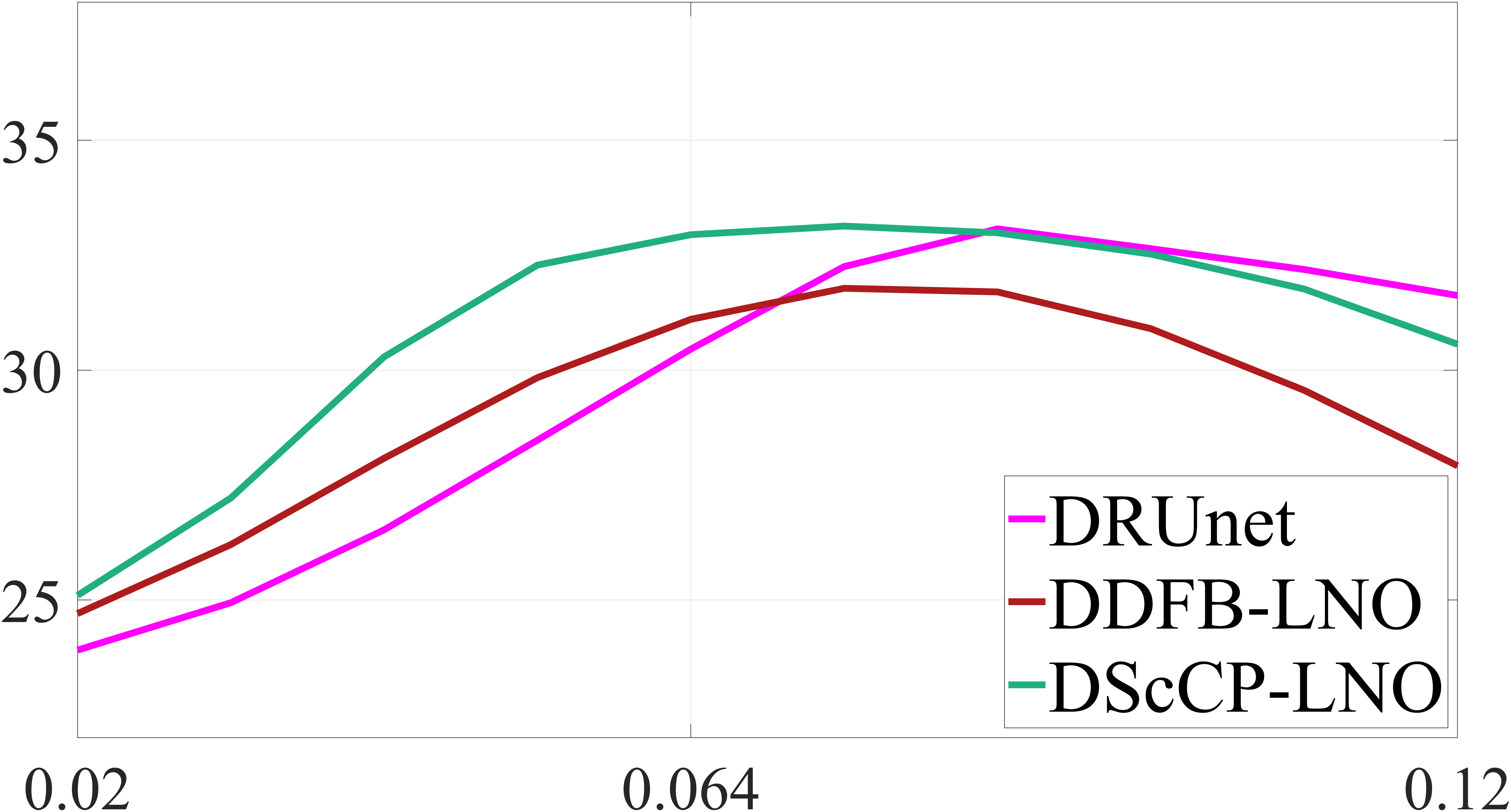}  \\
         400 photons & 300 photons &200 photons & 100 photons  
    \end{tabular}
    \caption{
    \small \textbf{{\UNN} robustness comparison on non-Gaussian denoising (Training Setting 2):} 
    \textbf{Top: Poisson denoising.} PSNR values obtained on Poisson noise for four noise levels (higher photon numbers indicate lower noise levels), on one image, and varying the regularization parameter $\nu$. 
    \textbf{Bottom: Poisson-Gaussian denoising.} PSNR values obtained on Poisson-Gaussian noise for four noise levels (varying the Poisson noie level, and for fixed standard deviation $\delta_G=0.02$ for the Gaussian noise), on one image, and varying the regularization parameter $\nu$. 
    }
    \label{fig:denoise-poisson}
\end{figure*}

\begin{figure*}[h]
\setlength{\tabcolsep}{2pt}
    \centering\scriptsize
    \begin{tabular}{rccccc}
    \rotatebox{90}{\hspace*{1.2cm}Poisson} &
      \includegraphics[width=2.9cm]{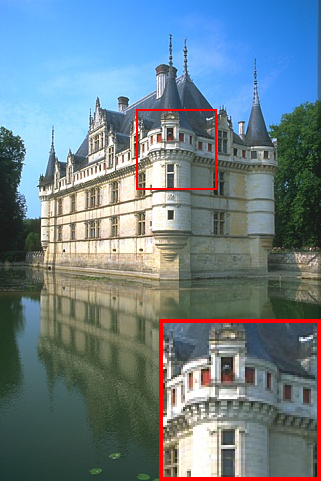}&
     \includegraphics[width=2.9cm]{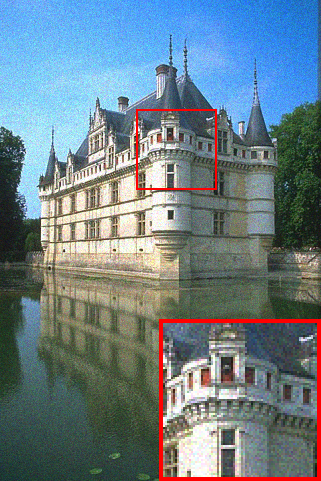}&
     \includegraphics[width=2.9cm]{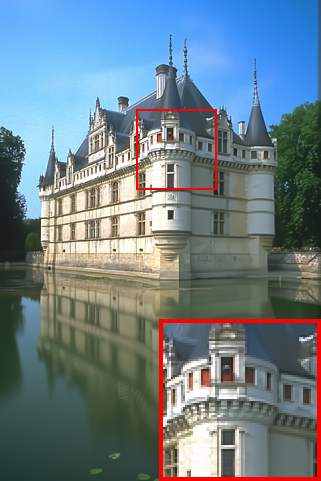}&
     \includegraphics[width=2.9cm]{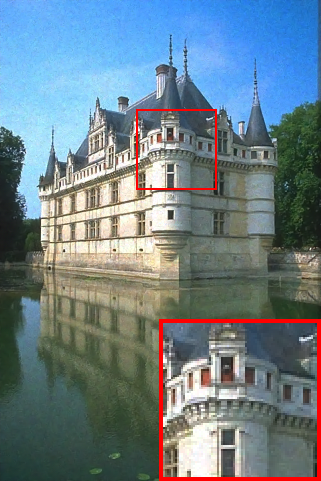}&
     \includegraphics[width=2.9cm]{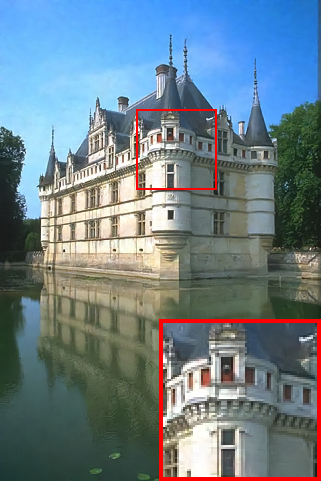}\\
     &Ground truth &Noisy -- $26.5$dB & DRUnet -- $34.8$dB & DDFB-LNO -- {$\mathbf{34.0dB}$} & DScCP-LNO -- \textcolor{red}{$\mathbf{35.4dB}$} \\
    \rotatebox{90}{\hspace*{1.2cm}Poisson-Gauss} &
     \includegraphics[width=2.9cm]{Figures/poisson_clean.png}&
     \includegraphics[width=2.9cm]{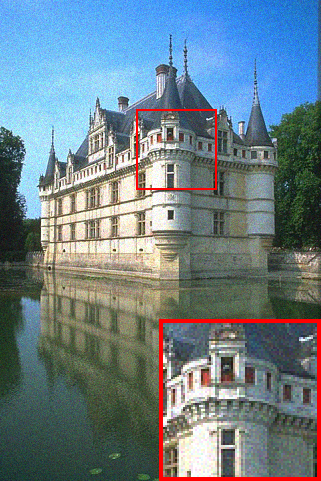}&
     \includegraphics[width=2.9cm]{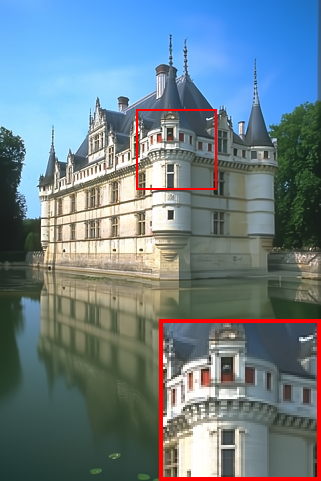}&
     \includegraphics[width=2.9cm]{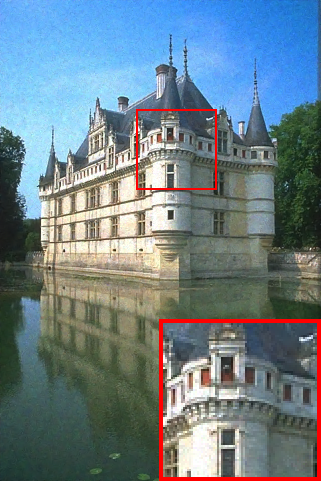}&
     \includegraphics[width=2.9cm]{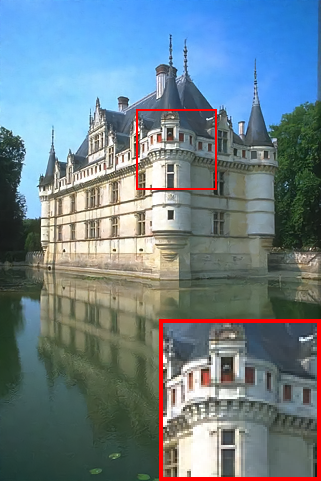}\\
     &Ground truth &Noisy -- $27.2$dB & DRUnet -- $35.5$dB & DDFB-LNO -- {$\mathbf{34.6dB}$} & DScCP-LNO -- \textcolor{red}{$\mathbf{35.8dB}$}
    \end{tabular}
    
    \vspace*{-0.1cm}
    
    \caption{
    \small \textbf{{\UNN} robustness comparison on non-Gaussian denoising (Training Setting 2):} 
    Example of denoised images (and PSNR values) for Poisson noise with the Poisson parameter set to 200 (\textbf{top)} and Poisson-Gauss noise with Gaussian standard deviation set to 0.02 and Poisson parameter set to 300 (\textbf{bottom)}, obtained with DRUnet and the proposed DDFB-LNO and DScCP-LNO (with $(K,J)=(20,64)$).}
    
    \label{fig:visual-denoise-poisson}

    \vspace*{-0.2cm}
\end{figure*}

To further assess the robustness of the proposed DDFB-LNO and DScCP-LNO, we evaluate them on denoising tasks that are different from those used during the training process. For the sake of completeness, we also compare our {\UNN}s with the state-of-the-art denoising network DRUnet.  
All networks have been trained as Gaussian denoisers (\textbf{Training Setting 2}). We observed in Section~\ref{Ssec:exp:perf-denoise} that DRUnet, with $1000$ times more parameters than our {\UNN}s, shows impressive performances in terms of Gaussian denoising (in PSNR). 
However, we also observed that DRUnet has a higher Lipschitz constant than the proposed DDFB-LNO and DScCP-LNO. Hence we will further assess the robustness of the networks byusing them for denoising Poisson and Poisson-Gaussian noises 
(i.e., on different noise settings than the training Gaussian setting).
For the Poisson noise, we vary the noise level to observe from $400$ photons (low noise level) to $100$ photons (high noise level). For the Poisson-Gaussian noise we first add a white Gaussian noise with standard deviation $\delta=0.02$, and then a Poisson noise varying from $400$ photons to $100$ photons.
%
%
The evaluation includes both quantitative and visual analysis, presented in Figures~\ref{fig:denoise-poisson} and~\ref{fig:visual-denoise-poisson}, respectively. 
For Poisson-Gaussian noise, DRUnet and the proposed DScCP-LNO give similar reconstruction quality, while DDFB-LNO leads to slightly lower reconstruction quality. For the Poisson noise, DScCP-LNO gives slighlty better estimates in terms of PSNR than the two other networks. 


\subsection{Robustness comparison: Image deblurring with PnP}
\label{s:pnp}

\noindent\textbf{Deblurring problem}\\
Another measure to assess the proposed unfolded NNs is to use them in a Plug-and-Play framework, for image deblurring.
In this context, the objective is to find an estimate $\estmsd^\dagger \in \RR^N$ of an original unknown image $\overline{\estmsd} \in \RR^N$, from degraded measurements $\datar \in \RR^M$ obtained through
\begin{equation}\label{degradation-model}
   \datar=\mathrm{A} \overline{\estmsd}+ \mathrm{b},
\end{equation}
where $\mathrm{A} \colon \mathbb{R}^N\to \mathbb{R}^M$ is a linear blurring operator, and $\mathrm{b} \in \mathbb{R}^M$ models an additive white Gaussian noise, with standard deviation $\sigma>0$. 
A common method to solve this inverse problem is then to find the MAP estimate $\estmsd^\dagger$ of $\overline{\estmsd}$, defined as a minimizer of a penalized least-squares objective function. 
A general formulation is given by
\begin{equation}\label{inverse-problem}
\text{find } \estmsd^\dagger_{\mathrm{MAP}} \in \Argmin{\estmsd\in C  }\;\frac{1}{2} \|\mathrm{A}\estmsd-\datar\|^2+ \lambda g(\D x),
\end{equation}
where $C \subset \RR^N$, $\D\colon \RR^N \to \RR^{\mid\mathbb{F}\mid }$ and $g \colon \RR^{\mid\mathbb{F}\mid}\to (-\infty,+\infty]$ are defined as in \eqref{eq:prox-function}, and $\lambda \propto \sigma^2$ (i.e., there exists $\beta>0$ such that $\lambda = (\beta\sigma)^2$) is a regularization parameter.

\medskip

\noindent\textbf{PnP-FB algorithm}\\
The idea of PnP algorithms is to replace the penalization term (often handled by a proximity operator) by a powerful denoiser. There are multiple choices of denoisers, that can be classified into two main categories: hand-crafted denoisers (e.g. BM3D \cite{Dabov_2007_j-tip_image_den_bm3d}) and learning-based denoisers (e.g., DnCNN \cite{zhang2017beyond} and UNet \cite{ronneberger2015u}).
PnP methods with NNs have recently been extensively studied in the literature, and widely used for image restoration (see, e.g., \cite{pesquet2021learning, Kamilov_U_2023_j-ieee-spm, Hurault_S_2021_arxiv, Repetti_A_2022_eusipco_dual_fbu}). 

It is to be noted that an unfolded {\UNN} could be designed for solving directly~\eqref{inverse-problem}, similarly to~\cite{Jiu_M_2021_jstsp_deep_pdp}. However PnP methods have the advantage over unfolded networks that they are generalisable when changing the measurement operator $\mathrm{A}$ in the restoration problem~\eqref{degradation-model}. Hence the objective of this section is to follow a similar approach as in \cite{Repetti_A_2022_eusipco_dual_fbu}, and to
plug the proposed unfolded {\UNN}s in a FB algorithm to solve~\eqref{degradation-model}. The objective is to further assess the robustness of the proposed unfolded strategies. Following the approach proposed in \cite{Repetti_A_2022_eusipco_dual_fbu}, the PnP-FB algorithm is given by
\begin{equation}\label{algo:fbpnp}
    \begin{array}{l}
    \text{Let } \estmsd_0 \in \RR^N, \dualvar_0 \in \RR^{| \mathbb{F}|} \\
    \text{For } t= 0,1,\dots\\
    \left\lfloor 
    \begin{array}{l}
        \datad_t = \estmsd_t-\gamma \mathrm{A}^\top\left(\mathrm{A}\estmsd_t - \datar\right), \\[0.1cm]
        (\estmsd_{t+1}, \dualvar_{t+1}) = f^K_{\datad_t, \lambda\gamma , \Theta} (\datad_t, \dualvar_t ),
    \end{array}
    \right. 
    \end{array}
\end{equation}
where, for every $t\in \NN$, $f^K_{\datad_t, \lambda\gamma , \Theta}$ is either D(i)FB or D(Sc)CP. 
In algorithm~\eqref{algo:fbpnp}, parameters $(\lambda, \gamma)$ are given as inputs of $f^K_{\datad_t, \lambda\gamma , \Theta}$. Precisely, the regularization parameter $\nu$ for the denoising problem~\eqref{prob:2-terms-minimization} is chosen to be the product between the regularization parameter $\lambda$ for the restoration problem~\eqref{inverse-problem} and the stepsize of the algorithm $\gamma$, i.e., $\nu = \lambda \gamma$.

The following result is a direct consequence of Corollary~\ref{cor:limit-NN} combined with convergence results of the FB algorithm \cite{combettes2005signal}.
\begin{theorem} \label{thm:cvg-pnpfb}
Let $(\estmsd_t)_{t\in \NN}$ be a sequence generated by \eqref{algo:fbpnp}, with $f^K_{\datad_t, \lambda\gamma , \Theta}$ being DD(i)FB-LNO or D(Sc)CP-LNO.
Assume that $\gamma \in (0, 2/\| \A \|_S^2)$ and that, for every $k\in \{1, \ldots, K\}$, $\D_{k, \mathcal{D}} = \D$ and $\D_{k, \mathcal{P}} = \D^\top$ for $\D \colon \RR^N \to \RR^{|\mathbb{F}|}$. 
Under the same conditions as Corollary~\ref{cor:limit-NN}, if $K \to \infty$, then $(\estmsd_t)_{t\in \NN}$ converges to a solution $\estmsd^\dagger$ to problem~\eqref{inverse-problem}, and 
\begin{equation}\label{thm:cvg-fb:mono}
    (\forall t \in \NN^*)\quad 
    \| \estmsd_{t+1} - \estmsd_t \| \le \| \estmsd_t - \estmsd_{t-1} \|.
\end{equation}
\end{theorem}
A few comments can be made on Theorem~\ref{thm:cvg-pnpfb} and on the PnP algorithm~\eqref{algo:fbpnp}.
First, \cite[Prop. 2]{Repetti_A_2022_eusipco_dual_fbu} is a particular case of Theorem~\ref{thm:cvg-pnpfb} for DDFB-LNO. 
Second, in practice, only a fixed number of layers $K$ are used in the {\UNN}s, although Theorem~\ref{thm:cvg-fb:mono} holds for $K\to +\infty$. 
In \cite{Repetti_A_2022_eusipco_dual_fbu} the authors studied the behavior of \eqref{algo:fbpnp}, using DDFB-LNO. They empirically emphasized that using warm restart for the network (i.e., using both primal and dual outputs from the network of the previous iteration) could add robustness to the PnP-FB algorithm, even when the number of layers $K$ is fixed, due to the monotonic behaviour of the FB algorithm on the dual variable. 
Finally, the regularization parameter $\lambda $ in \eqref{inverse-problem} aims to balance the data-fidelity term and the NN denoising power \cite{Repetti_A_2022_eusipco_dual_fbu}. 
The proposed {\UNN}s take as an input a parameter $\nu>0$ that has similar interpretation for the denoising problem \eqref{eq:prox-function}. In the PnP algorithm, we have $\nu = \lambda \gamma$, hence $\delta^2 = \beta^2 \sigma^2 \gamma$, where $\delta^2$ is the training noise level, and $\sigma^2$ is the noise level of the deblurring problem~\eqref{degradation-model}. The $\beta^2$ allows for flexibility in the choice of $\lambda$, to possibly improve the reconstruction quality.

\smallskip

In the remainder of the section, we will focus on unfolded {\UNN}s trained using \textbf{Training setting 2} described in Section~\ref{Ssec:exp:training} (i.e., with variable noise level) to better fit the noise level of the inverse problem~\eqref{degradation-model}.

\medskip

\noindent\textbf{Robustness comparison}\\
In the context of PnP methods, robustness can be measured in terms of convergence of the global algorithm. 
In this context, it is known that the PnP-FB algorithm converges if the NN is firmly non-expansive (see, e.g., \cite{pesquet2021learning} for details). 
According to \cite[Prop. 2.1]{pesquet2021learning}, an operator $f_\Theta$ is firmly non-expansive if and only if $h_\Theta = 2 f_\Theta - \operatorname{Id}$ is a $1$-Lipschitz operator, i.e., $ \chi_h = \max_\datad \Vert \operatorname{J} h_\Theta (\datad) \Vert_S < 1$. In the same paper, the authors used this result to develop a training strategy to obtain firmly non-expansive NNs. 

\begin{figure}[t!]
    \centering
    \begin{center}
        \includegraphics[trim={4.8cm 0cm 4.8cm 0cm},clip,scale=0.3]{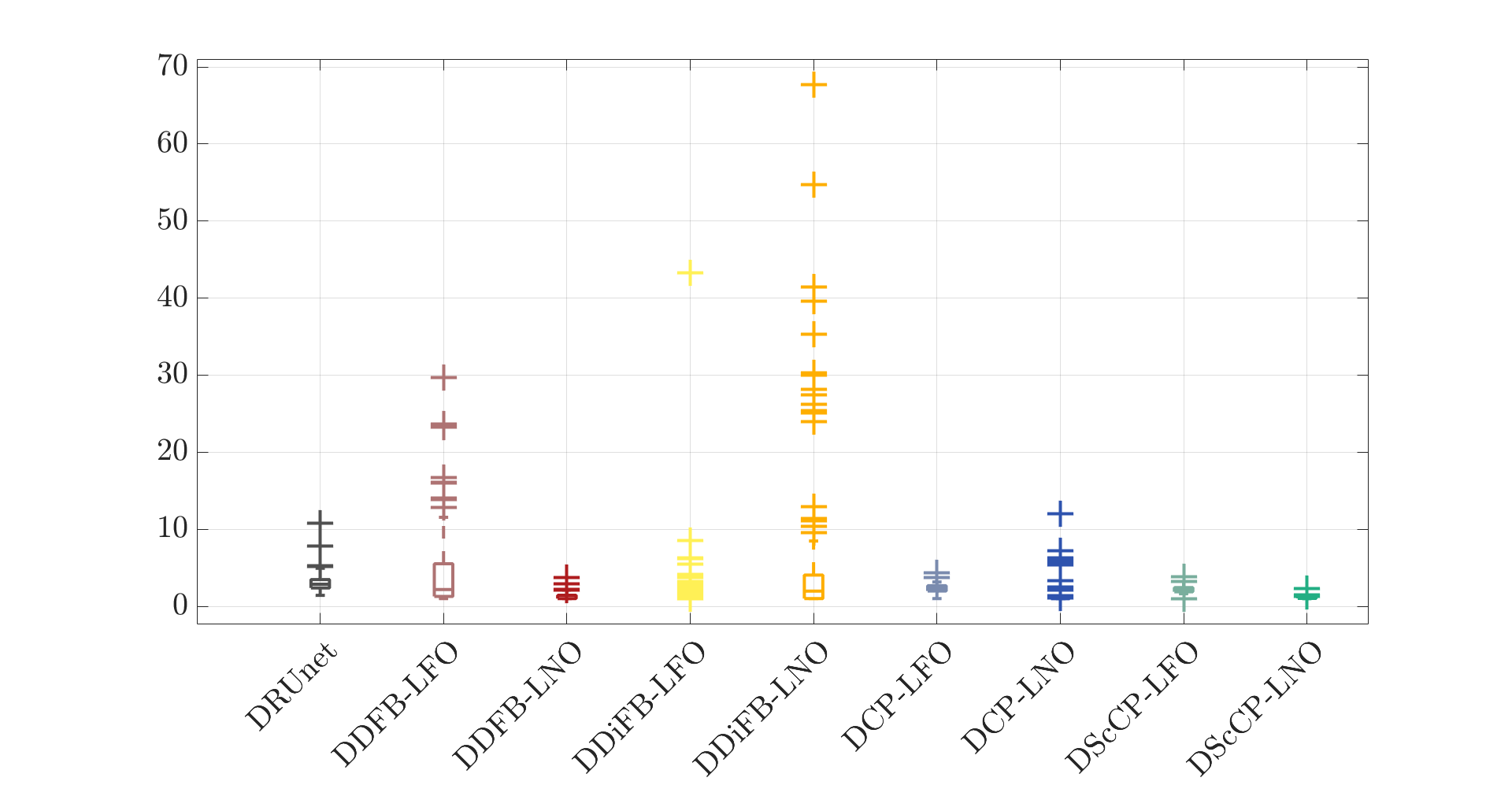}

        \vspace{0.1cm}
        {\footnotesize
        \setlength{\tabcolsep}{0.12cm}
        \begin{tabular}{@{}l||c|cc|cc|cc|cc@{}}
            &\multirow{2}{*}{DRUnet} & \multicolumn{2}{c|}{DDFB} & \multicolumn{2}{c|}{DDiFB} & \multicolumn{2}{c|}{DCP} & \multicolumn{2}{c}{DScCP} \\
            \cline{3-10}
            && LFO & LNO & LFO & LNO & LFO & LNO & LFO & LNO \\
            \hline\hline
            Mean    & $3.06$ & $4.42$ & $1.30$ & $2.17$ & $6.84$ & $2.34$ & $1.47$ & $2.21$ & $1.09$ \\
            Median  & $2.86$ & $2.17$ & $1.00$ & $1.32$ & $1.97$ & $2.38$ & $0.99$ & $2.13$ & $1.04$
        \end{tabular}}
    \end{center}

    \vspace{-0.2cm}
    
    \caption{\textbf{{\UNN} robustness comparison (firm non-expansiveness, Training Setting 2).} Distribution of $\| \operatorname{J} h_\Theta(\datad_s) \|_S$, where $h_\Theta = 2f_{\Theta}- \operatorname{Id}$ for $100$ images extracted from BSDS500 validation dataset $\mathbb J$, for the proposed {\UNN}s and DRUnet.}
    \label{fig:jac_f}
\end{figure}

Here we propose to use this result as a measure of robustness of the proposed {\UNN}s. Similarly to Section~\ref{Ssec:exp:robust-lip}, we approximate $\chi_h$ by computing $\chi_h \approx \max_{s \in \mathbb J} \| \operatorname{J} h_\Theta(\datad_s) \|_S$, where $\mathbb J$ contains $100$ images randomly selected from BSD500 validation set, and $\datad_s$ are noisy images with standard deviation uniformly distributed in $[0, 0.01]$. Then, the closer this value is to $1$, and the closer the associated NN $f_\Theta$ is to be firmly non-expansive. 
Figure~\ref{fig:jac_f} gives the box plots showing the distribution of $ (\Vert \operatorname{J} h_\Theta (\datad_s) \Vert_S )_{s \in \mathbb J}$. 
Conclusions on these results are very similar to those in Figure~\ref{fig:fne}, in particular that DDFB-LNO and DScCP-LNO have the smallest $\chi_h$ values.

\begin{figure}[t!]
	\setlength{\tabcolsep}{0.1cm}
	\scriptsize
    \hspace*{-0.1cm}\begin{tabular}{@{}c@{}cc@{}c@{}}
        \multicolumn{2}{c}{DDFB-LNO} & \multicolumn{2}{c}{DDFB-LFO} \\
        \includegraphics[trim={0.2cm 0.1cm 0.8cm 0.6cm},clip, width=0.25\columnwidth]{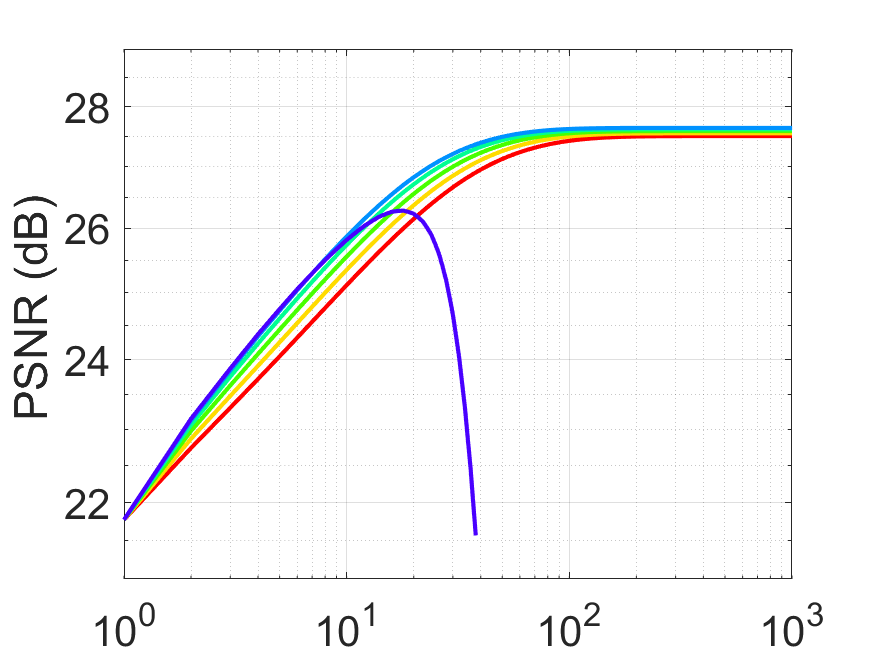}&
        \includegraphics[trim={0.2cm 0.1cm 0.8cm 0.4cm},clip, width=0.25\columnwidth]{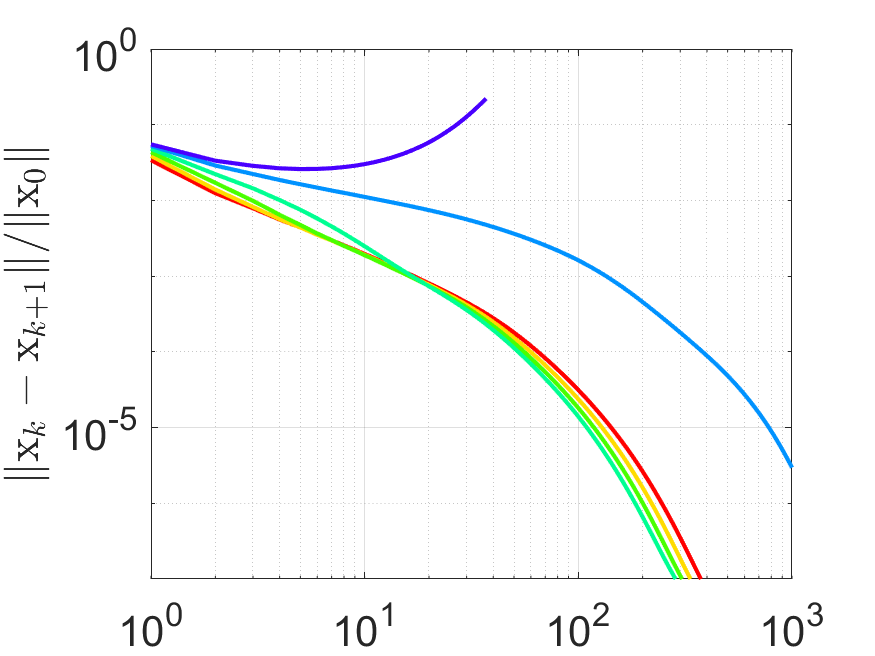}&
        \includegraphics[trim={0.2cm 0.1cm 0.8cm 0.6cm},clip, width=0.25\columnwidth]{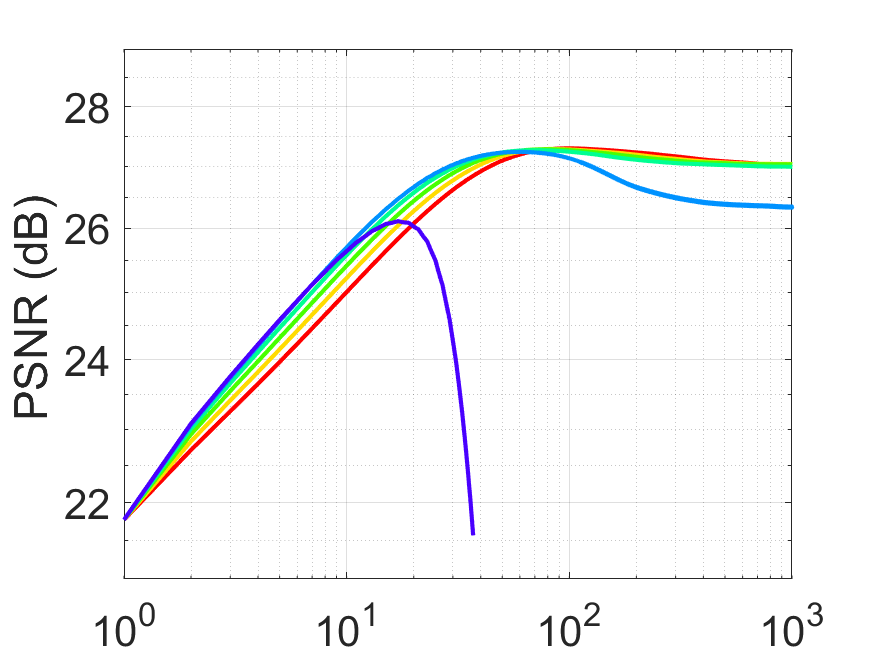}&
        \includegraphics[trim={0.2cm 0.1cm 0.8cm 0.4cm},clip, width=0.25\columnwidth]{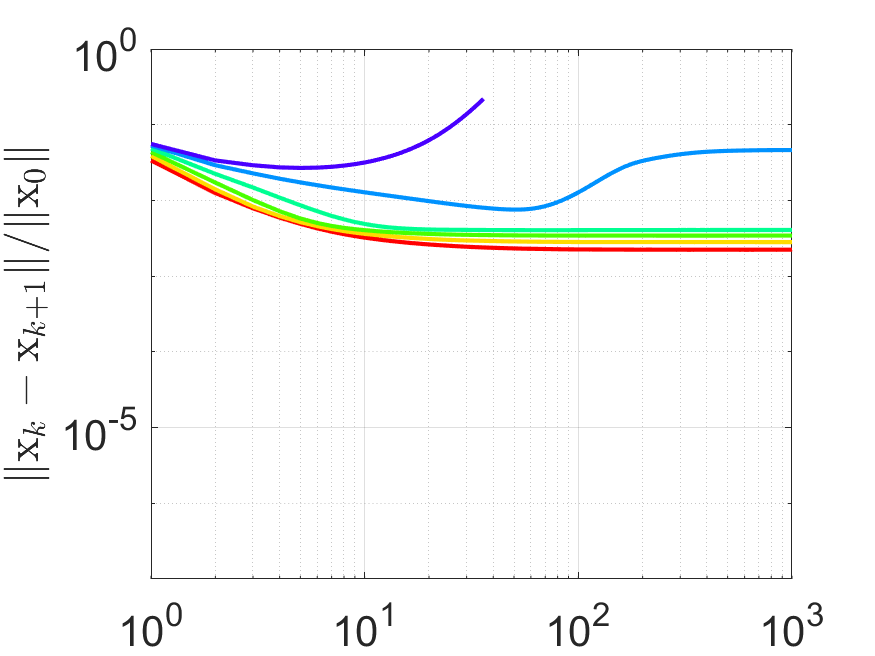} \\
        \multicolumn{2}{c}{DDiFB-LNO} & \multicolumn{2}{c}{DDiFB-LFO} \\
        \includegraphics[trim={0.2cm 0.1cm 0.8cm 0.6cm},clip, width=0.25\columnwidth]{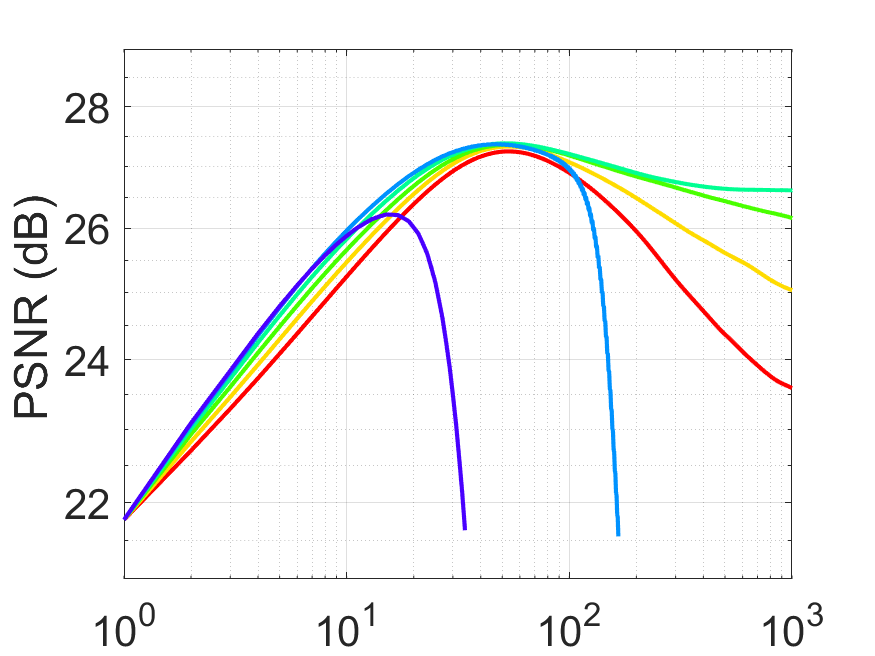}&
        \includegraphics[trim={0.2cm 0.1cm 0.8cm 0.4cm},clip, width=0.25\columnwidth]{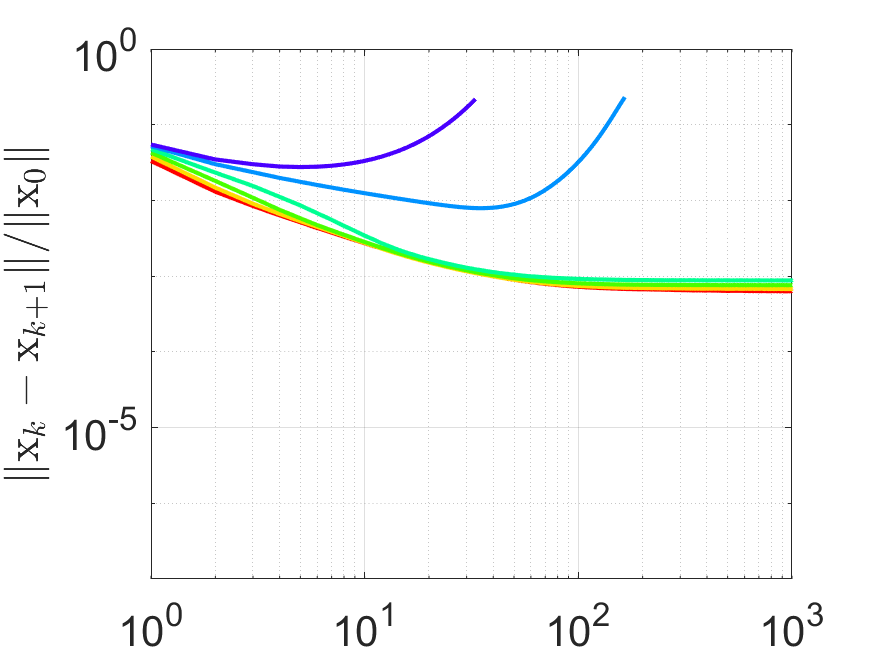}&
        \includegraphics[trim={0.2cm 0.1cm 0.8cm 0.6cm},clip, width=0.25\columnwidth]{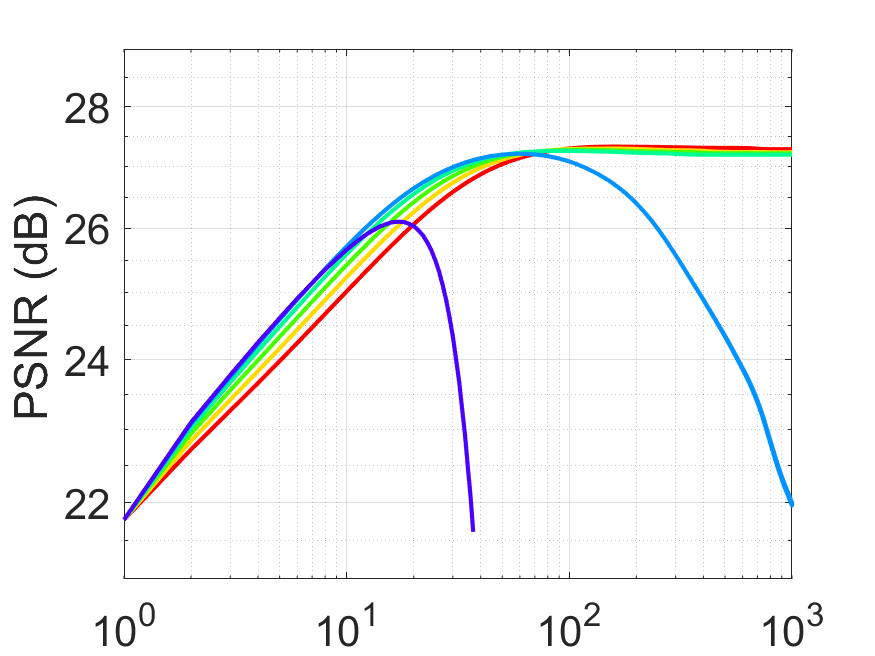}&
        \includegraphics[trim={0.2cm 0.1cm 0.8cm 0.4cm},clip, width=0.25\columnwidth]{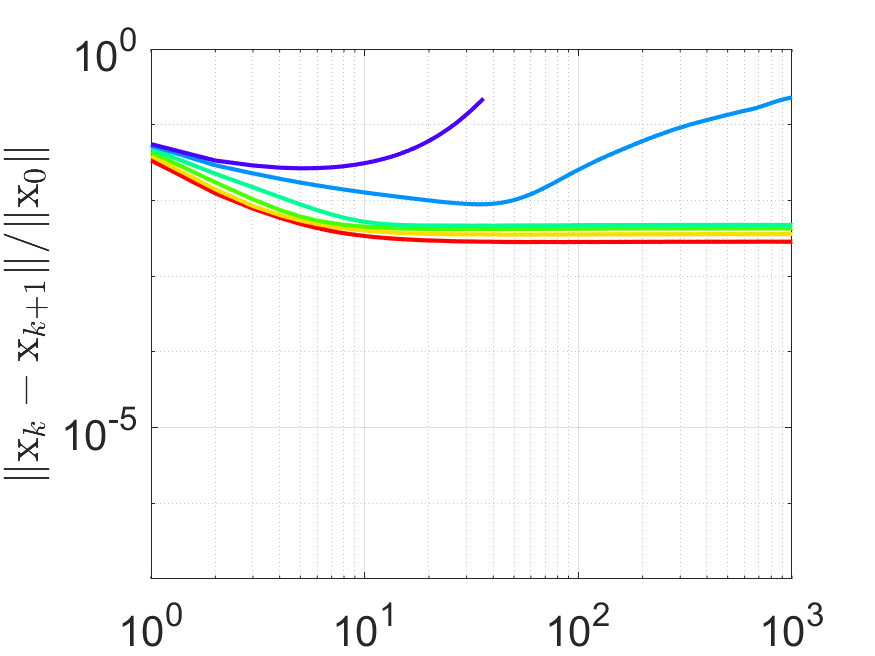} \\
        \multicolumn{2}{c}{DCP-LNO} & \multicolumn{2}{c}{DCP-LFO} \\
        \includegraphics[trim={0.2cm 0.1cm 0.8cm 0.6cm},clip, width=0.25\columnwidth]{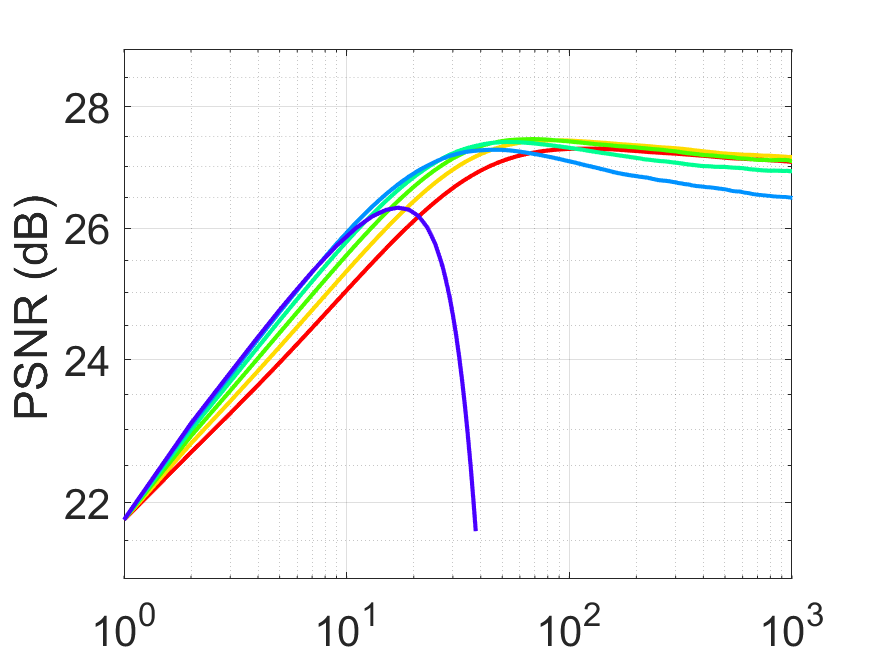}&
        \includegraphics[trim={0.2cm 0.1cm 0.8cm 0.4cm},clip, width=0.25\columnwidth]{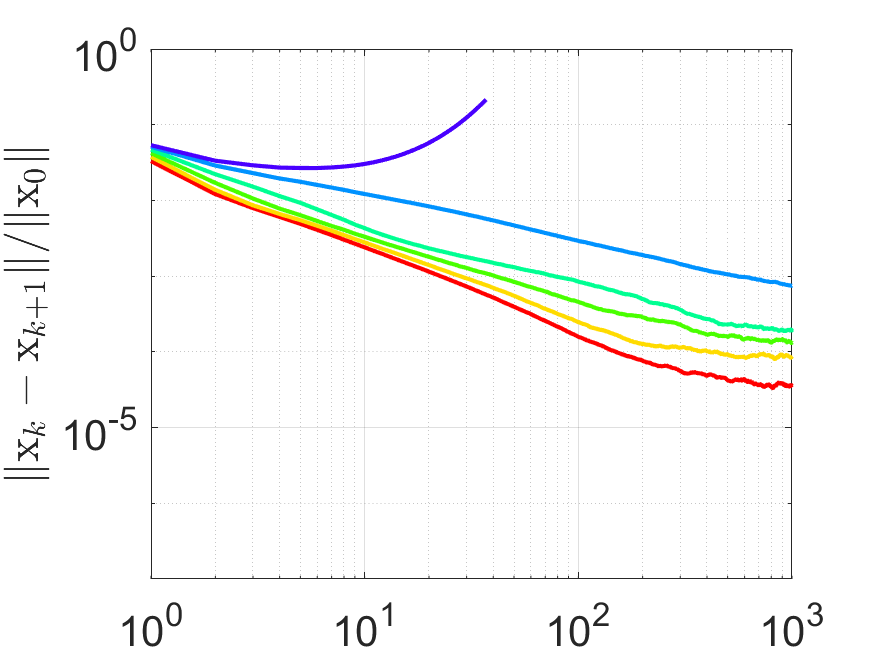}&
        \includegraphics[trim={0.2cm 0.1cm 0.8cm 0.6cm},clip, width=0.25\columnwidth]{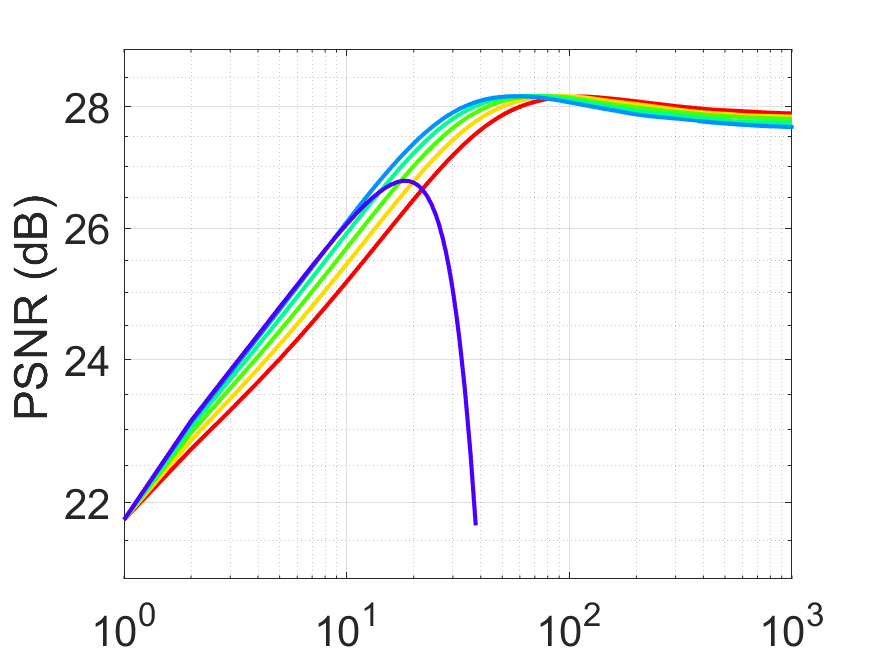}&
        \includegraphics[trim={0.2cm 0.1cm 0.8cm 0.4cm},clip, width=0.25\columnwidth]{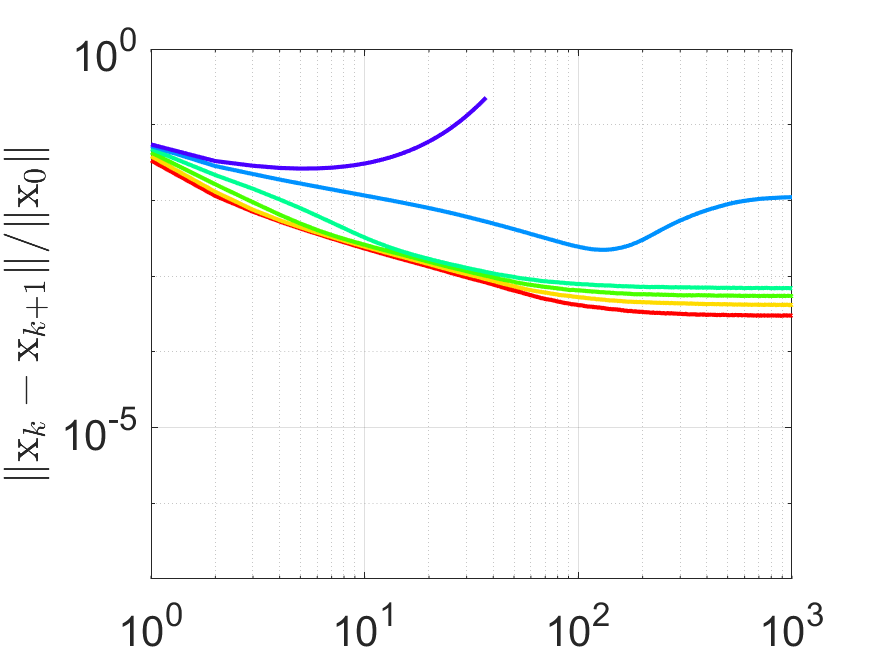} \\
        \multicolumn{2}{c}{DScCP-LNO} & \multicolumn{2}{c}{DScCP-LFO} \\
        \includegraphics[trim={0.2cm 0.1cm 0.8cm 0.6cm},clip, width=0.25\columnwidth]{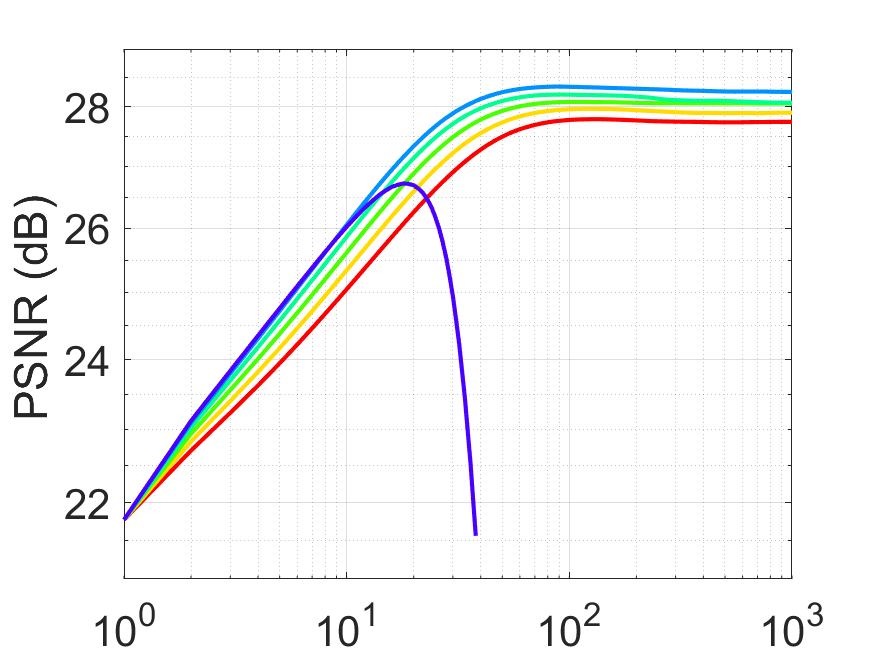}&
        \includegraphics[trim={0.2cm 0.1cm 0.8cm 0.4cm},clip, width=0.25\columnwidth]{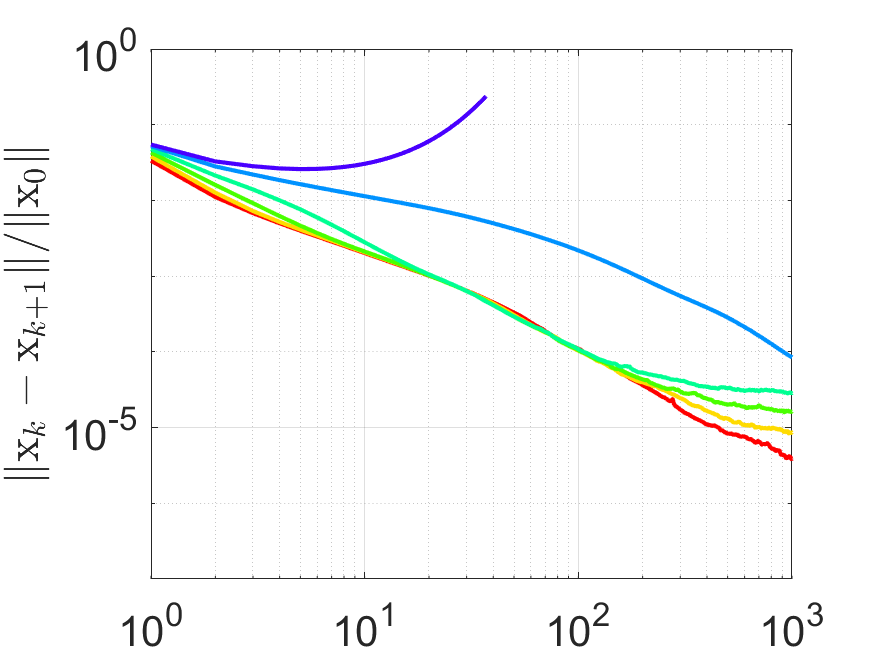}&
        \includegraphics[trim={0.2cm 0.1cm 0.8cm 0.6cm},clip, width=0.25\columnwidth]{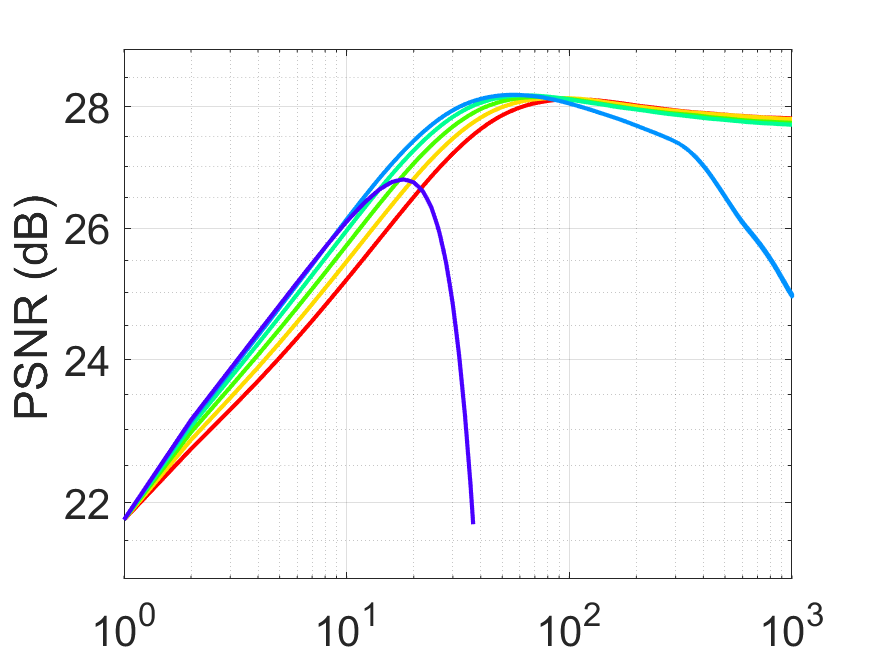}&
        \includegraphics[trim={0.2cm 0.1cm 0.8cm 0.4cm},clip, width=0.25\columnwidth]{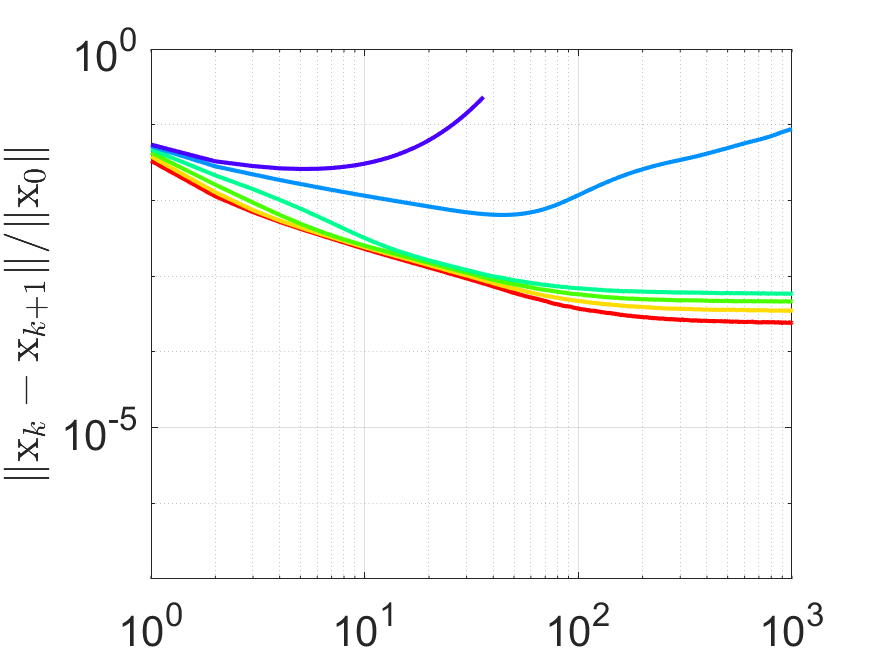} \\[0.1cm]
        \multicolumn{2}{c}{DRUnet} & \multicolumn{2}{c}{BM3D} \\
        \includegraphics[trim={0.2cm 0.1cm 0.8cm 0.6cm},clip, width=0.25\columnwidth]{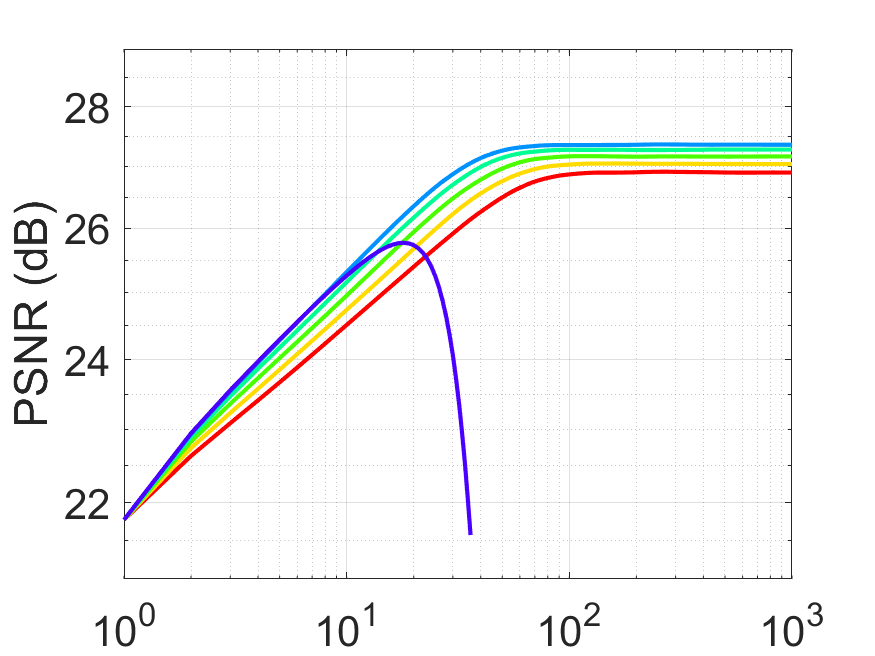}&
        \includegraphics[trim={0.2cm 0.1cm 0.8cm 0.4cm},clip, width=0.25\columnwidth]{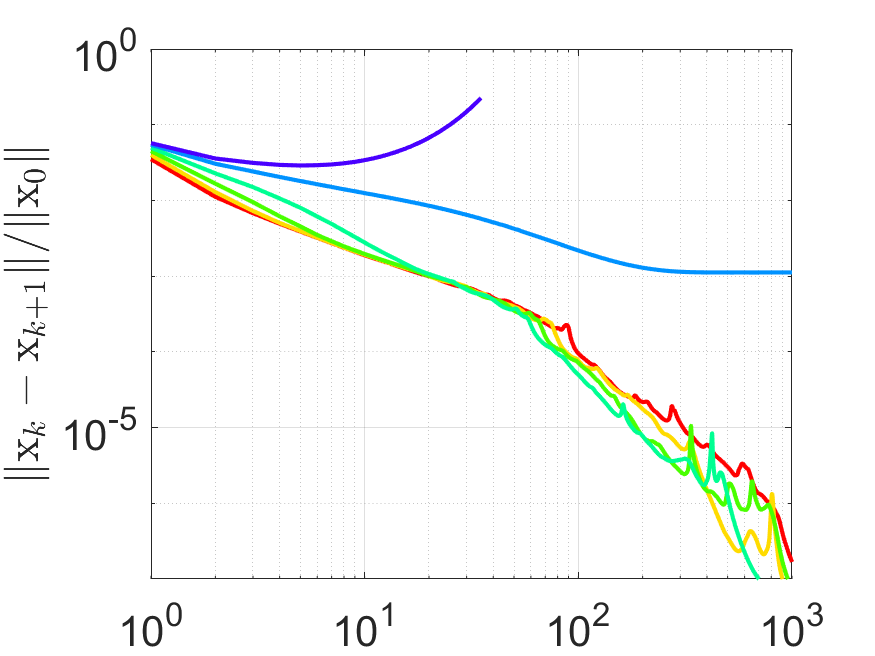}&
        \includegraphics[trim={0.2cm 0.1cm 0.8cm 0.6cm},clip, width=0.25\columnwidth]{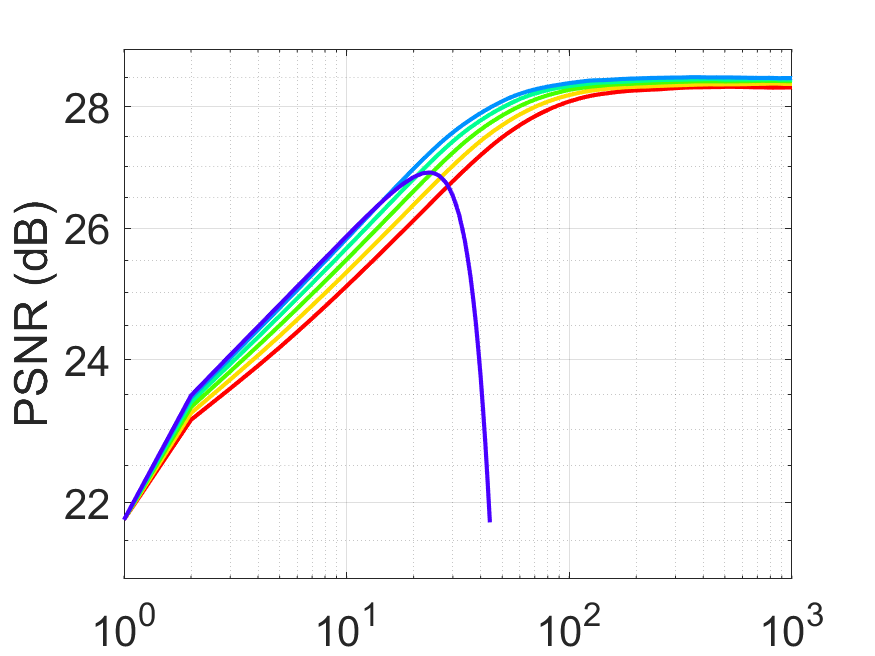}&
        \includegraphics[trim={0.2cm 0.1cm 0.8cm 0.4cm},clip, width=0.25\columnwidth]{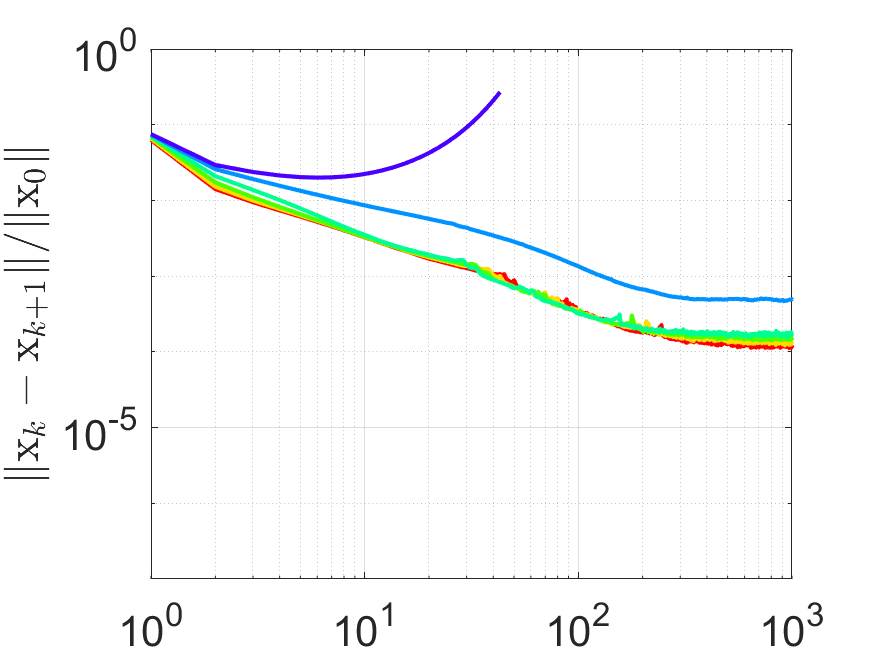} \\[-0.1cm]
         \multicolumn{4}{c}{\includegraphics[width=6cm]{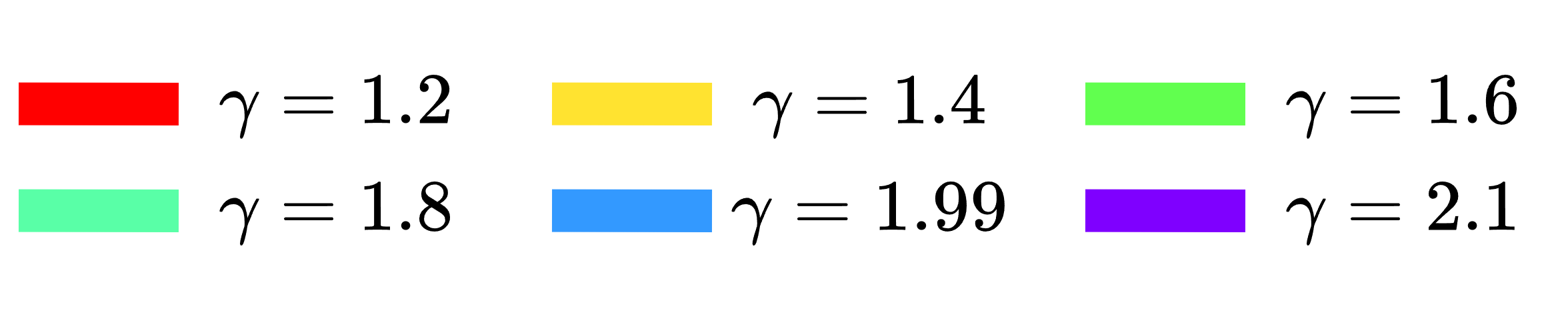}} 
    \end{tabular}

\vspace*{-0.3cm}

 \caption{\textbf{Restoration performance for deblurring (Training Setting 2): Parameter choice ($\gamma$).} Convergence behavior (PSNR and relative error norm) of the PnP-FB algorithms for fixed $\beta=1$, and varying $\gamma\in \{1.2,1.4,1.6,1.8,1.99,2.1\}$.}
 \label{fig:PnP-FB-dif-gamma}
\end{figure}

\begin{figure}[t!]
     \centering
     \scriptsize
    \begin{tabular}{c}
         \includegraphics[trim={3.8cm 0.cm 0 0cm},clip,width=10cm]{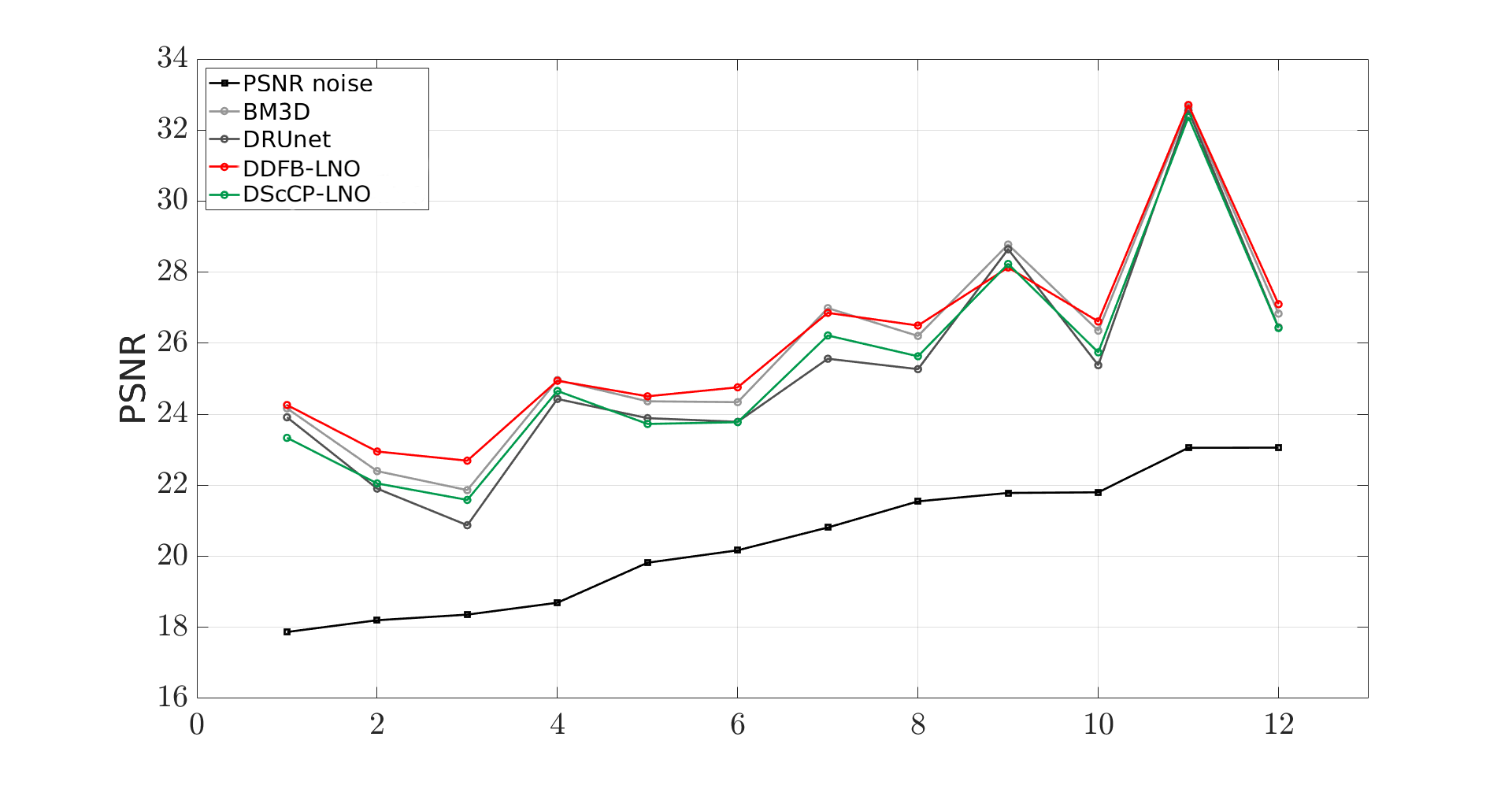}\\[-0.2cm] 
         PSNR with optimal $\beta$
    \end{tabular}
    \caption{\textbf{Restoration performance for deblurring (Training Setting 2).} 
    Best PSNR values obtained with DDFB-LNO, DScCP-LNO, DRUnet and BM3D, on $12$ images from BSDS500 validation set degraded according to~\eqref{degradation-model}, with $\sigma=0.03$. 
   }
    \label{fig:IR-compare}

    \vspace*{-0.2cm}
\end{figure}

\medskip

\noindent\textbf{Parameter choices}\\
As emphasized previously, the denoising NNs in PnP-FB algorithm~\eqref{algo:fbpnp} depend on two parameters: the step-size $\gamma$ and the regularization parameter $\lambda=\beta^2\sigma^2$. 
In this section we investigate the impact of the step-size on the results, while the regularization parameter is fixed to $\beta=1$.

\smallskip

To ensure the convergence of the PnP-FB algorithm, the stepsize must satisfy $\gamma\in (0,2/\Vert \mathrm{A} \Vert_S^2)$. 
Figure~\ref{fig:PnP-FB-dif-gamma} aims to evaluate the stability of the proposed unfolded NNs by looking at the convergence of the associated PnP-FB iterations, varying $\gamma  \in \{1.2, 1.4, 1.6, 1.8, 1.99, 2.1\}$ (where $\|\mathrm{A}\|_S=1$). For this experiment, we fix $\lambda=\sigma^2$ (i.e., $\beta=1$). 
Further, we also consider PnP algorithms using either DRUnet or BM3D as denoisers. 
The plots show the convergence profiles for the deblurring of one image in terms of PSNR values and  relative error norm of consecutive iterates $\|\estmsd_{t+1} - \estmsd_{t}\|/\|\estmsd_0\|$, with respect to the iterations. 
In theory, $(\|\estmsd_{t+1} - \estmsd_{t}\|)_t$ should decrease monotonically.
Interestingly, we can draw similar conclusions from the curves in Figure~\ref{fig:PnP-FB-dif-gamma} as for Figures~\ref{fig:jac_f} and~\ref{fig:fne}. Looking at the relative error norms, the most robust NNs are DDFB-LNO and DScCP-LNO, both converging monotonically for any choice of $\gamma \le 1.99$. DCP-LNO seems to have similar convergence profile, with a slower convergence rate. None of the other PnP schemes seems to be stable for $\gamma = 1.99$.
For $\gamma \le 1.8$, DRUnet shows an interesting convergence profile, however the error norm is not decreasing monotonically. 
The remaining PnP schemes do not seem to converge in iterates, as the error norms reach a plateau. 
In terms of PSNR values, DScCP-LNO and BM3D have the best performances, followed by DDFB-LNO and DRUnet. For these four schemes, $\gamma=1.99$ leads to the best PSNR values.

\begin{table}[t!]
    \centering 
    \caption{\textbf{Restoration performance for deblurring (Training Setting 2).} 
    Average PSNR values over $12$ images from BSDS500 validation set, obtained with different PnP-FB schemes.
    For each NN, $\beta$ was chosen to obtain the highest PSNR.}
    \label{tab:dif_noise_level}
    \footnotesize
    \begin{tabular}{@{}c|c|c|c|c|c@{}}
         $\sigma$& Noisy&BM3D&DRUnet& DDFB-LNO&DScCP-LNO\\\hline\hline
        $0.015$& 20.80&  \textcolor{red}{\textbf{28.33}}& 26.47 &\textcolor{blue}{\textbf{ 28.16}}&27.81 \\ \hline
        $0.03$ & 20.43  & \textcolor{blue}{\textbf{25.82}} & 25.14 &\textcolor{red}{\textbf{26.00}}  & 25.31\\ \hline
        $0.05$ & 19.68& \textcolor{blue}{\textbf{24.27}}&  23.98 &   \textcolor{red}{\textbf{24.37}}&  23.87 
    \end{tabular}
\end{table}

\medskip

\noindent\textbf{Restoration performance comparison}\\
In this section, we perform further comparisons of the different PnP schemes on $12$ random images selected from BSDS500 validation set, degraded as per model~\eqref{degradation-model}. In particular, we will run experiments for three different noise levels $\sigma\in \{0.015, 0.03, 0.05\}$. 
Since in the previous sections we observed that DDFB-LNO and DScCP-LNO are the more robust unfolded strategies, in this section we only focus on these two schemes, comparing them to BM3D and DRUnet. 
For each denoiser, we choose $\delta = \lambda \gamma$, with $\gamma= 1.99$ and $\lambda = \beta^2\sigma^2$. 
As explained after Theorem~\ref{thm:cvg-pnpfb}, parameter $\beta^2$ allows for flexibility to possibly improve the reconstruction quality. In the remainder we choose it to optimize the reconstruction quality of each image (PSNR value).

\begin{figure}[t!]
\setlength{\tabcolsep}{0.1cm}
\centering
\scriptsize
    \begin{tabular}{@{}cc@{}}
         Ground truth & Noisy ($\sigma = 0.015$) -- $20.11$~dB\\
         \includegraphics[width=5.5cm]{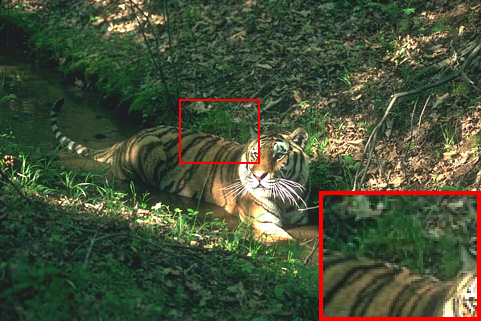}&
         \includegraphics[width=5.5cm]{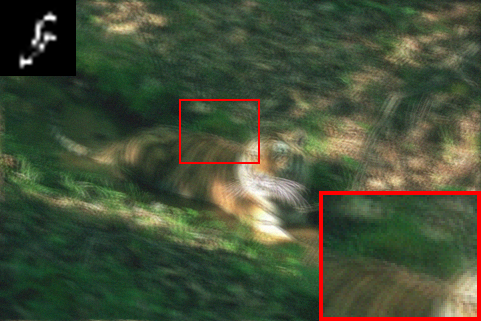}\\
         BM3D -- \textbf{\textcolor{blue}{27.10}}~dB & DRUnet -- $25.09$~dB \\
          \includegraphics[width=5.5cm]{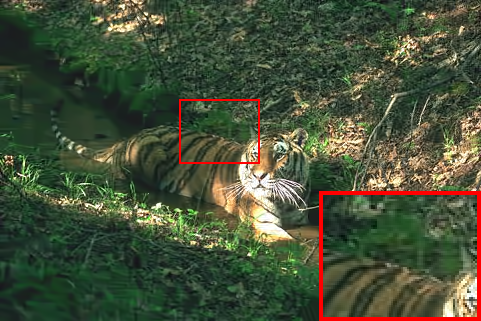}&
         \includegraphics[width=5.5cm]{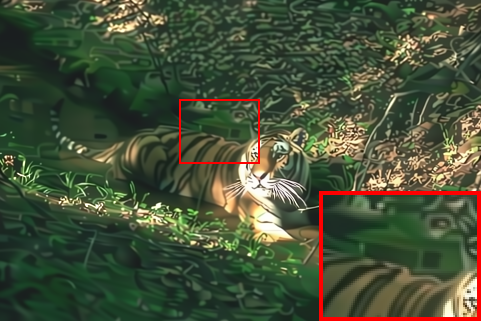}\\
         DDFB-LNO -- \textbf{\textcolor{red}{27.23}}~dB & DScCP-LNO -- $26.48$~dB\\
         \includegraphics[width=5.5cm]{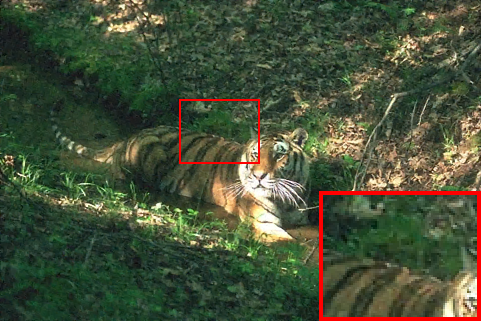}&
         \includegraphics[width=5.5cm]{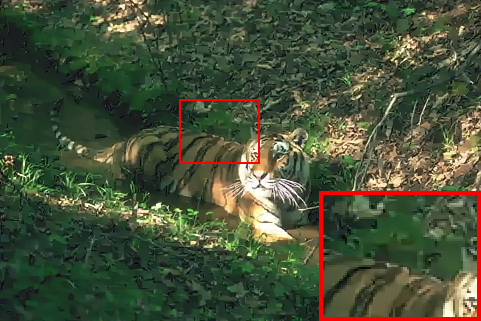}
    \end{tabular}
    \caption{\textbf{Restoration performance for deblurring (Training Setting 2).}  Restoration example for $\sigma=0.015$, with parameters $\gamma=1.99$ and $\beta$ chosen optimally for each scheme.}
    \label{fig:PnP-FB-results-ex1}
\end{figure}

\begin{figure}[t!]
\setlength{\tabcolsep}{0.1cm}
\centering
\scriptsize
    \begin{tabular}{@{}cc@{}}
         Ground truth & Noisy ($\sigma = 0.03$) -- $23.06$~dB \\
         \includegraphics[width=5.5cm]{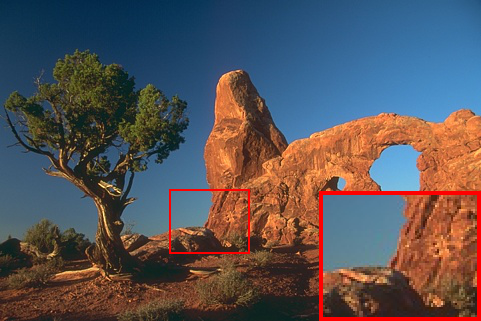}&
         \includegraphics[width=5.5cm]{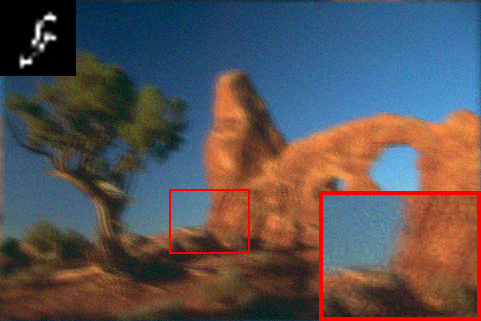}\\
         BM3D -- \textbf{\textcolor{blue}{26.83}}~dB& DRUnet -- $26.44$~dB \\
          \includegraphics[width=5.5cm]{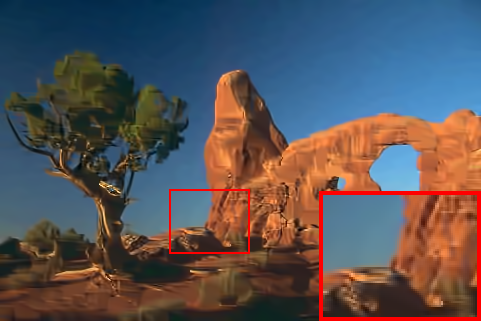}&
         \includegraphics[width=5.5cm]{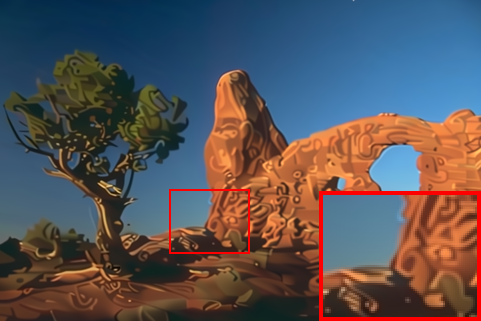}\\
         DDFB-LNO -- \textbf{\textcolor{red}{27.09}}~dB &DScCP-LNO -- $26.43$~dB\\
         \includegraphics[width=5.5cm]{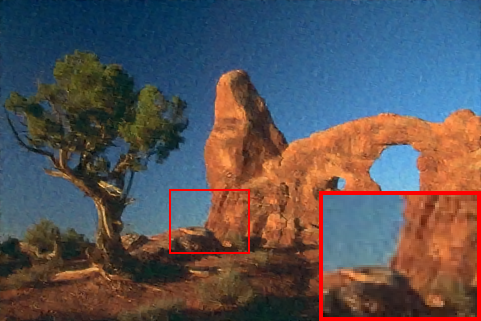}&
         \includegraphics[width=5.5cm]{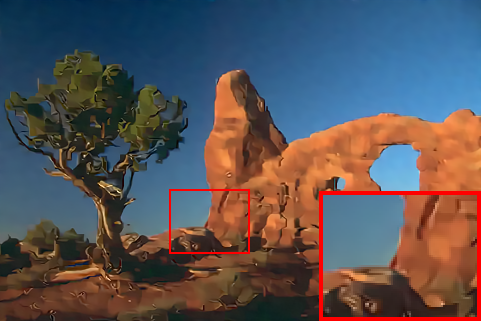}
    \end{tabular}
    \caption{\textbf{Restoration performance for deblurring (Training Setting 2).}  Restoration example for $\sigma=0.03$, with parameters $\gamma=1.99$ and $\beta$ chosen optimally for each scheme.}
    \label{fig:PnP-FB-results-ex2}
\end{figure}

In Figure~\ref{fig:IR-compare}, we provide best PSNR values obtained (for optimized values of $\beta$) with DDFB-LNO, DScCP-LNO, DRUnet and BM3D, on $12$ images from BSDS500 validation set degraded according to~\eqref{degradation-model}, with $\sigma=0.03$. 
In addition, the averaged PSNR values obtained with the four different schemes for the different noise levels $\sigma\in \{0.015, 0.03, 0.05\}$ are given in Table~\ref{tab:dif_noise_level}. 
It can be observed that, regardless the noise level $\sigma$, DDFB-LNO and BM3D always have the highest PSNR values, outperforming DRUnet and DScCP-LNO. However, DDFB-LNO has a much cheaper computation time than BM3D (see Table~\ref{tab:runtime+flop} for details). For low noise level $\sigma=0.015$, DScCP-LNO also outperforms DRUnet. For highest noise levels $\sigma \in \{0.03, 0.05\}$, DScCP-LNO and DRUnet have similar performances. 

For visual inspection, we also provide in Figures~\ref{fig:PnP-FB-results-ex1} and \ref{fig:PnP-FB-results-ex2} examples of two different images, for noise levels $\sigma=0.015$, $\sigma=0.03$ and $\sigma=0.05$, respectively. 
We observe that DRUnet and DScCP-LNO tend to better eliminate noise, compared to BM3D and DDFB-LNO. This is consistent with the denoising performance results observed in Section~\ref{Ssec:exp:perf-denoise}. However, when integrated in a PnP for the restoration problem, DRUnet seems to smooth out high-frequency details, sometimes resembling an AI generator and generate unrealistic patterns. On the contrary, DDFB-LLNO retains more high-frequency details. DScCP-LNO produces images much smoother that DDFB-LNO, but without unrealistic patterns.
Hence, if an image contains a significant amount of high-frequency details compared to piecewise smooth or piecewise constant patterns, unfolded NN denoisers like DScCP-LNO or DDFB-LNO are more suitable choices than DRUnet.

\section{Conclusion}
\label{s:conclusion}

In this work, we have presented a unified framework for building denoising {\UNN}s with learned linear operators, whose architectures are based on D(i)FB and (Sc)CP algorithms. 
We show through simulations that the proposed {\UNN}s have similar denoising performances to DRUnet, although being much lighter ($\sim 1000$ less parameters). In particular, we observed that the use of inertia in the unfolding strategies improves the denoising efficiency, i.e., DDiFB (resp. DScCP) is a better denoiser than DDFB (resp. DCP). Similarly, we observe that adopting a learning strategy close to theoretical convergence conditions leads to higher denoising performance.
Beside, we show that the proposed {\UNN}s are generally more robust than DRUnet, in particular for DDFB-LNO, DCP-LNO and DScCP-LNO. 
Robustness is assessed in terms of the network Lipschitz constants, by applying the networks non-Gaussian noises, and by injecting the networks in PnP algorithms for a deblurring task. In particular, DDFB-LNO and DScCP-LNO exhibit the best trade-off in terms of denoising/reconstruction performances, and robustness.  

In future work, we aim to extend our study to {\UNN}s designed to handle different imaging problems. 
Firstly, a similar approach could be investigated for non-Gaussian denoising, where the least-squares data-fidelity term would be replaced by a bespoke function associated with the distribution of interest. 
Secondly, a similar study could be performed for {\UNN}s designed to solve image restoration problems. In this case, the data-fidelity term would not be strongly convex anymore, hence different acceleration strategies may need to be explored.

\section{Annex}
\label{s:annex}

The proposed ScCP iterations~\eqref{eq:sccp} for minimizing~\eqref{eq:prox-function} is equivalent to the original ScCP algorithm proposed in \cite{chambolle2011first}:
\begin{align}
\label{eq:sccp-inter}
\begin{array}{l}
    \text{for } k = 0, 1, \ldots \\
    \left\lfloor 
    \begin{array}{l}  
    \estmsd_{k+1} 
    =   \prox_{\mu_k(\frac12 \| \cdot - \datad \|^2 + \iota_C)} \left( \datad - \mu_k \D^\top\dualvar_k  \right), \\
    \dualvar_{k+1} =\prox_{\tau_k (\nu g)^*} \left( \dualvar_k+\tau_k \D \Big( (1+\alpha_k ) \estmsd_{k+1}-\alpha_{k}\estmsd_{k} \Big) \right).
	\end{array}
    \right. 
\end{array}
\end{align}
This can be shown by noticing that, for every $\mathrm{v}\in \mathbb{R}^N$,
\begin{align*}
    \prox_{\mu_k(\frac12 \| \cdot - \datad \|^2 + \iota_C)}  (\mathrm{v}) 
    & = \underset{\estmsd \in \mathbb{R}^N}{\text{argmin }} \frac{1+\mu_k}{2\mu_k} \|\estmsd - \frac{ \mathrm{v}+\mu_k \datad}{1+\mu_k}\|^2+ \iota_{C}\left( \estmsd \right)\\
    & = \proj_C \left( \frac{\mathrm{v} + \mu_k \datad }{1+\mu_k} \right).
\end{align*}

\small\bibliographystyle{IEEEbib}
\bibliography{strings,refs}
\end{document}